\documentclass[11.5pt]{article}
\usepackage[top=1.25in, bottom=1.25in, left=1.25in, right=1.25in]{geometry}
\usepackage{amssymb}

\usepackage{amsmath}
\usepackage{xspace}
\usepackage{color}
\usepackage{xcolor}
\usepackage[font=footnotesize]{caption} 
\usepackage{subcaption}
\usepackage{url}
\usepackage{comment}
\usepackage{verbatim}
\usepackage{amsthm} 
\usepackage{framed} 
\usepackage{epsfig}
\usepackage{graphicx}
\usepackage{bm}
\usepackage{bbm} 

\usepackage{rotating}
\usepackage{ulem}
\usepackage{algorithm}
\usepackage{algpseudocode}
\usepackage{soul} 
\usepackage{setspace} 
\usepackage{soul} 
\usepackage{afterpage} 
\usepackage{mathabx} 

\usepackage[open,openlevel=2,numbered]{bookmark}
\hypersetup{pdftex,  
  breaklinks=true, 
  colorlinks=true,
  linkcolor=blue,
  citecolor=blue,
  pdfpagemode=UseNone, 
}

\usepackage{etex}
\usepackage{pgf}
\usepackage{tikz}
\usetikzlibrary{automata}
\pgfdeclarelayer{background}
\pgfsetlayers{main}
\usetikzlibrary{shapes}
\usetikzlibrary{matrix} 

\newcommand{\inmath}	{\ensuremath}
\newcommand{\xmath}[1]	{\ensuremath{#1}\xspace}
\newcommand{\bmath}[1]	{\xmath{\bm{#1}}}	

\newcommand{\argmin}[1] 	{\operatorname*{arg\,min}_{#1}}

\newcommand{\minimize}[1] 	{\operatorname*{minimize}_{#1}}

\newcommand{\reals}			{\xmath{\mathbb{R}}}

\newcommand{\bmat}		{\begin{bmatrix}} 
\newcommand{\emat}		{\end{bmatrix}}

\newcommand{\braces}[1]	{\xmath{\left\{#1\right\}}}

\newcommand{\abs}[1]	{\xmath{\left| #1 \right|}}
\newcommand{\norm}[1]	{\xmath{\left\| #1 \right\|}}

\newcommand{\inprod}[1]	{\xmath{\mathop{\langle #1\rangle}\nolimits}}

\newcommand{\inv}		{\ensuremath{^{-\!1}}}

\newcommand{\dstyle}{\displaystyle}

\newcommand{\wordbrace}[3][]{\,\inmath{\mathrm{#2}#1\!\braces{#3}}}

\newcommand{\diag}[1]	{\wordbrace{diag}{#1}}

\newcommand{\ba}[1]		{\left[ \begin{array}{#1}}
\newcommand{\ea}		{\end{array} \right]}
\newcommand{\be}		{\begin{equation}}
\newcommand{\ee}[1]		{\label{#1}\end{equation}}

\newcommand{\ie}		{\emph{i.e\@.}\xspace}
\newcommand{\eg}		{\emph{e.g\@.}\xspace}

\newcommand{\apriori}	{\emph{a priori}\xspace}

\theoremstyle{plain}

\theoremstyle{definition}

\theoremstyle{remark}

\newcommand{\imwidth}  {0.23\linewidth}
\newcommand{\imheight} {0.21\linewidth}

\newcommand{\VSPACE}{\vspace{10pt}}
\newcommand{\VSPACEE}{\vspace{10pt}}
\newcommand{\imwidthh}  {0.23\linewidth}

\newcommand{\myfbox}[1]{\setlength{\fboxrule}{1.25pt}\fbox{#1}\setlength{\fboxrule}{0.25pt}}

\definecolor{DarkBlue}{rgb}{0,0.2,0.65}
\definecolor{black}{rgb}{0,0,0}
\definecolor{myblack}{rgb}{0,0,0}
\definecolor{myred}{rgb}{0.85,0,0}
\definecolor{mygreen}{rgb}{0,0.6,0}
\definecolor{myblue}{rgb}{0,0,0.5}

\definecolor{myorange}{rgb}{1,0.6,0.2}
\definecolor{myyellow}{rgb}{1,1,0}
\definecolor{mypurple}{rgb}{.7,0,.7}
\definecolor{mywhite}{rgb}{1,1,1}

\newcommand{\dored}[1]		{{\color{myred}{#1}}}

\newcommand{\mytilde}{\raise.17ex\hbox{$\scriptstyle\mathtt{\sim}$}}

\newcommand{\Bu} {\xmath{\bmath{u}}}
\newcommand{\Bv} {\xmath{\bmath{v}}}
\newcommand{\Bw} {\xmath{\bmath{w}}}

\newcommand{\Bz} {\xmath{\bmath{z}}}

\newcommand{\BA} {\xmath{\bmath{A}}}

\newcommand{\BH} {\xmath{\bmath{H}}}
\newcommand{\BI} {\xmath{\bmath{I}}}

\newcommand{\Bzeta}			{\xmath{\bmath{\zeta}}}

\newcommand{\BLambda}		{\xmath{\bmath{\Lambda}}}

\newcommand{\Bphi}			{\xmath{\bmath{\phi}}}

\newcommand{\w}			{{\bmath{w}}\xspace}

\renewcommand{\b}			{{\bmath{b}}\xspace}

\newcommand{\x}			{{\bmath{x}}\xspace}
\newcommand{\xtil}			{{\bmath{\tilde{x}}}\xspace}
\renewcommand{\r}			{{\bmath{r}}\xspace}

\newcommand{\q}			{{\bmath{q}}\xspace}
\newcommand{\wtil}			{{\bmath{\widetilde{w}}}\xspace}
\renewcommand{\u}			{{\bmath{u}}\xspace}

\newcommand{\va}			{{\bmath{v_1}}\xspace}
\newcommand{\vb}			{{\bmath{v_2}}\xspace}
\newcommand{\vc}			{{\bmath{v_3}}\xspace}
\newcommand{\vd}			{{\bmath{v_4}}\xspace}
\newcommand{\ua}			{{\bmath{u_1}}\xspace}
\newcommand{\ub}			{{\bmath{u_2}}\xspace}
\newcommand{\uc}			{{\bmath{u_3}}\xspace}
\newcommand{\ud}			{{\bmath{u_4}}\xspace}

\newcommand{\Ctil}			{{\bmath{\widetilde{C}}}\xspace}

\newcommand{\A}			{{\bmath{A}}\xspace}
\newcommand{\B}			{{\bmath{B}}\xspace}
\newcommand{\C}			{{\bmath{C}}\xspace}

\newcommand{\Q}			{{\bmath{Q}}\xspace}
\newcommand{\I}			{{\bmath{I}}\xspace}
\newcommand{\U}			{{\bmath{U}}\xspace}
\newcommand{\X}			{{\bmath{X}}\xspace}
\newcommand{\Y}			{{\bmath{Y}}\xspace}
\newcommand{\YXw}			{{\bmath{YXw}}\xspace}

\newcommand{\bzero}			{{\bmath{0}}\xspace}

\newcommand{\Abar}			{{\bmath{\bar{A}}}\xspace}
\newcommand{\Bbar}			{{\bmath{\bar{B}}}\xspace}
\newcommand{\xbar}			{{\bmath{\bar{x}}}\xspace}
\newcommand{\ybar}			{{\bmath{\bar{y}}}\xspace}

\newcommand{\fbar}			{{\bmath{\bar{f}}}\xspace}
\newcommand{\gbar}			{{\bmath{\bar{g}}}\xspace}

\newcommand{\prox}		{\xmath{\mathrm{Prox}}} 
\newcommand{\soft}		{\xmath{\mathrm{Soft}}}

\newcommand{\loss}		{\xmath{\ell}}
\newcommand{\Loss}		{\xmath{\mathcal{L}}}

\newcommand{\Reg}{\xmath{\mathcal{R}}}

\newcommand{\elltwo} {\xmath{\ell_2}}
\newcommand{\ellone} {\xmath{\ell_1}}

\newcommand{\normsq}[1]{\norm{#1}^2}

\newcommand{\sign}[1] {\text{sign}\left(#1\right)} 

\newcommand{\iter}{^{(t)}}
\newcommand{\iterp}{^{(t+1)}}

\newcommand{\pbar}{{\xmath{\bar{p}}}}
\newcommand{\qbar}{{\xmath{\bar{q}}}}

\newcommand{\ptil}{{\xmath{\tilde{p}}}}
\newcommand{\etil}{{\xmath{\tilde{e}}}}

\newcommand{\xstar} {\xmath{\x^*}}
\newcommand{\pstar} {{\xmath{p^*}}}

\newcommand{\varepsADMM}{\varepsilon}

\usepackage{lineno}
\newcommand*\patchAmsMathEnvironmentForLineno[1]{%
  \expandafter\let\csname old#1\expandafter\endcsname\csname #1\endcsname
  \expandafter\let\csname oldend#1\expandafter\endcsname\csname end#1\endcsname
  \renewenvironment{#1}%
     {\linenomath\csname old#1\endcsname}%
     {\csname oldend#1\endcsname\endlinenomath}}%
\newcommand*\patchBothAmsMathEnvironmentsForLineno[1]{%
  \patchAmsMathEnvironmentForLineno{#1}%
  \patchAmsMathEnvironmentForLineno{#1*}}%
\AtBeginDocument{%
\patchBothAmsMathEnvironmentsForLineno{equation}%
\patchBothAmsMathEnvironmentsForLineno{align}%
\patchBothAmsMathEnvironmentsForLineno{flalign}%
\patchBothAmsMathEnvironmentsForLineno{alignat}%
\patchBothAmsMathEnvironmentsForLineno{gather}%
\patchBothAmsMathEnvironmentsForLineno{multline}%
}

\allowdisplaybreaks 
\newcommand{\citep}[1]{\cite{#1}}

\DeclareTextFontCommand{\emph}{\itshape}
\begin{document}
	\title{Disease Prediction based on Functional Connectomes using a \\ Scalable and Spatially-Informed Support Vector Machine}
\author{Takanori Watanabe$^a$, Daniel Kessler$^c$, Clayton Scott$^{a,b}$, \\ \vspace{10pt}
	Michael Angstadt$^c$, Chandra Sripada$^c$\\	
	$^a$Department of Electrical Engineering and Computer Science\\
	$^b$Department of Statistics\\	
	$^c$Department of Psychiatry\\	
	University of Michigan, Ann Arbor, MI, USA
	}
\date{}
\maketitle


\begin{abstract}
Substantial evidence indicates that major psychiatric disorders are associated with distributed \mbox{neural} dysconnectivity, leading to strong interest in using neuroimaging methods to accurately predict disorder status.
In this work, we are specifically interested in a multivariate approach that uses features derived from whole-brain resting state functional connectomes.
However, functional connectomes reside in a high dimensional space, which complicates model interpretation and introduces numerous statistical and computational challenges. 
Traditional feature selection techniques are used to reduce data dimensionality, but are blind to the spatial structure of the connectomes.
We propose a regularization framework where the $6$-D structure of the functional connectome (defined by pairs of points in $3$-D space) is explicitly taken into account via the fused Lasso or the \mbox{GraphNet} regularizer.
Our method only restricts the loss function to be convex and margin-based, allowing non-differentiable loss functions such as the hinge-loss to be used.
Using the fused Lasso or GraphNet regularizer with the hinge-loss leads to a structured sparse support vector machine (SVM) with embedded feature selection.
We introduce a novel efficient optimization algorithm based on the augmented Lagrangian and the classical alternating direction method, which can solve both fused Lasso and GraphNet regularized SVM with very little modification.
We also demonstrate that the inner subproblems of the algorithm can be solved efficiently in analytic form by coupling the variable splitting strategy with a data augmentation scheme.
Experiments on simulated data and resting state scans from a large schizophrenia dataset show that our proposed approach can identify predictive regions that are spatially contiguous in the $6$-D ``connectome space,'' offering an additional layer of interpretability that could provide new insights about various disease processes.
\end{abstract}

	\section{Introduction}
	\label{intro}
	There is substantial interest in establishing neuroimaging-based biomarkers that reliably distinguish individuals with psychiatric disorders from healthy individuals. 
Towards this end, neuroimaging affords a variety of specific modalities including structural imaging, diffusion tensor imaging (DTI) and tractography, and activation studies under conditions of cognitive challenge (\ie, task-based functional magnetic resonance imaging (fMRI)).
In addition, resting state fMRI has emerged as a mainstream approach that offers robust, sharable, and scalable ability to comprehensively characterize patterns of connections and network architecture of the brain.

Recently a number of groups have demonstrated that substantial quantities of discriminative information regarding psychiatric diseases reside in resting state functional connectomes \citep{Fox:2010,Castellanos:2013}.
In this article, we define the functional connectomes as the cross-correlation matrix that results from parcellating the brain into hundreds of distinct regions, and computing cross-correlation matrices across time \citep{Varoquaux:2013}.
Even with relatively coarse parcellation schemes with several hundred regions of interest (ROI), the resulting connectomes encompass hundreds of thousands of connections or more. 
The massive size of connectomes offers new possibilities, as patterns of connectivity across the entirety of the brain are represented.  
Nonetheless, the high dimensionality of connectomic data presents critical statistical and computational challenges.
In particular, mass univariate strategies that perform separate statistical tests at each edge of the connectome require excessively stringent corrections for multiple comparisons. 
Multivariate methods are promising, but these require specialized approaches in the context where the number of parameters dominate the number of observations, a setting commonly referred to as the ``large $p$ small $n$ problem,'' denoted $p\gg n$ \citep{Buhlmann:2011, West:2003}.

In the $p\gg n$ regime, it is important to leverage any potential structure in the data, and sparsity is a natural assumption that arises in many applications \citep{Candes:2008,Fan:2010}.
For example, in the context of connectomics, it is reasonable to believe that only a fraction of the functional connectome is impacted under a specific disorder, an assumption that has been supported in nearly all extant studies (see \cite{Castellanos:2013}).
Furthermore, when sparsity is coupled with a linear classifier, the nonzero variables can be interpreted as pairs of brain regions that allow reliable discrimination between controls and patients. 
In other words, sparse linear classifiers have the potential of revealing \emph{connectivity-based biomarkers} that characterize mechanisms of the disease process of interest \citep{Atluri:2013}. 

The problem of identifying the subset of variables relevant for prediction is called feature selection \citep{Jain:2000, Guyon:2003}, which can be done in a univariate or a multivariate fashion.
In the univariate approach, features are independentally ranked based on their statistical relationship with the target label (\eg, two sample t-test, mutual information), and only the top features are submitted to the classifier.
While this method is commonly used \citep{Zeng:2012, Sripada:2013}, it ignores the multivariate nature of fMRI.
On the other hand, multivariate approaches such as \emph{recursive feature elimination} \citep{Guyon:2003} can be used to capture feature interactions \citep{Craddock:2009,Dai:2012}, but these methods are computationally intensive and rely on suboptimal heuristics.
However, a more serious shortcoming common to all the methods above is that outside of sparsity, no structural information is taken into account.
In particular, we further know that functional connectomes reside in a structured space, defined by pairs of coordinate points in $3$-D brain space.
Performing prediction and feature selection in a spatially informed manner could potentially allow us to draw more neuroscientifically meaningful conclusions.
Fortunately, \emph{regularization methods} allow us to achieve this in a natural and principled way.

Regularization is a classical technique to prevent overfitting \citep{Tikhonov:1963,Stein:1961}, achieved by encoding prior knowledge about the data structure into the estimation problem.
Sparsity promoting regularization methods, such as Lasso \citep{Tibshirani:1996} and Elastic-net \citep{Zou:2005}, have the advantage of performing prediction and feature selection jointly \citep{Grosenick:2008, Yamashita:2008}; however, they also have the issue of neglecting additional structure the data may have.  Recently, there has been strong interest in the machine learning community in designing a convex regularizer that promotes \emph{structured sparsity} \citep{Mairal:2011,Chen:2012,Micchelli:2013}, which extends the standard concept of sparsity. 
Indeed, spatially informed regularizers have been applied successfully in task-based detection, \ie, \emph{decoding}, where the goal is to localize in $3$-D space the brain regions that become active under an external stimulus \citep{Baldassarre:2012,Michel:2011,Jenatton:2012,Grosenick:2013, Gramfort:2013}. 
Connectomic maps exhibit rich spatial structure, as each connection comes from a pair of localized regions in $3$-D space, giving each connection a localization in $6$-D space (referred to as ``connectome space'' hereafter).
However, to the best of our knowledge, no framework currently deployed exploits this spatial structure in the functional connectome. 

Based on these considerations, the main contributions of this paper are two-fold.
First, we propose to explicitly account for the $6$-D spatial structure of the functional connectome by using either the fused Lasso \citep{Tibshirani:2005} or the GraphNet regularizer \citep{Grosenick:2013}.
Second, we introduce a novel scalable algorithm based on the classical alternating direction method \citep{Gabay:1976,Glowinski:1975,Boyd:2011} for solving the nonsmooth, large-scale optimization problem that results from these spatially-informed regularizers.
Variable splitting and data augmentation strategies are used to break the problem into simpler subproblems that can be solved efficiently in closed form.
The method we propose only restricts the loss function to be convex and margin-based, which allows non-differentiable loss functions such as the hinge-loss to be used.  This is important, since using the fused Lasso or the GraphNet regularizer with the hinge-loss function leads to a structured sparse support vector machine (SVM)~\citep{Grosenick:2013,Gui-Bo-Ye:2011}, where feature selection is \emph{embedded} \citep{Guyon:2003}, \ie, feature selection is conducted jointly with classification.
We demonstrate that the optimization algorithm we introduce can solve both fused Lasso and GraphNet regularized SVM with very little modification.
To the best of our knowledge, this is the first application of structured sparse methods in the context of disease prediction using functional connectomes.
Additional discussions of technical contributions are reported in Sec.~\ref{subsec:optim}.
We perform experiments on simulated connectomic data and resting state scans from a large schizophrenia dataset to demonstrate that the proposed method identifies predictive regions that are spatially contiguous in the connectome space, offering an additional layer of interpretability that could provide new insights about various disease processes.

\paragraph{Notation}
We let lowercase and uppercase bold letters denote vectors and matrices, respectively.  
For every positive integer $n\in\mathbb{N}$, we define an index set $[n]:=\{1,\dots,n\}$, and also let $\BI_n\in\reals^{n\times n}$ denote the identity matrix.
Given a matrix $\BA\in\reals^{n\times p}$, we let $\BA^T$ denote its matrix transpose, and $\BA^H$ denote its Hermitian transpose.
Given $\w,\Bv\in\reals^n$, we invoke the standard notation $\inprod{\w,\Bv}:=\sum_{i=1}^n w_iv_i$ to express the inner product in $\reals^n$.  
We also let $\norm{\w}_p=(\sum_{i=1}^n w_i^p)^{1/p}$ denote the $\ell_p$-norm of a vector, $p\geq 1$, with the absence of subscript indicating the standard Euclidean norm, $\norm{\cdot}=\norm{\cdot}_2$.

	\section{Material and methods}
	\label{methods}
\subsection{Defining Functional Connectomes}\label{subsec:FC,def}
In this work, we produced a whole-brain resting state functional connectome as follows.
First, $347$ non-overlapping spherical nodes are placed throughout the entire brain in a regularly-spaced grid pattern, with a spacing of $18\times 18 \times 18$ mm; each of these nodes represents a pseudo-spherical ROI with a radius of $7.5$ mm, which encompasses $33$ voxels (the voxel size is $3\times 3\times 3$ mm).
For a schematic representation of the parcellation scheme, see Fig.~\ref{fig:roi,grid,slice}.
Next, for each of these nodes, a single representative time-series is assigned by spatially averaging the BOLD signals falling within the ROI.
Then, a cross-correlation matrix is generated by computing Pearson's correlation coefficient between these representative time-series.
Finally, a vector~\x of length $\binom{347}{2}=60,031$ is obtained by extracting the lower-triangular portion of the cross-correlation matrix.
This vector $\x\in\reals^{60,031}$ represents the whole-brain functional connectome, which serves as the feature vector for disease prediction.

The grid-based scheme for brain parcellation used in this work provides numerous advantages. 
Of note, this approach has been validated in previous studies \citep{Sripada:2013,Sripada:2013b,Sripada:2014}. 
Furthermore, the uniformly spaced grid is a good fit with our implementation of fused Lasso and GraphNet, as it provides a natural notion of nearest-neighbor and ordering among the coordinates of the connectome. 
This property also turns out to be critical for employing our optimization algorithm, which will be discussed in Sec.~\ref{subsec:optim}. 
This is in contrast to alternative approaches, such as methods that rely on anatomical \citep{AAL:2002,Zeng:2012} or functional parcellation schemes \citep{Dosenbach:2010}. 
Anatomical parcellations in particular have been shown to yield inferior performance to alternative schemes in the literature \citep{Power:2011}. 
Additionally, grid-based approaches provide scalable density: there is a natural way to increase the spatial resolution of the grid when computational feasibility allows. 
In particular, to increase node density, one could reduce the inter-node distance and also reduce the node size such that suitable inter-node space remains. 
This scalable density property turns out to be quite important, as our grid-based scheme is considerably more dense than standard functional parcellations (\eg, \cite{Dosenbach:2010, Shirer:2011}) that use as many as several hundred fewer nodes, and thus have tens of thousands fewer connections in the connectome.  
Finally, the use of our grid-based scheme naturally leaves space between the nodes. 
While on the surface this may appear to yield incomplete coverage, this is in fact a desirable property to avoid inappropriate inter-node smoothing. 
This may result as a function of either the point-spread process of fMRI image acquisition or be introduced as a standard preprocessing step. 
In recognition of these advantages, we have elected to use a grid scheme composed of pseudo-spherical nodes spaced at regular intervals.

One pragmatic advantage of using an \apriori parcellation scheme as opposed to one that combines parcellation and connectome calculation is that it permits the usage of a grid, and thus yields all the advantages outlined above. 
Moreover, it allows for easier comparison across studies since an identical (or at least similar) parcellation can be brought to bear on a variety of connectomic investigations. 
Secondly, while an approach that embeds both parcellation and connectome calculation in a single step may be suitable for recovering a more informative normative connectome, it would not necessarily be appropriate for recovering discriminative information about diseases in the connectome unless features were selected based on their disease-versus-healthy discriminative value. 
This approach, however, would require nesting parcellation within cross validation and would lead to highly dissimilar classification problems across cross validation folds and present challenges to any sort of inference or aggregation of performance. 
In light of these challenges, we have elected to use our \apriori grid-based scheme.

\renewcommand{\imwidth}  {0.2425\linewidth}
\renewcommand{\imwidthh}  {0.19\linewidth}
\begin{figure}[t!]
	\centering
	\textbf{\large{Grid-based Brain Parcellation Scheme with $\bmath{347}$-nodes}} \vspace{8pt}\\
	\begin{subfigure}[t]{\imwidth}
		\centering
		\includegraphics[height=100pt]{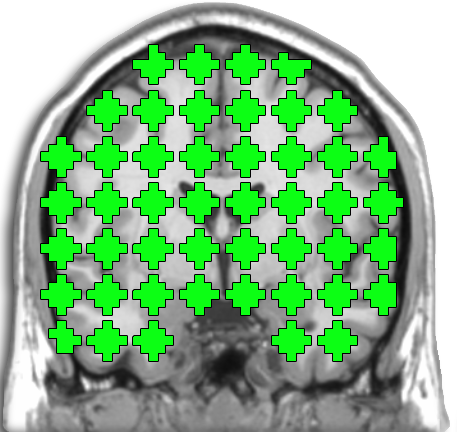}
		\caption{Coronal}
	\end{subfigure}\hspace{12pt}
	\begin{subfigure}[t]{\imwidth}
		\centering
		\includegraphics[height=100pt]{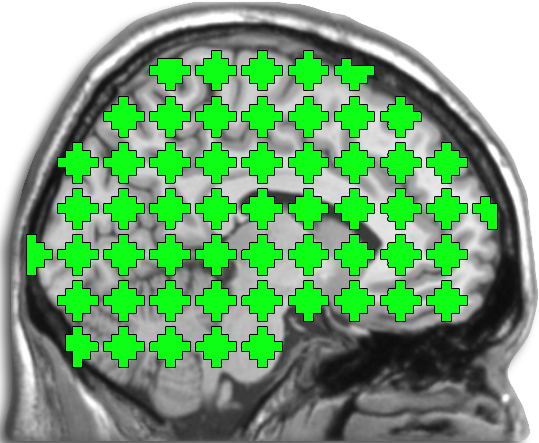}
		\caption{Sagittal}
	\end{subfigure}\hspace{12pt}
	\begin{subfigure}[t]{\imwidth}
		\centering
		\includegraphics[height=100pt]{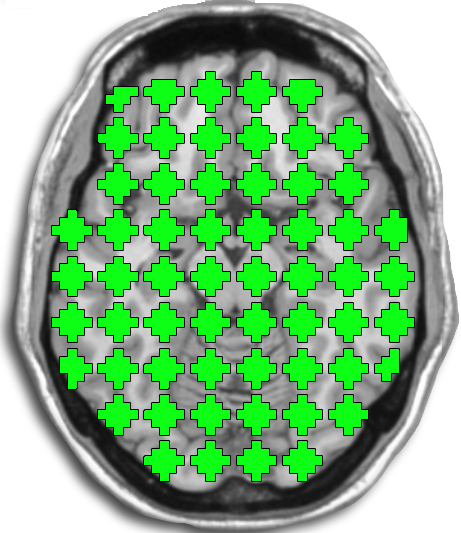}
		\caption{Axial}
	\end{subfigure}
	\begin{subfigure}[t]{\imwidthh}
		\centering
		\raisebox{15pt}{\includegraphics[height=60pt]{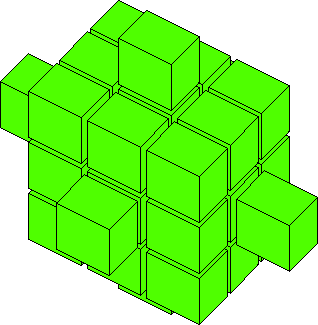}}
		\caption{Node (33 voxels)}
	\end{subfigure}
	\caption{
	Coronal, sagittal, and axial slices depicting the coverage of our brain parcellation scheme along with $3$-D rendering of one pseudo-sphereical node. 
	Each contiguous green region represents a pseudo-spherical node representing an ROI containing $33$-voxels.
	Overall, there are $347$ non-overlapping nodes placed throughout the entire brain.
	These nodes are placed on a grid with $18$ mm spacing between node centers in the $X$, $Y$, and $Z$ dimensions.
	}
	\label{fig:roi,grid,slice}
\end{figure}

\subsection{Statistical learning framework}
We now formally introduce the statistical learning framework adopted to perform joint feature selection and disease prediction with spatial information taken into consideration.

\subsubsection{Regularized empirical risk minimization and feature selection}
In this work, we are interested in the supervised learning problem of linear binary classification.
Suppose we are given a set of training data \sloppy{$\left\{(\x_1,y_1),\cdots,(\x_n,y_n)\right\}$}, where $\x_i\in\reals^p$ is the input feature vector and ${y_i\in\{-1,+1\}}$ is the corresponding class label for each $i\in[n]$.
In our application, $\x_i$ represents functional connectome and $y_i$ indicates the diagnostic status of subject $i\in[n]$, where we adopt the convention of letting $y=+1$ indicate ``disorder'' and $y=-1$ indicate ``healthy'' in this article.
The goal is to learn a linear decision function $\sign{\inprod{\x,\w}}$, parameterized by weight vector $\w\in\reals^p$, that predicts the label $y\in\{-1,+1\}$ of a new input $\x\in\reals^p$.
A standard approach for estimating \w is solving a regularized empirical risk minimization (ERM) problem with the form 
\begin{equation}
	\argmin{\w\in\reals^p} \frac{1}{n} \sum_{i=1}^n \loss\left(y_i \inprod{\w,\x_i}\right) + \lambda\Reg(\w) \;.
	\label{eqn:reg,erm}
\end{equation}
The first term $\frac{1}{n}\sum_{i=1}^n \loss\left(y_i \inprod{\w,\x_i}\right)$ corresponds to the \emph{empirical risk} of a margin-based loss function $\loss:\reals\to\reals_+$ (\eg, hinge, logistic, exponential), which quantifies how well the model fits the data.
The second term $\Reg:\reals^p\to\reals_+$ is a \emph{regularizer} that curtails overfitting and enforces some kind of structure on the solution by penalizing weight vectors that deviate from the assumed structure.
The user-defined regularization parameter $\lambda\geq0$ controls the tradeoff between data fit and regularization. 
Throughout this work, we assume the loss function and the regularizer to be convex, but not necessarily differentiable.
Furthermore, we introduce the following notations 
\[
	\begin{array}{ccc}
		\Y:=\diag{y_1,\cdots,y_n}			, &
		\X:=\ba{c} \x_1^T \\ \vdots \\ \x_n^T \ea , &
		\YXw=\ba{c} y_1 \inprod{\w,\x_1} \\ \vdots \\ y_n \inprod{\w,\x_n} \ea ,
	\end{array}
\]
which allow us to express the empirical risk succinctly by defining a functional \sloppy{${\Loss:\reals^n\to\reals_+}\;$} which aggregates the total loss ${\Loss(\YXw):=\sum_{i=1}^n \loss(y_i \inprod{\w,\x_i})}\,$.

Regularized ERM \eqref{eqn:reg,erm} has a rich history in statistics and machine learning, and many well known estimators can be recovered from this framework.
For example, when the hinge loss ${\loss(t):=\text{max}(0,1-t)}$ is used with the smoothness promoting \elltwo-regularizer $\norm{\w}^2_2$, we recover the SVM \citep{Cortes:1995}.
However, while smoothness helps prevent overfitting, it is problematic for model interpretation, as all the coefficients from the weight vector contribute to the final prediction function.
Automatic feature selection can be done using the \ellone-regularizer $\norm{\w}_1$ known as the Lasso \citep{Tibshirani:1996}, which causes many of the coefficients in \w to be exactly zero.
Because the prediction function is described by a linear combination between the weight \w and the feature vector \x, we can directly identify and visualize the regions that are relevant for prediction.

While the \ellone-regularizer possesses many useful statistical properties, several works have reported poor performance when the features are highly correlated.
More precisely, if there are clusters of correlated features, Lasso will select only a single representative feature from each cluster group, ignoring all the other equally predictive features.
This leads to a model that is overly sparse and sensitive to data resampling, creating problems for interpretation.
To address this issue, \cite{Zou:2005} proposed to combine the \ellone and \elltwo regularizers, leading to the Elastic-net, which has the form $\norm{\w}_1+\frac{\gamma}{2\lambda}\norm{\w}^2_2$, where $\gamma\geq 0$ is a second regularization parameter. 
The \ellone-regularizer has the role of encouraging sparsity, whereas the \elltwo-regularizer has the effect of allowing groups of highly correlated features to enter the model together, leading to a more stable and arguably a more sensible solution.  
While Elastic-net addresses part of the limitations of Lasso and has been demonstrated to improve prediction accuracy \citep{Carroll:2009,Ryali:2010}, it does not leverage the $6$-D structure of connectome space.  To address this issue, we employ the fused Lasso and GraphNet \citep{Grosenick:2013}.

\subsubsection{Spatially informed feature selection and classification via fused Lasso and GraphNet}
The original formulation of fused Lasso \citep{Tibshirani:2005} was designed for encoding correlations among successive variables in $1$-D data, such as mass spectrometry and comparative genomic hybridization (CGH) data \citep{Gui-Bo-Ye:2011}.
More specifically, assuming the weight vector $\w\in\reals^p$ has a natural ordering among its coordinates $j\in[p]$, the regularized ERM problem with the fused Lasso has the following form:
\begin{equation}
	\argmin{\w\in\reals^p}\frac{1}{n}\Loss(\YXw) + \lambda\norm{\Bw}_1 + 
		\gamma\sum_{j=2}^p\abs{w^{(j)} - w^{(j-1)}} \;,
	\label{eqn:fused,lasso,1d}
\end{equation}
where $w^{(j)}$ indicates the $j$-th entry of \w.
Like Elastic-net, this regularizer has two components: the first component is the usual sparsity promoting \ellone-regularizer, and the second component penalizes the absolute deviation among adjacent coordinates.
Together, they have the net effect of promoting sparse and piecewise constant solutions.

The idea of penalizing the deviations among neighboring coefficients can be extended to other situations where there is a natural ordering among the feature coordinates.
For instance, the extension of the $1$-D fused Lasso \eqref{eqn:fused,lasso,1d} for $2$-D imaging data is to penalize the \emph{vertical} and \emph{horizontal} difference between pixels; here, the coordinates are described via lexicographical ordering.
This type of generalization applies to our \mbox{$6$-D} functional connectomes by the virtue of the grid pattern in the nodes, and the ERM formulation reads
\begin{equation}
	\argmin{\w\in\reals^p}\frac{1}{n}\Loss(\YXw) + \lambda\norm{\Bw}_1 + 
		\gamma\sum_{j=1}^p\sum_{k\in\mathcal{N}_j} \abs{\;w^{(j)}-w^{(k)}} \;,
	\label{eqn:fused,lasso,6d}
\end{equation}
where $\mathcal{N}_j$ is the first-order neighborhood set corresponding to coordinate $j$ in \mbox{$6$-D} connectome space.
The spatial penalty $\gamma\sum_{j=1}^p\sum_{k\in\mathcal{N}_j} \abs{\,w^{(j)}-w^{(k)}}$ accounts for the \mbox{$6$-D} structure in the connectome by penalizing deviations among \emph{nearest-neighbor} edges, encouraging solutions that are spatially coherent in the connectome space.
This type of regularizer is known as an anisotropic total variation (TV) penalty in the image processing community \citep{Wang:2008tv}, and an analogous isotropic TV penalty was applied by \cite{Michel:2011} for the application of $3$-D brain decoding.

When the absolute value penalty in the spatial regularizer ${|\,w^{(j)}-w^{(k)}|}$ in \eqref{eqn:fused,lasso,6d} is replaced by the squared penalty $\frac{1}{2}(w^{(j)}-w^{(k)})^2$, we recover the GraphNet model proposed by \mbox{\cite{Grosenick:2013}}:
\begin{equation}
	\argmin{\w\in\reals^p}\frac{1}{n}\Loss(\YXw) + \lambda\norm{\Bw}_1 + 
		\frac{\gamma}{2}\sum_{j=1}^p\sum_{k\in\mathcal{N}_j} \left(w^{(j)}-w^{(k)}\right)^2 \;.
	\label{eqn:graphnet}
\end{equation}
GraphNet also promotes spatial contiguity, but instead of promoting sharp piecewise constant patches, it encourages the clusters to appear in smoother form by penalizing the quadratic deviations among the nearest-neighbor edges (\ie, the coordinates of the functional connectome \x).
We emphasize that the optimization algorithm we propose can be used to solve both fused Lasso~\eqref{eqn:fused,lasso,6d} and GraphNet~\eqref{eqn:graphnet} with very little modification.

To gain a better understanding of the neighborhood set $\mathcal{N}_j$ in the context of our application, let us denote $(x,y,z)$ and $(x',y',z')$ the pair of $3$-D points in the brain that define the connectome coordinate $j$.
Then, the first-order neighborhood set of $j$ can be written precisely as
\footnote{If $(x,y,z)$ or $(x',y',z')$ are on the boundary of the brain volume, then neighboring points outside the brain volume are excluded from $\mathcal{N}_j$.}
\begin{equation}
	\mathcal{N}_j:=
	\braces{
	\begin{array}{l}
			\big(x\pm 1,y,z,x',y',z'\big), \;
			\big(x,y\pm 1,z,x',y',z'\big), \;
			\big(x,y,z\pm 1,x',y',z'\big),\\
			\big(x,y,z,x'\pm 1,y',z'\big), \;
			\big(x,y,z,x',y'\pm 1,z'\big), \;
			\big(x,y,z,x',y',z'\pm 1\big)
	\end{array}}\; .
	\nonumber
\end{equation}
Fig.~\ref{fig,conn,neighbor} provides a pictorial illustration of $\mathcal{N}_j$ in the case of a $4$-D connectome, where the nodes reside in $2$-D space.

There are multiple reasons why fused Lasso and GraphNet are justified approaches for our problem.
For example, fMRI is known to possess high spatio-temporal correlation between neighboring voxels and time points, partly for biological reasons as well as from preprocessing (\eg, spatial smoothing).
Consequently, functional connectomes contain rich correlations among nearby coordinates in the connectome space.
In addition, there is a neurophysiological basis for why the predictive features are expected to be spatially contiguous rather than being randomly dispersed throughout the brain; this point will be  thoroughly discussed in Sec.~\ref{subsec:why,flasso}.
Finally, the spatial coherence that fused Lasso and GraphNet promote helps decrease model complexity and facilitates interpretation.

Letting $\C\in\reals^{e\times p}$ denote the $6$-D \emph{finite differencing matrix} (also known as the \emph{incidence matrix}), the spatial regularization term for both fused Lasso and GraphNet can be written compactly as
\begin{equation}
	\norm{\C\w}_q^q=\sum_{j=1}^p\sum_{k\in\mathcal{N}_j} |\,w^{(j)}-w^{(k)}|^q,\;\;\; q\in\{1,2\} \;,
	\nonumber
\end{equation}
where each row in \C contains a single $+1$ and a $-1$ entry, and $e$ represents the total number of adjacent coordinates in the connectome.
This allows us to write out the regularized ERM formulation for both fused Lasso~\eqref{eqn:fused,lasso,6d} and GraphNet~\eqref{eqn:graphnet} in the following unified form: 
\begin{equation}
	\argmin{\w\in\reals^p} \frac{1}{n}\Loss(\YXw) + \lambda\norm{\w}_1 + \frac{\gamma}{q}\norm{\C\w}^q_q  \; ,\; q\in\{1,2\}\;.
	\label{eqn:costfx}
\end{equation}
We will focus on this matrix-vector representation hereafter, as it is more intuitive and convenient for analyzing the variable splitting framework in the upcoming section.

\begin{figure}
	{\begin{center}
	\resizebox{0.39\linewidth}{!}{\begin{tikzpicture}[-,red, line width=0.1cm,font=\bfseries,draw=black] 
	\tikzstyle{every node}=[circle,ultra thick,draw=black,fill=white,text=black,minimum size=0cm,line width=0.04cm,inner sep=1pt]
	\tikzstyle{main} =[fill=red]
	\tikzstyle{main1} =[fill=blue!30]
	\tikzstyle{main2} =[fill=green!50]
	\matrix [rectangle,row sep=4pt, column sep=8pt,draw=black,fill=blue!5,line width=0.01cm] %
	{
	\node  (1_1){1,1} ;& 	\node  (1_2) {2,1} ;&	\node  (1_3) {3,1} ;&	\node  (1_4) {4,1} ;&	\node  (1_5) {5,1} ;&	\node  (MAIN2T)[main2] {6,1}  ;&	\node  (1_7) {7,1} ;\\
	\node  (2_1){1,2} ;& 	\node  (2_2) {2,2} ;&	\node  (2_3) {3,2} ;&	\node  (2_4) {4,2} ;&	\node  (MAIN2L)[main2] {5,2} ;&	\node  (MAIN2)[main] {6,2} ;&	\node  (MAIN2R)[main2] {7,2} ;\\
	\node  (3_1){1,3} ;& 	\node  (MAIN1T)[main1] {2,3} ;&	\node  (3_3) {3,3} ;&	\node  (3_4) {4,3} ;&	\node  (3_5) {5,3} ;&	\node  (MAIN2B)[main2] {6,3} ;&	\node  (3_7) {7,3} ;\\
	\node(MAIN1L)[main1]  {1,4} ;& 	\node  (MAIN1)[main]{2,4} ;&	\node  (MAIN1R)[main1] {3,4} ;&	\node  (4_4) {4,4} ;&	\node  (4_5) {5,4} ;&	\node  (4_6) {6,4} ;&	\node  (4_7) {7,4} ;\\
	\node  (5_1){1,5} ;& 	\node  (MAIN1B)[main1] {2,5} ;&	\node  (5_3) {3,5} ;&	\node  (5_4) {4,5} ;&	\node  (5_5) {5,5} ;&	\node  (5_6) {6,5} ;&	\node  (5_7) {7,5} ;\\
};
	\draw (MAIN1) to (MAIN2) [red];

	\draw[green] (MAIN1) to [out=90,in=180] (MAIN2L) ;
	\draw[green] (MAIN1) to [out=90,in=180] (MAIN2T) ;
	\draw[green] (MAIN1) to [out=-15,in=-115] (MAIN2R) ;
	\draw[green] (MAIN1) to [out=-15,in=-135](MAIN2B) ;
	
	\draw[blue] (MAIN2) to  [out=155,in=90] (MAIN1L) ;
	\draw[blue] (MAIN2) to  [out=155,in=75](MAIN1T) ;
	\draw[blue] (MAIN2) to [out=-125,in=10](MAIN1R) ;
	\draw[blue] (MAIN2) to [out=-125,in=0](MAIN1B) ;
\end{tikzpicture}}\\
	\textbf{{$\mathcal{N}_j$ in $\bmath{4}$-D Connectome Space}} \vspace{-15pt}\\
	\end{center}	
	}
	\caption{
	Illustration of the neighborhood structure of the connectome when the nodes reside in $2$-D space.  
	The red edge represents coordinate $j=\big\{(2,4),(6,2)\big\}$ in $4$-D connectome space, and its neighborhood set $\mathcal{N}_j$ is represented by the blue and green edges.  
	This idea extends directly to $6$-D connectomes generated from $3$-D resting state volumes.
	}
	\label{fig,conn,neighbor}
\end{figure}

\subsection{Optimization}\label{subsec:optim}
Solving the optimization problem \eqref{eqn:costfx} is challenging since the problem size $p$ is large and the three terms {in the cost function} can each be \mbox{non-differentiable}.
To address these challenges, we now introduce a scalable optimization framework based on augmented Lagrangian (AL) methods.
In particular, we introduce a variable splitting scheme that converts the unconstrained optimization problem of the form \eqref{eqn:costfx} into an equivalent constrained optimization problem, which can be solved efficiently using the alternating direction method of multipliers (ADMM) algorithm \citep{Boyd:2011,Glowinski:1975, Gabay:1976}. 
We demonstrate that by augmenting the weight vector with zero entries at appropriate locations, the inner subproblems associated with ADMM can be solved efficiently in closed form.

\subsubsection{Alternating Direction Method of Multipliers}
The ADMM algorithm is a powerful algorithm for solving convex optimization problems having the separable structure \citep{Boyd:2011}
\begin{equation}
	\minimize{\xbar,\ybar}\,\fbar(\xbar) + \gbar(\ybar) \;\;\;\; 
	\text{subject to } \Abar\xbar+\Bbar\ybar=\bzero \;,
	\label{eqn:canonical,admm}
\end{equation}
where $\xbar\in\reals^\pbar$ and $\ybar\in\reals^\qbar$ are unknown primal variables, $\fbar:\reals^\pbar\to\reals\cup\{+\infty\}$ and $\gbar:\reals^\qbar\to\reals\cup\{+\infty\}$ are closed convex functions, and $\Abar\in\reals^{c\times \pbar}$ and $\Bbar\in\reals^{c\times \qbar}$ are matrices representing $c$ linear constraints.  
More specifically, the ADMM algorithm solves for the primal variables in \eqref{eqn:canonical,admm} through the following iterative procedure:
\begin{align}
	&\hspace{0.17\linewidth}\xbar\iterp\leftarrow\displaystyle\argmin{\xbar} \fbar(\xbar) + \frac{\rho}{2}\norm{\Abar\xbar + \Bbar\ybar\iter + \u\iter}^2 \label{eqn:admm,xbar,update}\\ 
	&\hspace{0.17\linewidth}\ybar\iterp\leftarrow\displaystyle\argmin{\ybar} \gbar(\ybar) + \frac{\rho}{2}\norm{\Abar\xbar\iterp + \Bbar\ybar + \u\iter}^2 \label{eqn:admm,ybar,update}\\ 
	&\hspace{0.17\linewidth}\u\iterp\leftarrow \u\iter+ \left(\Abar\xbar\iterp+\Bbar\ybar\iterp\right)  \; , & \label{eqn:admm,dual,update}
\end{align}
where superscript $t$ denotes the iteration count and $\u\in\reals^c$ denotes the (scaled) dual variable.

The convergence of the ADMM algorithm has been established in Theorem $1$ of \cite{Mota:2011}, which states that if matrices \Abar and \Bbar are full column-rank and the problem \eqref{eqn:canonical,admm} is solvable (\ie, it has an optimal objective value), the ADMM iterations \eqref{eqn:admm,xbar,update} - \eqref{eqn:admm,dual,update} converges to the optimal solution.
While the AL parameter $\rho>0$ does not affect the convergence property of ADMM, it can impact its convergence speed.
We use the value $\rho=1$ in all of our implementations, although this value can be empirically tuned in practice.

\subsubsection{Variable splitting and data augmentation}
\label{subsec:var,split}
The original formulation of our problem \eqref{eqn:costfx} does not have the structure of \eqref{eqn:canonical,admm}.
However, we can convert the unconstrained optimization problem \eqref{eqn:costfx} into an equivalent constrained optimization problem \eqref{eqn:canonical,admm} by introducing auxiliary constraint variables, a method known as \emph{variable splitting} \citep{Afonso:2010}.  
While there are several different ways to introduce the constraint variables, the heart of the strategy is to select a splitting scheme that decouples the problem into more manageable subproblems.
For example, one particular splitting strategy we can adopt for problem \eqref{eqn:costfx} is
\begin{equation}
	\begin{array}{c}
		\dstyle\minimize{\substack{\w,\va \\ \vb,\vc,\vd}} 
			\frac{1}{n}\Loss(\va)+\lambda\norm{\vb}_1+\frac{\gamma}{q}\norm{\vc}^q_q \; \; 
			\vspace{0.01\linewidth}\\
		\text{ subject to }  \YXw =\va, \; \w=\vb, \; \C\vd=\vc, \;  \w=\vd \;, 
	\end{array}
	\label{eqn:splitting1}
\end{equation}
where $\va,\vb,\vc,\vd$ are the constraint variables.
It is easy to see that problems \eqref{eqn:costfx} and \eqref{eqn:splitting1} are equivalent, and the correspondence with the ADMM formulation \eqref{eqn:canonical,admm} is as follows:
\begin{equation}
	\begin{array}{c}
		\fbar(\xbar)=\dstyle\frac{\gamma}{q}\norm{\vc}^q_q , \;\;\;
		\gbar(\ybar)=\dstyle\frac{1}{n}\Loss(\va) + \lambda\norm{\vb}_1  
		\vspace{0.02\linewidth} \\
		\Abar = 	\ba{cc} 	\Y\X	& \bzero \\ \I	& \bzero \\ \bzero 	& \I \\ \I & \bzero \ea 	, \;
		\xbar = \bmat \w \\ \vc \emat, 
		\Bbar = 	\ba{rrr} 	-\I 	& \bzero	& \bzero \\ \bzero & -\I & \bzero  \\ 
						 \bzero & \bzero & -\C \\ \bzero & \bzero & -\I \ea,
		\ybar = \bmat \va \\ \vb \\ \vd \emat 	.
	\end{array}
	\label{eqn:admm,objective1}
\end{equation}
However, there is an issue with this splitting strategy: one of the resulting subproblems from the ADMM algorithm requires us to invert a matrix involving the Laplacian matrix $\C^T\C\in\reals^{p\times p}$, which is prohibitively large.
Although this matrix is sparse, it has a distorted structure due to the irregularities in the coordinates of \x.
These irregularities arise from two reasons:
(1) the nodes defining the functional connectome \x are placed only on the brain, not the entire rectangular field of view (FOV), and
(2) \x lacks a complete $6$-D representation since it only contains the lower-triangular part of the cross-correlation matrix.
Fig.~\ref{fig:lap,noncirc} displays the Laplacian matrix that results from the $347$-node functional connectome defined in Section~\ref{subsec:FC,def}, and the distorted structure is clearly visible.

To address this issue, we introduce an \emph{augmentation matrix} $\A\in\reals^{\ptil\times p}$, whose rows are either the zero vector or an element from the trivial basis $\braces{\bmath{e}_j \; |\; j\in[p]}$, and has the property $\A^T\A=\I_p$.
Furthermore, we define the \emph{augmented weight vector} $\wtil:=\A\w$, where \A rectifies the irregularities in the coordinates of \w (and \x) by padding extra zero entries, accommodating for:
(1) the nodes that were not placed in the FOV (\ie, the regions outside the brain), and
(2) the diagonal and upper-triangular part of the cross-correlation matrix, which were disposed due to redundancy; further details regarding this augmentation scheme is reported in \ref{appendix:data,aug}.
As a result, we now have a new differencing matrix $\Ctil\in\reals^{\tilde{e}\times\ptil}$ corresponding to $\wtil\in\reals^\ptil$, whose Laplacian matrix $\Ctil^T\Ctil\in\reals^{\ptil\times\ptil}$ has a systematic structure, as shown in Fig.~\ref{fig:lap,circ}.
In fact, this matrix has a special structure known as \emph{block-circulant with circulant-blocks} (BCCB), which is critical since the matrix inversion involving $\Ctil^T\Ctil$ can be computed efficiently in closed form using the fast Fourier transform (FFT) (the utility of this property will be elaborated more in Section~\ref{subsec:admm,steps}).  
It is important to note that this BCCB structure in the Laplacian matrix arises from the grid structure introduced from the parcellation scheme we adopted for producing the functional connectome.

\renewcommand{\imwidth}  {0.38\linewidth}
\begin{figure}
	\centering
	\begin{subfigure}[t]{\imwidth}
		 \centering
		 \includegraphics[height=.7\linewidth]{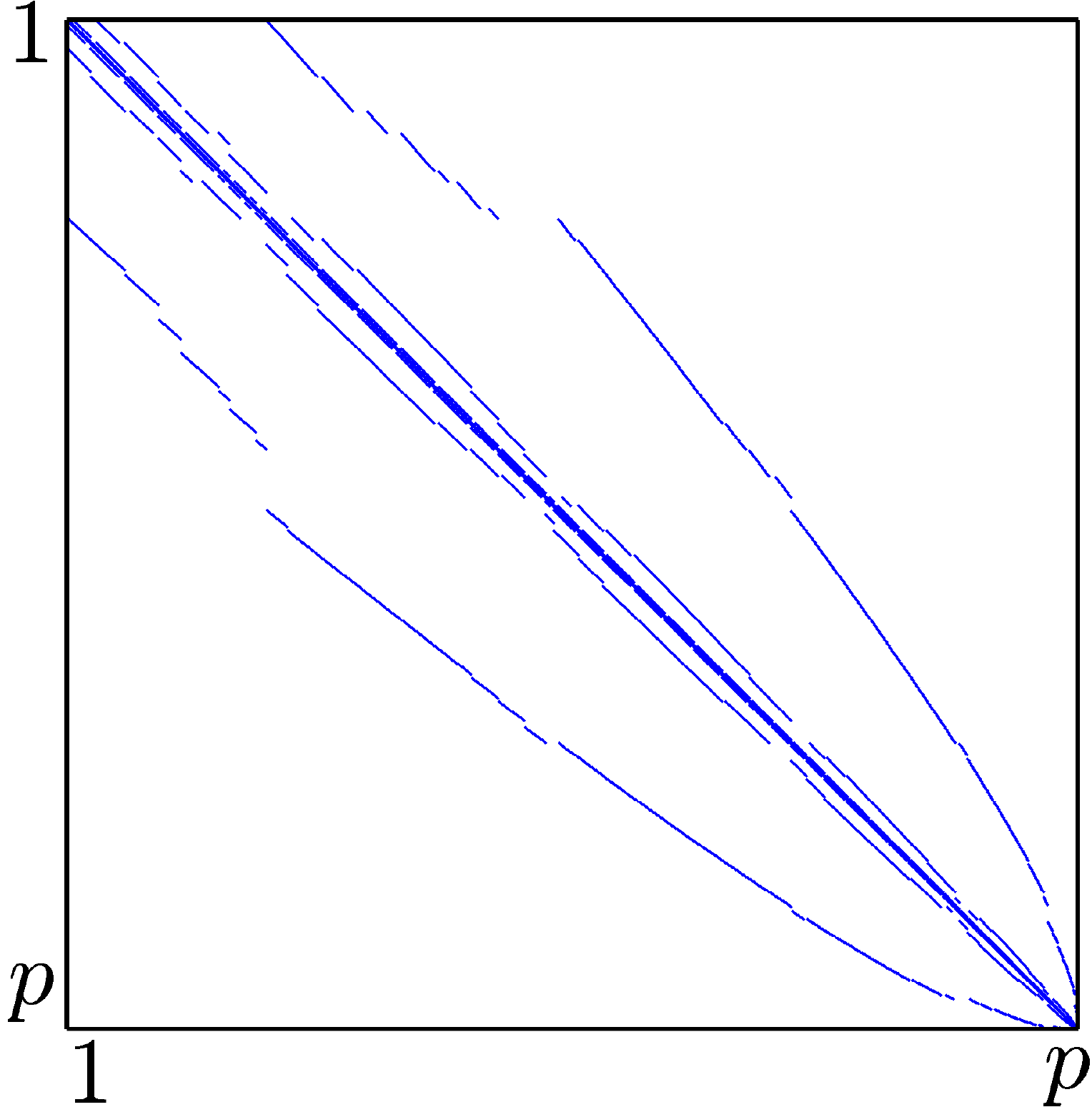}
		 \caption{Laplacian matrix: $\C^T\C$}
		 \label{fig:lap,noncirc}
	\end{subfigure}
	\hspace{0.05\linewidth}
	\begin{subfigure}[t]{\imwidth}
		 \centering
		 \includegraphics[height=.7\linewidth]{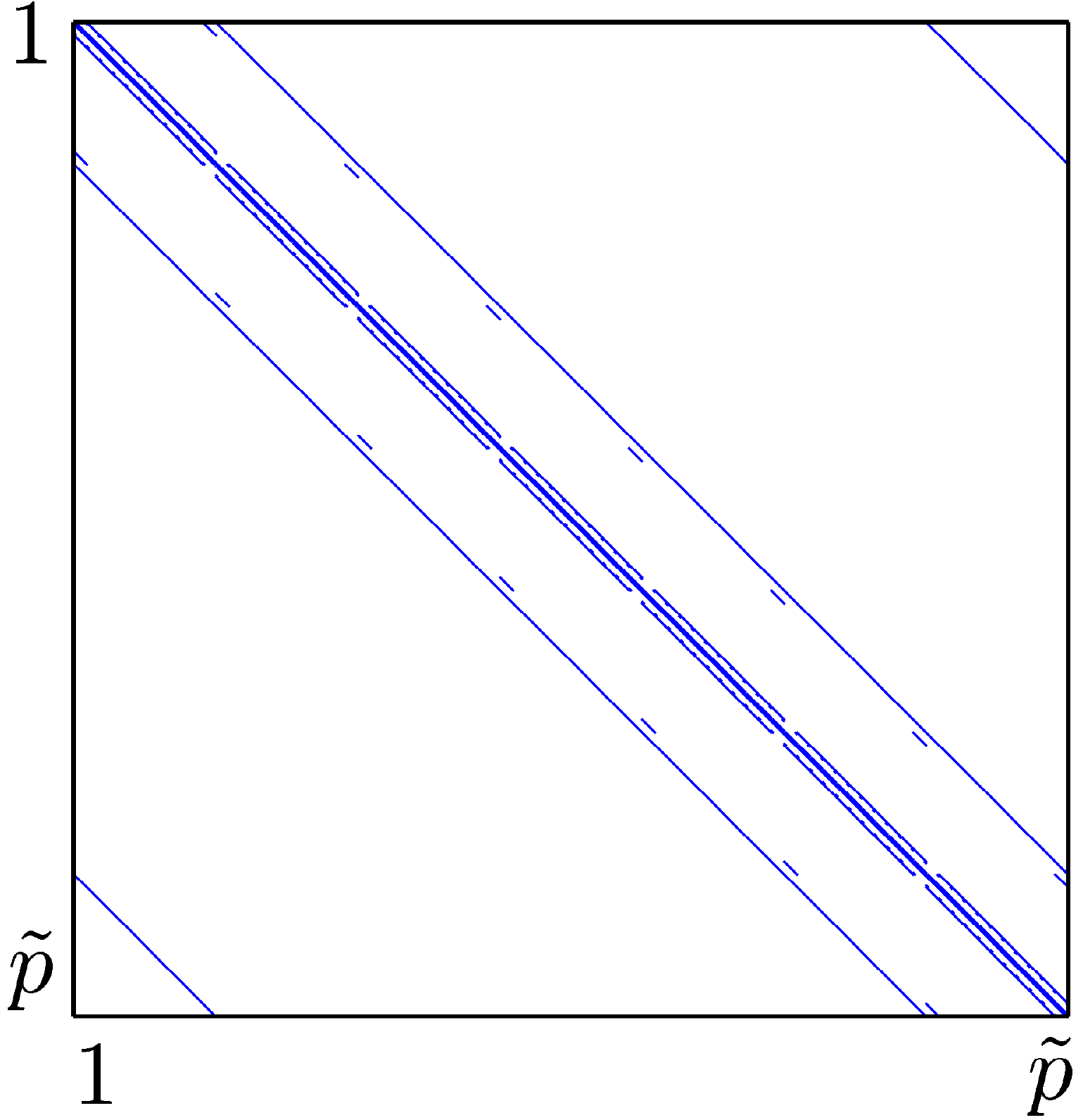}
		 \caption{Augmented Laplacian matrix: $\Ctil^T\Ctil$}
		 \label{fig:lap,circ}
	\end{subfigure}

	\caption{Laplacian matrix corresponding to the original data $\C^T\C$ and the augmented data $\Ctil^T\Ctil$, where the rows and columns of these matrices represent the coordinates of the original and augmented functional connectome.  Note that the irregularities in the original Laplacian matrix are rectified by data augmentation.  The augmented Laplacian matrix has a special structure known as \emph{block-circulant with circulant-blocks} (BCCB), which has important computational advantages that will be exploited in this work.}
	 \label{fig:lap}
\end{figure}

Finally, by introducing a diagonal masking matrix $\B\in\{0,1\}^{\etil\times\etil}$, we have $\|\B\Ctil\wtil\|^q_q = \norm{\C\w}_q^q$ for $\q\in\{1,2\}$.  
Note that this masking strategy was adopted from the recent works of \cite{Allison:2013} and \cite{Matakos:2013}, and has the effect of removing artifacts that are introduced from the data augmentation procedure when computing the $\norm{\cdot}_q^q$-norm.
This allows us to write out the fused Lasso and GraphNet problem \eqref{eqn:costfx} in the following equivalent form:
\begin{equation}
	\argmin{\w\in\reals^p} \frac{1}{n}\Loss(\YXw) + \lambda\norm{\w}_1 + 
		\frac{\gamma}{q}\norm{\B\Ctil\A\w}^q_q  \; ,\; q\in\{1,2\}
	\nonumber
\end{equation}
Moreover, this can be converted into a constrained optimization problem
\begin{equation}
	\begin{array}{c}
		\dstyle\minimize{\substack{\w,\va \\ \vb,\vc,\vd}} 
			\frac{1}{n}\Loss(\va)+\lambda\norm{\vb}_1+\frac{\gamma}{q}\norm{\B\vc}^q_q \; \; 
			\vspace{0.01\linewidth}\\
		\text{ subject to }  \YXw =\va, \; \w=\vb, \; \Ctil\vd=\vc, \;  \A\w=\vd \;, 
	\end{array}
	\label{eqn:admm,splitting2}
\end{equation}
and the correspondence with the ADMM formulation \eqref{eqn:canonical,admm} now becomes:
\begin{equation}
	\begin{array}{c}
		\fbar(\xbar)=\dstyle\frac{\gamma}{q}\norm{\B\vc}^q_q , \;\;\;
		\gbar(\ybar)=\dstyle\frac{1}{n}\Loss(\va) + \lambda\norm{\vb}_1  
		\vspace{0.02\linewidth} \\
		\Abar = 	\ba{cc} 	\Y\X	& \bzero \\ \I	& \bzero \\ \bzero 	& \I \\ \A & \bzero \ea 	, \;
		\xbar = \bmat \w \\ \vc \emat 			, 
		\Bbar = 	\ba{rrr} 	-\I 	& \bzero	& \bzero \\ \bzero & -\I & \bzero  \\ 
						 \bzero & \bzero & -\Ctil \\ \bzero & \bzero & -\I \ea 		, 
		\ybar = \bmat \va \\ \vb \\ \vd \emat \;.
	\end{array}
	\label{eqn:admm,objective2}
\end{equation}
The dual variables corresponding to $\va,\vb,\vc,$ and $\vd$ are written in block form $\u=[\ua^T,\ub^T,\uc^T,\ud^T]^T$.
Note that functions \fbar and \gbar are convex, and matrices \Abar and \Bbar are full column-rank, so the convergence of the ADMM iterations \eqref{eqn:admm,xbar,update}-\eqref{eqn:admm,dual,update} is guaranteed (see Theorem~1 in \cite{Mota:2011}).

\subsubsection{ADMM: efficient closed-form updates}
\label{subsec:admm,steps}

With the variable splitting scheme \eqref{eqn:admm,splitting2} and ADMM formulation \eqref{eqn:admm,objective2}, the ADMM update for the primal variable $\xbar$ \eqref{eqn:admm,xbar,update} decomposes into subproblems
\begin{alignat}{1} 
	\w\iterp 	\leftarrow \argmin{\w}&
		 \Bigg\{\normsq{\Y\X\w-\left(\va\iter-\ua\iter\right)} 
		+ \normsq{\w-\left(\vb\iter-\ub\iter\right)} \nonumber \\
		  &+ \normsq{\A\w-\left(\vd\iter-\ud\iter\right)} \Bigg\} \label{eqn:w,update1} \\
	\vc\iterp \leftarrow \argmin{\vc}&
		\Bigg\{\frac{\gamma}{q}\norm{\B\vc}_q^q
		  +\frac{\rho}{2}\normsq{\vc-\left(\Ctil\vd\iter-\uc\iter\right)}\Bigg\}\;, \label{eqn:v3,update1}
\end{alignat}
whereas the updates for primal variable $\ybar$ \eqref{eqn:admm,ybar,update} are
\begin{alignat}{1} 
	\va\iterp \leftarrow \argmin{\va}
		& \left\{ \frac{1}{n}\Loss(\va) + \frac{\rho}{2}\normsq{\va-\left(\Y\X\w\iterp+\ua\iter\right)} \right\}
			\label{eqn:v1,update1}\\
	\vb\iterp \leftarrow \argmin{\vb} 
		& \left\{\lambda\norm{\vb}_1 + \frac{\rho}{2}\normsq{   \vb-\left(\w\iterp+\ub\iter\right)} 
			\label{eqn:v2,update1}\right\}\\
	\vd\iterp \leftarrow \argmin{\vd}
		& \bigg\{\normsq{\Ctil\vd-\left(\vc\iterp+\uc\iter\right)}
		 + \normsq{\vd-\left(\A\w\iterp+\ud\iter\right)}\bigg\} \;.  \label{eqn:v4,update1}
\end{alignat}
The update for the dual variable \u is a trivial matrix-vector \mbox{multiplication \eqref{eqn:admm,dual,update}} (see Algorithm~\ref{alg:admm} line $14$-$17$).

We now demonstrate that the minimization problems \eqref{eqn:w,update1}-\eqref{eqn:v4,update1} each admits an efficient, closed form solution.

\newcommand{\Bvarrho}{\bmath{\varrho}}
\paragraph{\w update}
The quadratic minimization problem \eqref{eqn:w,update1} has the following closed form solution:
\begin{align}
	\w\iterp \leftarrow \left(\X^T\X + 2\BI_p\right)\inv  
		\Big(&\X^T\Y^T [\va\iter-\ua\iter] 
		+[\vb\iter-\ub\iter]  
		 +\A^T[\vd\iter-\ud\iter] \Big) \,. \label{eqn:w,update2}
\end{align}
Note we used the fact that $\Y^T\Y=\I_n$ and $\A^T\A=\I_p$ to arrive at this expression.
Applying update \eqref{eqn:w,update2} brute force will require an inversion of a $(p\times p)$ matrix, but this can be converted into an $(n\times n)$ inversion problem by invoking the \emph{matrix inversion Lemma}
\begin{equation}
	\left( \X^T\X + 2\BI_p   \right)\inv = 
	\frac{1}{2}\I_p - \frac{1}{4}\X^T \big( \I_n + \frac{1}{2}\X\X^T \big)\inv\X \; .
	\label{eqn:inv,lemma}
\end{equation}
In the context of our work, $n$ denotes the number of scanned subjects, which is typically on the order of a few hundred.
The matrix $(\X^T\X + 2\BI_p)\inv$ can be stored in memory if $p$ is small, but the massive dimensionality of the functional connectome in our application dismisses this option.
Therefore, we instead precompute the $(p\times n)$ matrix $\BH:=\frac{1}{4}\X^T(\I_n+\frac{1}{2}\X\X^T)\inv$ in~\eqref{eqn:inv,lemma}, and let
\[\Bvarrho\iter:=\X^T\Y^T [\va\iter-\ua\iter] +[\vb\iter-\ub\iter] +\A^T[\vd\iter-\ud\iter]\;.\]
This way, the update \eqref{eqn:w,update2} can be implemented as follows:
\begin{equation}
	w\iterp \leftarrow(\X^T\X + 2\I_p)\inv\Bvarrho\iter=\frac{1}{2}\Bvarrho\iter - \BH\X\Bvarrho\iter\;,
	\label{eqn:inv,lemma2}
\end{equation}
which allows us to carry out the \w-update without having to store a $(p\times p)$ matrix in memory.

\paragraph{\va and \vb update}
The minimization problems \eqref{eqn:v1,update1} and \eqref{eqn:v2,update1} have the form of the (scaled) proximal operator $\prox_{\tau F}:\reals^p\to\reals^p$ \citep{Rockafellar:1998:book}, defined by 
\begin{equation}
	\prox_{\tau F}(\Bv)= \argmin{\Bu\in\reals^p}\tau F(\Bu)+\frac{1}{2}\norm{\Bv-\Bu}^2 ,\;\; \tau>0\;,
	\label{eqn:prox}
\end{equation}
where $F:\reals^p\to\reals\cup\{+\infty\}$ is a closed convex function.
Using standard subdifferential calculus rules \citep{Borwein:2006:book}, it is straightforward to show that a point $\u^*\in\reals^p$ solves the minimization in \eqref{eqn:prox} if and only if the condition 
\begin{equation}
	\bzero\in \partial F(\u^*)+(\u^*-\Bv)/\tau
	\label{eqn:prox,opt,cond}
\end{equation} 
holds.
Here, $\partial F(\u^*)$ denotes the subdifferential of function $F$ at $\u^*$, defined by 
\begin{equation}
	\partial F(\u^*):= \left\{ \Bz\in\reals^p :
		 F(\u^*) + \inprod{\Bz,\u-\u^*} \leq F(\u),\; \forall \u\in\reals^p
						\right\}     .
\nonumber
\end{equation}

In addition, both updates \eqref{eqn:v1,update1} and \eqref{eqn:v2,update1} are fully separable across their coordinates, decomposing into the following sets of elementwise scalar optimization problems:
\begin{alignat}{3} 
  \left[\va\iterp\right]_i \;\;\leftarrow \;\;&
  	\prox_{\frac{\loss}{n\rho}}\left(\big[ \Y\X\w\iterp + \ua\iter \big]_i\right) ,
  	& \;\;\;\;\;\; i\in[n] \label{eqn:v1,update2} \\
  \left[\vb\iterp\right]_j \;\;\leftarrow \;\;&
  	\prox_{\frac{\lambda}{\rho}\abs{\cdot}}\left(\big[ \w\iterp+\ub\iter \big]_j\right) ,
  	& j\in[p] &\;,\label{eqn:v2,update2} 
\end{alignat}
where $[\,\cdot\,]_i$ and $[\,\cdot\,]_j$ each index the $i$-th and $j$-th element of a vector in $\reals^n$ and $\reals^p$ respectively.
For some margin-based loss functions, their corresponding proximal operator \eqref{eqn:v1,update2} can be derived in closed form using the optimality condition~\eqref{eqn:prox,opt,cond}.
Fig.~\ref{fig:prox} plots a few commonly used margin-based losses and their corresponding proximal operators, and Table~\ref{table:prox} provides their closed form expressions.
The choice of the margin-based loss is application dependent, such as whether differentiability is desired or not.
The proximal operator of the \ellone-norm~\eqref{eqn:v2,update1} and the absolute loss function~\eqref{eqn:v2,update2} corresponds to the well known \emph{soft-threshold operator} \citep{Donoho:1995}
\begin{equation}
	\text{Soft}_\tau(t):=
		\begin{cases}
			t-\tau & \text{if } t > \tau \\
			0		& \text{if } \abs{t}\leq \tau \\
			t+\tau	& \text{if } t < -\tau
		\end{cases} \;\;.
	\label{eqn:soft,thresh}
\end{equation}
The absolute loss and the soft-threshold operator are also included in Fig.~\ref{fig:prox} and Table.~\ref{table:prox} for completeness.

\afterpage{
\renewcommand{\imwidth}  {0.365\linewidth}
\renewcommand{\imwidthh}  {0.88\linewidth}
\begin{figure}[t!]
	\centering
	\begin{subfigure}[t]{\imwidth}
		 \centering
		 \includegraphics[width=\imwidthh]{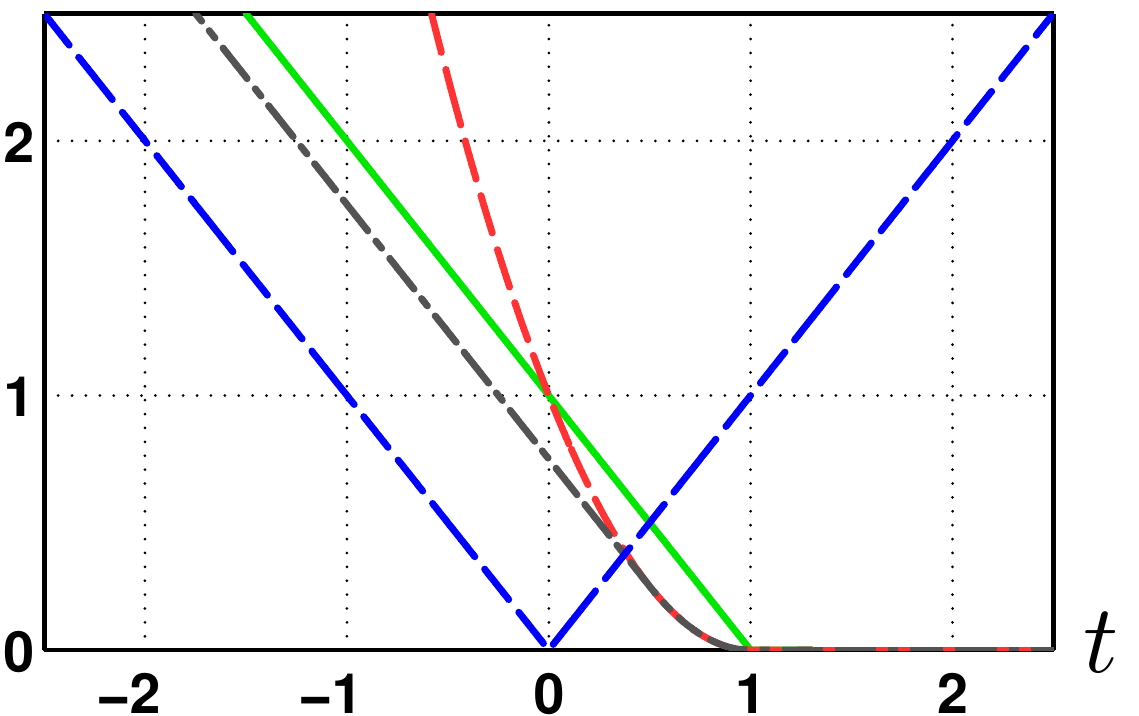}
		 \caption{{Loss functions $\loss(t)$}}
		 \label{subfig:loss,fcn}
	\end{subfigure}
	\begin{subfigure}[t]{0.225\linewidth}
		 \centering
		 \raisebox{0.17\linewidth}{\includegraphics[width=\imwidthh]{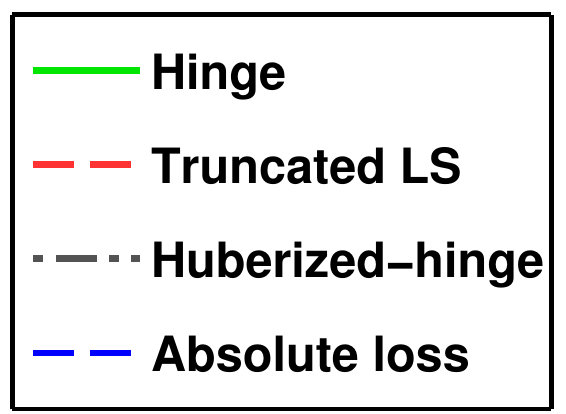}}
	\end{subfigure}
	\begin{subfigure}[t]{\imwidth}
		 \centering
		 \includegraphics[width=\imwidthh]{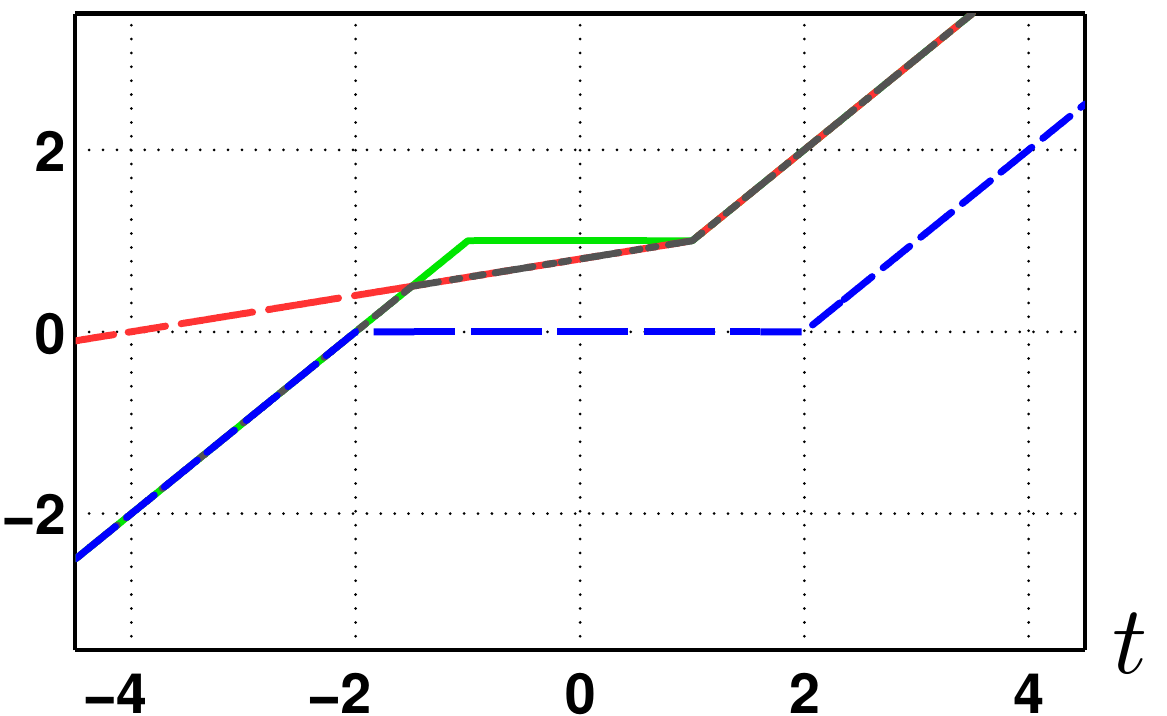}
		 \caption{{Proximal operator $\prox_{\tau\loss}(t)$}}
		 \label{subfig:loss,prox}
	\end{subfigure}%
	\caption{Plots of scalar convex loss functions that are relevant in this work, along with their associated proximal operators. 
	  Table~\ref{table:prox} provides the closed form expression for these functions.  
	  Parameter values of $\tau=2$ and $\delta=0.5$ are used in the plot for the proximal operator and the huberized hinge-loss respectively.}
	\label{fig:prox}
\end{figure}
\begin{table}[h!]\small
	\centering
	\begin{tabular}{c|c|l}
		& \normalsize{$\loss(t)$} & \hspace{0.1\linewidth}\normalsize{$\prox_{\tau\loss}(t)$}\\
		\hline\hline  
		 Hinge 			& $\max(0,1-t)$ 	& 
			 $\begin{cases}
				t 		& \text{if } t>1\\
				1 		& \text{if } 1-\tau\leq t \leq 1 \\
				t+\tau	& \text{if } t < 1-\tau
			 \end{cases}$\\ \hline
		 \begin{tabular}{c}Truncated\\ least squares\end{tabular}	
		 	& $\big\{\max(0,1-t)\big\}^2$	& 
		 	$\begin{cases}
		 		t			& \text{if } t>1 \\
		 		\dstyle\frac{t+2\tau}{1+2\tau}	& \text{if } t\leq 1 \\
		 	\end{cases}$\\ \hline
		  \begin{tabular}{c}Huberized\\ hinge \\ \citep{Wang:2008} \end{tabular}		& 
			 $\begin{cases}
				0 					&\text{if } t>1 \\
			 	\dstyle\frac{(1-t)^2}{2\delta} 	&\text{if } 1-\delta \leq t \leq 1 \\
			 	 1-t-\frac{\delta}{2}	&\text{if } t < 1-\delta
			 \end{cases}$
			 & 
			 $\begin{cases}
				t 		&\text{if } t>1 \\
			 	\dstyle\frac{t+\tau/\delta}{1+\tau/\delta} 	&\text{if } 1-\delta-\tau \leq t \leq 1 \\
			 	t+\tau	&\text{if } t < 1-\delta-\tau
			 \end{cases}$ \\ \hline
		 \begin{tabular}{c}Absolute \\ loss\end{tabular}	
		 	& \begin{tabular}{c}
		 		$\abs{t}$\\ (from \ellone-regularization)
		 	  \end{tabular}	
		 	  & 
			 $\text{Soft}_\tau(t):=
			 \begin{cases}
			 	t-\tau & \text{if } t > \tau \\
			 	0		& \text{if } \abs{t}\leq \tau \\
			 	t+\tau	& \text{if } t < -\tau
			 \end{cases}$ \\
		\hline
	\end{tabular}
	\caption{Examples of scalar convex loss functions that are relevant for this work, along with their corresponding proximal operators in closed form.}
	\label{table:prox}
\end{table}
} 

\paragraph{\vc update}
The solution to the minimization problem \eqref{eqn:v3,update1} depends on the choice of $q\in\{1,2\}$, where $q=1$ recovers fused Lasso and $q=2$ recovers GraphNet.

In the fused Lasso case $q=1$, since the masking matrix $\B\in\{0,1\}^{\etil\times\etil}$ is diagonal, the update~\eqref{eqn:v3,update1} is fully separable.
Letting $\Bzeta\iter:=\Ctil\vd\iter-\uc\iter$, the minimization problem decouples into a set of scalar minimization problems of the form:
\begin{equation}
	\argmin{v_k\in\reals}\left\{ \gamma\; b_k\abs{v_k} + 
			\frac{\rho}{2} \left(v_k-\zeta_k\iter \right)^2  
		\right\} \;,\;\;\;\; k\in[\etil]
	\label{eqn:v3,scalar}
\end{equation}
where $b_k$ is the $k$-th diagonal entry of \B and $\zeta_k\iter$ is the $k$-th entry of \sloppy{${\Bzeta\iter\in\reals^\etil}$}.
On one hand, when $b_k=0$, the minimizer for problem~\eqref{eqn:v3,scalar} returns the trivial solution $\zeta_k\iter$.
On the other hand, when $b_k=1$, the minimizer will once again have the form of the proximal operator~\eqref{eqn:prox} corresponding to the absolute loss function $\abs{\cdot}$, recovering the soft-threshold operator~\eqref{eqn:soft,thresh}.
To summarize, when $q=1$, the update for \vc \eqref{eqn:v3,update1} can be done efficiently by conducting the following elementwise update for each $k\in[\etil]$:
\begin{equation}
	\left[\vc\iterp\right]_k \;\leftarrow
	\begin{cases}
		\soft_{\gamma/\rho}\left(\left[\Ctil\left(\vd\iter-\uc\iter\right)\right]_k\right) & \text{if } \B_{k,k}=1 \\
		\left[\Ctil\left(\vd\iter-\uc\iter\right)\right]_k & \text{if } \B_{k,k}=0
	\end{cases}	
	\label{eqn:v3,update,flasso}
\end{equation}
where $\left[\cdot\right]_k$ indexes the $k$-th element of a vector in $\reals^\etil$.

In the GraphNet case $q=2$, update \eqref{eqn:v3,update1} is a quadratic optimization problem with the closed form solution
\begin{equation}
	\vc\iterp \leftarrow \rho\big(\gamma\B+\rho\I_\etil)\inv\Ctil(\vd\iter-\uc\iter) \;,
	\label{eqn:v3,update,gnet}
\end{equation}
which is trivial to compute since the matrix $(\gamma\B+\rho\I_\etil)$ is diagonal.

\paragraph{\vd update}

The closed form solution to the quadratic optimization problem \eqref{eqn:v4,update1} is 
\begin{equation}
	\vd\iterp \leftarrow \left(\Ctil^T\Ctil + \I_\ptil\right)\inv
		\left(\Ctil^T[\vc\iter+\uc\iter]+\A\w\iterp+\ud\iter\right) \;.
	\label{eqn:v4,update2}
\end{equation}
To suppress notations, let us define $\Q\in\reals^{\ptil\times\ptil}$ and $\b\in\reals^\ptil$, where
$\Q:=\Ctil^T\Ctil + \I_\ptil$
and 
\[\b:=\Ctil^T[\vc\iter+\uc\iter]+\A\w\iterp+\ud\iter .\]
As stated earlier, the Laplacian matrix $\Ctil^T\Ctil$ is block-circulant with circulant-blocks (BCCB), and consequently, the matrix \Q is BCCB as well.  
It is well known that a BCCB matrix can be diagonalized as \citep{Davis:1979:book}
	\[\Q=\U^H\BLambda\U,\]
where $\U\in\reals^{\ptil\times\ptil}$ is the ($6$-D) DFT matrix and $\BLambda\in\reals^{\ptil\times\ptil}$ is a diagonal matrix containing the ($6$-D) DFT coefficients of the first column of \Q .
As a result, the update \eqref{eqn:v4,update2} can be carried out efficiently using the (6-D) FFT 
\begin{equation}
 	\Q\inv\b = 
 		\left(\U^H \BLambda\inv \U\right)\b = 
 		\mathrm{ifft}\Big(	\mathrm{fft}(\b)\odiv\Bphi \Big) \;,
 	\label{eqn:v4,fft}
\end{equation}
where $\text{fft}$ and $\text{ifft}$ denote the ($6$-D) FFT and inverse-FFT operation\footnote{These multidimensional FFT and inverse FFT operations are implemented using \texttt{fftn} and \texttt{iffn} functions in MATLAB.}, \Bphi is a vector containing the diagonal entries of \BLambda, and $\odiv$ indicates elementwise division 
(more precisely, vectors \b and \Bphi are reshaped into $6$-D arrays prior to the $6$-D FFT and inverse-FFT operations, and the result of these operations is re-vectorized).

AL-based optimization methods that involve this kind of FFT-based inversion have been applied in image processing \citep{Afonso:2010,Allison:2013, Matakos:2013}.
Problems such as image denoising, reconstruction, and restoration are typically cast as a regularized ERM problem involving the squared loss function.
The data augmentation scheme we propose allows us to apply this FFT-based technique with $6$-D functional connectomes in the context of binary classification with margin-based loss functions. 

Finally, note that the ADMM algorithm was also used to solve the fused Lasso regularized SVM problem in \citep{Gui-Bo-Ye:2011} under a different variable splitting setup.
However, their application focuses on $1$-D data such as mass spectrometry and array CGH.
Consequently, the Laplacian matrix corresponding to their feature vector is tridiagonal with no irregularities present.
Furthermore, the variable splitting scheme they propose requires an iterative algorithm to be used for one of the ADMM subproblems.
In contrast, the variable splitting scheme and the data augmentation strategy we propose allow the ADMM subproblems to be decoupled in a way that all the updates can be carried out efficiently and non-iteratively in closed form.

\paragraph{Summary: the final algorithm and termination criteria}
Algorithm~\ref{alg:admm} outlines the complete ADMM algorithm for solving both the fused Lasso and GraphNet regularized ERM problem \eqref{eqn:costfx}, and is guaranteed to converge.
In our implementations, all the variables were initialized at zero.
The algorithm is terminated when the relative difference between two successive iterates falls below a user-specified threshold:
\begin{equation}
	\frac{\norm{\w\iterp - \w\iter}}
		{\norm{\w\iter}}
	\leq \varepsADMM \;.
	\label{eqn:admm,termin}
\end{equation}

\newcommand{\indentAlg}{\hspace{0.05\linewidth}}
\begin{algorithm}[t!]
\caption{ADMM for solving fused Lasso $(q=1)$ or GraphNet $(q=2)$}
\label{alg:admm}
\begin{algorithmic}[1]
\begin{spacing}{1.2} 
\State Initialize primal variables $\w,\va,\vb,\vc,\vd$
\State Initialize dual variables $\ua,\ub,\uc,\ud$
\State Set $t=0$, assign $\lambda\geq 0,\, \gamma\geq 0$
\State Precompute $\BH:=\frac{1}{4}\X^T(\I_n+\frac{1}{2}\X\X^T)\inv$
\Repeat
	\State \xbar-update \eqref{eqn:admm,xbar,update}
		\State\indentAlg  $\w\iterp\leftarrow \left(\X^T\X + 2\BI_p\right)\inv  
		\big(\X^T\Y^T [\va\iter-\ua\iter] 
		+[\vb\iter-\ub\iter]  
		 +\A^T[\vd\iter-\ud\iter] \big)$\Statex\Comment{apply update \eqref{eqn:inv,lemma2}}
		\State\indentAlg $\vc\iterp \leftarrow 
			\begin{cases} 
				\text{solve using \eqref{eqn:v3,update,flasso}} & \text{if } q=1 \text{ (fused Lasso)}\\
				\text{solve using \eqref{eqn:v3,update,gnet}}   & \text{if } q=2 \text{ (GraphNet)}
			\end{cases}$
	\State \ybar-update \eqref{eqn:admm,ybar,update}
		\State\indentAlg $\va\iterp \leftarrow \prox_{\frac{\Loss}{n\rho}}\left(\Y\X\w\iterp+\ua\iter\right)$
			 \Comment{apply \eqref{eqn:v1,update2} elementwise}
		\State\indentAlg $\vb\iterp \leftarrow 
			\soft_{\lambda/\rho}\left(\w\iterp+\ub\iter\right)$
			\Comment{apply \eqref{eqn:v2,update2} elementwise}
		\State\indentAlg {\normalsize $\vd\iterp \leftarrow 
			\left(\Ctil^T\Ctil + \I_\ptil\right)\inv
			\left(\Ctil^T[\vc\iterp+\uc\iter]+\A\w\iterp+\ud\iter\right)$ \Statex\Comment{solve using FFT approach \eqref{eqn:v4,fft}}}
	\State \u-update \eqref{eqn:admm,dual,update}	
		\State\indentAlg $\ua\iterp \leftarrow \ua\iter + \Y\X\w\iterp-\va\iterp$
		\State\indentAlg $\ub\iterp \leftarrow \ub\iter + \w\iterp - \vb\iterp$
		\State\indentAlg $\uc\iterp \leftarrow \uc\iter + \vc\iterp - \Ctil\vd\iterp$
		\State\indentAlg $\ud\iterp \leftarrow \ud\iter + \A\w\iterp - \vd\iterp$
	\State $t\leftarrow t+1$
\Until{stopping criterion is met}
\end{spacing}
\end{algorithmic}
\end{algorithm}

\subsection{Generation of synthetic data: $4$-D functional connectomes}
\label{subsec:synthetic,4d,conn}
\newcommand{\calI}{\xmath{\mathcal{I}}}
\newcommand{\calK}{\xmath{\mathcal{K}}}
\newcommand{\calC}{\xmath{\mathcal{C}}}
\newcommand{\calE}{\xmath{\mathcal{E}}}
\newcommand{\calN}{\xmath{\mathcal{N}}}
To assess the validity of our method, we ran experiments on synthetic $4$-D functional connectome data.
The data were generated to imitate functional connectomes resulting from a single slice of our grid-based parcellation scheme (see Fig.~\ref{fig:roi,grid,slice}).
Specifically, we selected only the nodes that are present at axial slice $z=18$ in the MNI space; this slice was selected for its substantial $X$ and $Y$ coverage.
Fig.~\ref{subfig:sim,conn,struct,hc} provides a schematic representation of the selected nodes.

\newcommand{\muhat}{\xmath{\hat{\mu}}}
\newcommand{\mutil}{\xmath{\widetilde{\mu}}}
\newcommand{\sighat}{\xmath{\hat{\sigma}}}
\newcommand{\arctanh}{\text{arctanh}}
To mimic the \emph{control vs. patient} binary classification setup, we created two classes of functional connectomes sampled from random normal distributions. 
The mean and the variance for these distributions were assigned using the functional connectomes generated from the real resting state dataset described later in Sec.~\ref{subsec:real,data}.
Specifically, we first took the subject-level functional connectomes corresponding to the $67$ healthy controls in the dataset, and extracted the entries that represent the edges among the nodes at slice $z=18$.
Since there are $66$ nodes within this slice, this gives us $\binom{66}{2}=2145$ edges for each subjects.
Next, we applied Fisher transformation on these edges to map the correlation values to the real line.
For each of these transformed edges, we calculated the inter-subject sample mean and sample variance, which we denote by $\{\muhat(k),\sighat^2(k)\}$ with $k\in[2145]$ indexing the edges.
Finally, a synthetic subject-level ``control class'' connectome is realized by sampling edges individually from a set of random normal distributions having the above mean and variance, and then applying inverse Fisher transformation $\tanh:\reals\to (-1,+1)$ on these sampled edges, \ie,
\[
	\x=\left[
		\tanh\big(x^{(1)}\big),
		\dots,
		\tanh\big(x^{(2145)}\big)
	\right]^T
	\text{ where } x^{(k)}\sim\calN\left(\muhat(k),\sighat^2(k)\right),\; k\in[2145].
\]
Realizations of the ``patient class'' connectomes are generated in a similar manner, but here we introduced two clusters of \emph{anomalous nodes}, indicated by the red nodes in Fig.~\ref{subfig:sim,conn,struct,ds}.
These clusters participate in a disease-specific perturbation, where signal was added to all connections originating in one cluster and terminating in the other.
More formally, let $\calK\subset[2145]$ denote the index set corresponding to these disease-specific \emph{anomalous edges}, which consist of a complete bipartite graph formed by the anomalous node clusters $\calC_1=\{8,14,15,16,23\}$ and $\calC_2=\{41,48,49,50,56\}$, $\calC_1,\calC_2\subset[66]$.
Under these notations, a synthetic subject-level ``patient class'' connectome is realized by the following procedure:
\[
	\x=\left[
		\tanh\big(x^{(1)}\big),
		\dots,
		\tanh\big(x^{(2145)}\big)
	\right]^T 
	\text{where } 
	\begin{cases}
		x^{(k)}\sim\calN\Big(\muhat(k),\sighat^2(k)\Big) &\text{ if } k\notin\calK \vspace{3pt}\\
		x^{(k)}\sim\calN\Big(\muhat(k)+d\cdot\sighat(k),\;\sighat^2(k)\Big) &\text{ if } k\in\calK\;.
	\end{cases}
\]
In other words, if an edge $k$ is a member of the anomalous edge set \calK, a non-random signal $d\cdot\sighat(k)$ is added to the sampled edge-value.
Here, $d$ denotes Cohen's effect size \citep{Cohen:1988}, which we set at $d=0.6$ for our experiments.
Overall, since $\abs{\calC_1}=\abs{\calC_2}=5$, we have $\abs{\calK}=\abs{\calC_1}\cdot\abs{\calC_2}=25$, \ie, there are $25$ anomalous edges in the patient group; see Fig.~\ref{subfig:sim,conn,struct,ds} for a pictorial illustration of the anomalous edge set \calK in the $2$-D node space.
Fig.~\ref{subfig:sim,conn,struct,supp} presents a binary support matrix indicating the structure of the anomalous edges in the $4$-D connectome space, with the locations of the anomalous edges specified by the product set $\calC_1\times \calC_2\subset[66]\times[66]$. 

\begin{figure}
	\centering
	\begin{subfigure}[b]{0.25\linewidth}
		 \centering
		 \resizebox{0.94\linewidth}{!}{\begin{tikzpicture}[-,red, line width=0.06cm,font=\bfseries,draw=black] 
	\tikzstyle{every node}=[circle,ultra thick,draw=black,fill=white,text=black,minimum size=0.55cm,line width=0.03cm,inner sep=0.8pt]
	\tikzstyle{HC} =[fill=blue!20] 
	\tikzstyle{DS} =[fill=blue!20]  
	\matrix [rectangle,row sep=2.5pt, column sep=5pt,draw=black,fill=green!5,line width=0.025cm] %
	{
				    ;&				;&				 ;& \node(65)[HC]{65} ;& \node(66)[HC]{66} ;\\
		 		    ;& \node(59)[HC]{59} ;& \node(60)[HC]{60} ;& \node(61)[HC]{61} ;& \node(62)[HC]{62} ;& \node(63)[HC]{63} ;& \node(64)[HC]{64} ;& ;\\
		 		    ;& \node(52)[HC]{52} ;& \node(53)[HC]{53} ;& \node(54)[HC]{54} ;& \node(55)[HC]{55} ;& \node(56)[DS]{56} ;& \node(57)[HC]{57} ;& \node(58)[HC]{58} ;\\
	 \node(44)[HC]{44} ;& \node(45)[HC]{45} ;& \node(46)[HC]{46} ;& \node(47)[HC]{47} ;& \node(48)[DS]{48} ;& \node(49)[DS]{49} ;& \node(50)[DS]{50} ;& \node(51)[HC]{51} ;\\
	 \node(36)[HC]{36} ;& \node(37)[HC]{37} ;& \node(38)[HC]{38} ;& \node(39)[HC]{39} ;& \node(40)[HC]{40} ;& \node(41)[DS]{41} ;& \node(42)[HC]{42} ;& \node(43)[HC]{43} ;\\
	 \node(28)[HC]{28} ;& \node(29)[HC]{29} ;& \node(30)[HC]{30} ;& \node(31)[HC]{31} ;& \node(32)[HC]{32} ;& \node(33)[HC]{33} ;& \node(34)[HC]{34} ;& \node(35)[HC]{35} ;\\
	 \node(20)[HC]{20} ;& \node(21)[HC]{21} ;& \node(22)[HC]{22} ;& \node(23)[DS]{23} ;& \node(24)[HC]{24} ;& \node(25)[HC]{25} ;& \node(26)[HC]{26} ;& \node(27)[HC]{27} ;\\
	 \node(12)[HC]{12} ;& \node(13)[HC]{13} ;& \node(14)[DS]{14} ;& \node(15)[DS]{15} ;& \node(16)[DS]{16} ;& \node(17)[HC]{17} ;& \node(18)[HC]{18} ;& \node(19)[HC]{19} ;\\
		 		    ;& \node(6)[HC]{6} 	;& \node(7)[HC]{7} 	 ;& \node(8)[DS]{8}	  ;& \node(9)[HC]{9}   ;& \node(10)[HC]{10} ;& \node(11)[HC]{11} ;&  ;\\
	 			    ;&				;& \node(1)[HC]{1} 	 ;& \node(2)[HC]{2}	  ;& \node(3)[HC]{3}   ;& \node(4)[HC]{4}   ;& \node(1)[HC]{5}   ;&  ;\\
	};
\end{tikzpicture}}
		 \caption{Control}
		 \label{subfig:sim,conn,struct,hc}
	\end{subfigure}
	\hspace{-4pt}
	\begin{subfigure}[b]{0.5\linewidth}
		 \centering
		 \resizebox{0.94\linewidth}{!}{\begin{tikzpicture}[-,red, line width=0.06cm,font=\bfseries,draw=black] 
	\tikzstyle{every node}=[circle,ultra thick,draw=black,fill=white,text=black,minimum size=0.55cm,line width=0.03cm,inner sep=0.8pt]
	\tikzstyle{HC} =[fill=blue!20] 
	\tikzstyle{DS} =[fill=red!25]  
	\matrix [rectangle,row sep=2.5pt, column sep=5pt,draw=black,fill=green!5,line width=0.025cm] %
	{
				    ;&				;&				 ;& \node(65)[HC]{65} ;& \node(66)[HC]{66} ;\\
		 		    ;& \node(59)[HC]{59} ;& \node(60)[HC]{60} ;& \node(61)[HC]{61} ;& \node(62)[HC]{62} ;& \node(63)[HC]{63} ;& \node(64)[HC]{64} ;& ;\\
		 		    ;& \node(52)[HC]{52} ;& \node(53)[HC]{53} ;& \node(54)[HC]{54} ;& \node(55)[HC]{55} ;& \node(56)[DS]{56} ;& \node(57)[HC]{57} ;& \node(58)[HC]{58} ;\\
	 \node(44)[HC]{44} ;& \node(45)[HC]{45} ;& \node(46)[HC]{46} ;& \node(47)[HC]{47} ;& \node(48)[DS]{48} ;& \node(49)[DS]{49} ;& \node(50)[DS]{50} ;& \node(51)[HC]{51} ;\\
	 \node(36)[HC]{36} ;& \node(37)[HC]{37} ;& \node(38)[HC]{38} ;& \node(39)[HC]{39} ;& \node(40)[HC]{40} ;& \node(41)[DS]{41} ;& \node(42)[HC]{42} ;& \node(43)[HC]{43} ;\\
	 \node(28)[HC]{28} ;& \node(29)[HC]{29} ;& \node(30)[HC]{30} ;& \node(31)[HC]{31} ;& \node(32)[HC]{32} ;& \node(33)[HC]{33} ;& \node(34)[HC]{34} ;& \node(35)[HC]{35} ;\\
	 \node(20)[HC]{20} ;& \node(21)[HC]{21} ;& \node(22)[HC]{22} ;& \node(23)[DS]{23} ;& \node(24)[HC]{24} ;& \node(25)[HC]{25} ;& \node(26)[HC]{26} ;& \node(27)[HC]{27} ;\\
	 \node(12)[HC]{12} ;& \node(13)[HC]{13} ;& \node(14)[DS]{14} ;& \node(15)[DS]{15} ;& \node(16)[DS]{16} ;& \node(17)[HC]{17} ;& \node(18)[HC]{18} ;& \node(19)[HC]{19} ;\\
		 		    ;& \node(6)[HC]{6} 	;& \node(7)[HC]{7} 	 ;& \node(8)[DS]{8}	  ;& \node(9)[HC]{9}   ;& \node(10)[HC]{10} ;& \node(11)[HC]{11} ;&  ;\\
	 			    ;&				;& \node(1)[HC]{1} 	 ;& \node(2)[HC]{2}	  ;& \node(3)[HC]{3}   ;& \node(4)[HC]{4}   ;& \node(1)[HC]{5}   ;&  ;\\
	};
	\tikzstyle{col}=[red] 
	\draw(15)to[](41)[col];
	\draw(15)to[out=125,in=-160](48)[col];
	\draw(15)to[](49)[col];
	\draw(15)to[out=25,in=-115](50)[col];
	\draw(15)to[out=68,in=-119](56)[col];
	\tikzstyle{col}=[green!100] 
	\draw(14)to[out=130,in=190](41)[col];
	\draw(14)to[out=130,in=180](48)[col];
	\draw(14)to[out=130,in=200](49)[col];
	\draw(14)to[out=130,in=205](50)[col];
	\draw(14)to[out=130,in=180](56)[col];	
	\tikzstyle{col}=[blue] 
	\draw(8)to[out=-15,in=-70](41)[col];
	\draw(8)to[out=-15,in=-70](48)[col];
	\draw(8)to[out=-15,in=-55](49)[col];
	\draw(8)to[out=-15,in=-65](50)[col];
	\draw(8)to[out=-15,in=-55](56)[col];
\end{tikzpicture}
\begin{tikzpicture}[-,red, line width=0.06cm,font=\bfseries,draw=black] 
	\tikzstyle{every node}=[circle,ultra thick,draw=black,fill=white,text=black,minimum size=0.55cm,line width=0.03cm,inner sep=0.8pt]
	\tikzstyle{HC} =[fill=blue!20] 
	\tikzstyle{DS} =[fill=red!25]  
	\matrix [rectangle,row sep=2.5pt, column sep=5pt,draw=black,fill=green!5,line width=0.025cm] %
	{
				    ;&				;&				 ;& \node(65)[HC]{65} ;& \node(66)[HC]{66} ;\\
		 		    ;& \node(59)[HC]{59} ;& \node(60)[HC]{60} ;& \node(61)[HC]{61} ;& \node(62)[HC]{62} ;& \node(63)[HC]{63} ;& \node(64)[HC]{64} ;& ;\\
		 		    ;& \node(52)[HC]{52} ;& \node(53)[HC]{53} ;& \node(54)[HC]{54} ;& \node(55)[HC]{55} ;& \node(56)[DS]{56} ;& \node(57)[HC]{57} ;& \node(58)[HC]{58} ;\\
	 \node(44)[HC]{44} ;& \node(45)[HC]{45} ;& \node(46)[HC]{46} ;& \node(47)[HC]{47} ;& \node(48)[DS]{48} ;& \node(49)[DS]{49} ;& \node(50)[DS]{50} ;& \node(51)[HC]{51} ;\\
	 \node(36)[HC]{36} ;& \node(37)[HC]{37} ;& \node(38)[HC]{38} ;& \node(39)[HC]{39} ;& \node(40)[HC]{40} ;& \node(41)[DS]{41} ;& \node(42)[HC]{42} ;& \node(43)[HC]{43} ;\\
	 \node(28)[HC]{28} ;& \node(29)[HC]{29} ;& \node(30)[HC]{30} ;& \node(31)[HC]{31} ;& \node(32)[HC]{32} ;& \node(33)[HC]{33} ;& \node(34)[HC]{34} ;& \node(35)[HC]{35} ;\\
	 \node(20)[HC]{20} ;& \node(21)[HC]{21} ;& \node(22)[HC]{22} ;& \node(23)[DS]{23} ;& \node(24)[HC]{24} ;& \node(25)[HC]{25} ;& \node(26)[HC]{26} ;& \node(27)[HC]{27} ;\\
	 \node(12)[HC]{12} ;& \node(13)[HC]{13} ;& \node(14)[DS]{14} ;& \node(15)[DS]{15} ;& \node(16)[DS]{16} ;& \node(17)[HC]{17} ;& \node(18)[HC]{18} ;& \node(19)[HC]{19} ;\\
		 		    ;& \node(6)[HC]{6} 	;& \node(7)[HC]{7} 	 ;& \node(8)[DS]{8}	  ;& \node(9)[HC]{9}   ;& \node(10)[HC]{10} ;& \node(11)[HC]{11} ;&  ;\\
	 			    ;&				;& \node(1)[HC]{1} 	 ;& \node(2)[HC]{2}	  ;& \node(3)[HC]{3}   ;& \node(4)[HC]{4}   ;& \node(1)[HC]{5}   ;&  ;\\
	};
	\tikzstyle{col}=[black] 
	\draw(23)to[out=15,in=-90](41)[col];
	\draw(23)to[](48)[col];
	\draw(23)to[](49)[col];
	\draw(23)to[out=0,in=-90](50)[col];
	\draw(23)to[out=90,in=180](56)[col];

	\tikzstyle{col}=[myorange] 
	\draw(16)to[in=-80](41)[col];
	\draw(16)to[](48)[col];
	\draw(16)to[out=78,in=-120](49)[col];
	\draw(16)to[out=30,in=-80](50)[col];
	\draw(16)to[out=82,in=-123](56)[col];
\end{tikzpicture}}
		 \caption{Patient}
		 \label{subfig:sim,conn,struct,ds}
	\end{subfigure}
	\begin{subfigure}[b]{0.235\linewidth}
		 \centering
		 \raisebox{0pt}{\hspace{-8pt}\includegraphics[width=1.125\linewidth]{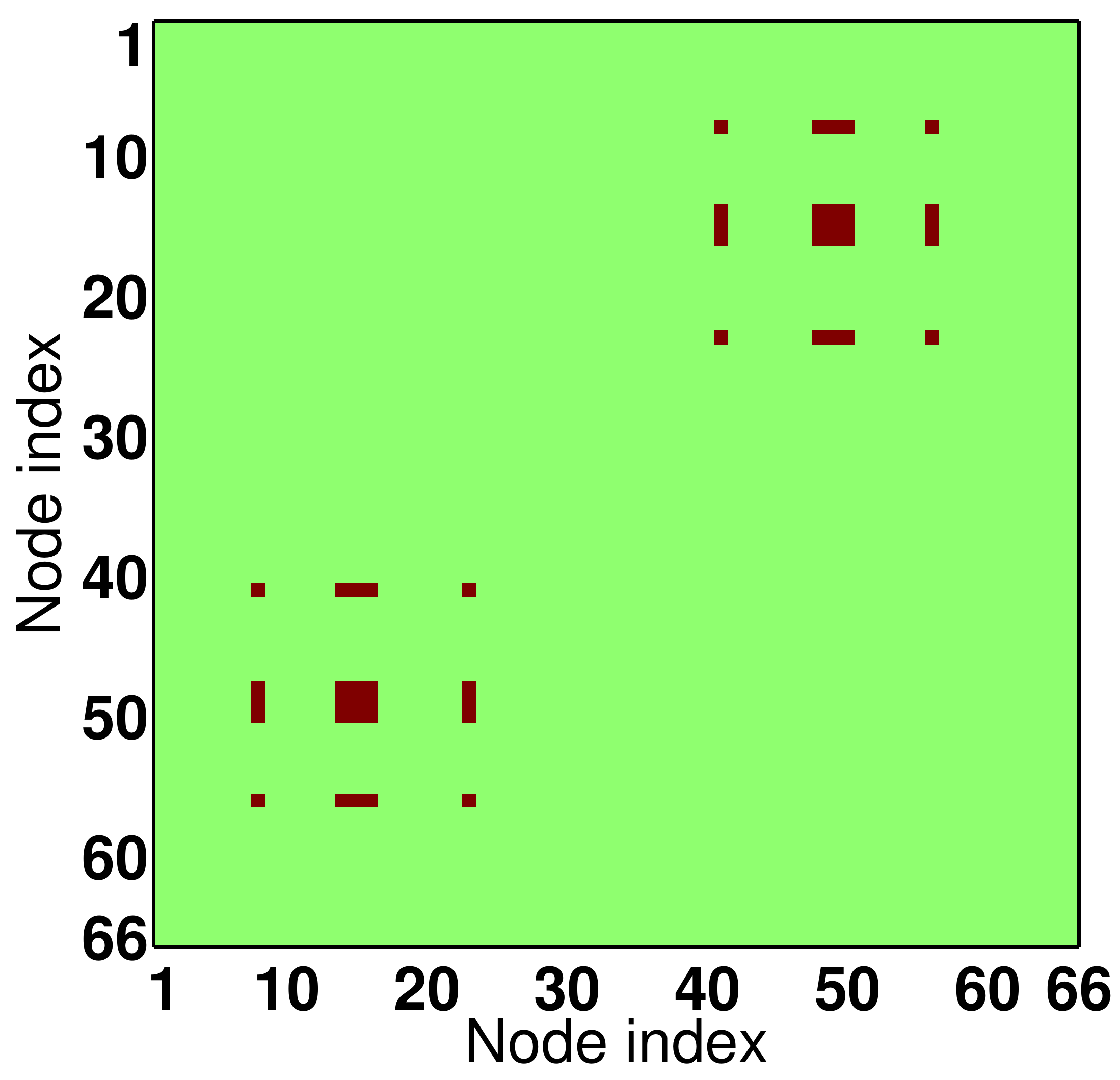}}\vspace{-5pt}\\
		 \caption{Edge support}
		 \label{subfig:sim,conn,struct,supp}
	\end{subfigure}
	\vspace{-12pt}
	\caption{
		Schematic representations of the synthetic $4$-D functional connectome data generated for the simulation experiments (best viewed in color). 
		(a) Node orientation representing the ``control class'' connectome, where the blue nodes indicate
the normal nodes.
		(b) Node orientation representing the ``patient class'' connectome, where there are $25$ \emph{anomalous edges} shared among the two \emph{anomalous node} clusters indicated in red (this subfigure is split into two side-by-side figures to improve visibility of the impacted edges).
		(c) Binary support matrix indicating the locations of the anomalous edges in the connectome space.
	}
	\label{fig:sim,conn,struct}
\end{figure}
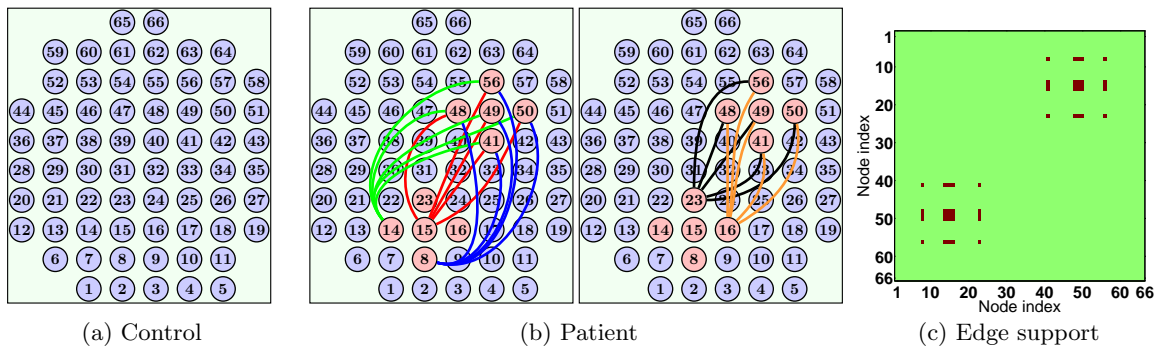

It is important to note that the inclusion of the clusters of anomalous nodes is motivated from the ``patchiness assumption'' of brain disorders, a view that has been born from multiple task-based and connectivity-based studies; this point will be expounded in finer detail in \ref{subsec:why,flasso}.
In short, the ``patchiness assumption'' is the view that major psychiatric disorders manifest in the brain by impacting moderately sized spatially contiguous regions,  which is what the clusters of anomalous nodes are intended to mimic in this simulation.

For training the classifiers, we sampled $100$ functional connectomes consisting of $50$ control samples and $50$ patient samples. 
For evaluating the performance of the classifiers, we sampled $500$ additional functional connectomes consisting of $250$ control samples and $250$ patient samples.

\subsection{Real experimental data: schizophrenia resting state dataset}
\label{subsec:real,data}
To further assess the utility of the proposed method, we also conducted experiments on real resting state scans.

\paragraph{Participants}
We used the Center for Biomedical Research Excellence (COBRE) dataset (\url{http://fcon_1000.projects.nitrc.org/indi/retro/cobre.html}) made available by the Mind Research Network. 
The dataset is comprised of $74$ typically developing control participants and $71$ participants with a DSM-IV-TR diagnosis of schizophrenia. 
Diagnosis was established by the Structured Clinical Interview for DSM-IV (SCID). Participants were excluded if they had mental retardation, neurological disorder, head trauma, or substance abuse or dependence in the last $12$ months. A summary of the participant demographic characteristics is provided in Table~\ref{table:schiz,demographic}. 

Data collection was performed at the Mind Research Network, and funded by a Center of Biomedical Research Excellence (COBRE) grant 5P20RR021938/P20GM103472 from the NIH to Dr. Vince Calhoun. The COBRE data set can also be downloaded from the Collaborative Informatics and Neuroimaging Suite data exchange tool (COINS~\citep{Scott:2011}; \url{http://coins.mrn.org/dx}).

\paragraph{Data Acquisition}
A multi-echo MPRAGE (MEMPR) sequence was used with the following parameters: TR/TE/TI = $2530/[1.64, 3.5, 5.36, 7.22, 9.08]$/$900$ ms, flip angle $= 7^{\circ}$, FOV $= 256\times 256$ mm, slab thickness $= 176$ mm, matrix size $= 256\times 256\times 176$, voxel size $= 1\times 1\times 1$ mm, number of echoes $= 5$, pixel bandwidth $=650$ Hz, total scan time $= 6$ minutes. 
With $5$ echoes, the TR and TI time to encode partitions for the MEMPR are similar to that of a conventional MPRAGE, resulting in similar GM/WM/CSF contrast. 
Resting state data were collected with single-shot full k-space echo-planar imaging (EPI) with ramp sampling correction using the intercomissural line (AC-PC) as a reference (TR: $2$ s, TE: $29$ ms, matrix size: $64\times 64$, $32$ slices, voxel size: $3\times 3\times 4 mm^3$).

\paragraph{Imaging Sample Selection}
Analyses were limited to participants with: 
(1) MPRAGE anatomical images, with consistent near-full brain coverage (\ie, superior extent included the majority of frontal and parietal cortex and inferior extent included the temporal lobes) with successful registration; 
(2) complete phenotypic information for main phenotypic variables (diagnosis, age, handedness); 
(3) mean framewise displacement (FD) within two standard deviations of the sample mean; 
(4) at least $50\%$ of frames retained after application of framewise censoring for motion (``motion scrubbing''; see below).
After applying these sample selection criteria, we analyzed resting state scans from $121$ individuals consisting of $67$ healthy controls (HC) and $54$ schizophrenic subjects (SZ). 
Demographic characteristics of the post-exclusion sample are shown in Table~\ref{table:schiz,demographic}.

\newcommand{\mycellcol}{\cellcolor{blue!25}}
\begin{table}[t!]
	\centering
	\begin{tabular}{l|cccc|cccc}
		\multicolumn{1}{c}{}&\multicolumn{4}{c}{\textbf{Healthy Controls}}&\multicolumn{4}{c}{\textbf{Schizophrenia}}\\
		\hline
			& $n$ & Age & \#male & \#RH & $n$ & Age & \#male & \#RH \\		
		\hline\hline
		 Pre-exclusion 	& $74$ & $35.8\pm 11.6$ & $51$ & $71$ 
		 		  		& $71$ & $38.1\pm 14.0$ & $57$ & $59$ \\
		 Post-exclusion 	& $67$ & $35.2\pm 11.7$ & $46$ & $66$ 
		 				& $54$ & $35.5\pm 13.1$ & $48$ & $46$ \\
		\hline
	\end{tabular}
	\caption{
		Demographic characteristics of the participants before and after sample exclusion criteria is applied \mbox{(RH = right-handed)}.
	}
	\label{table:schiz,demographic}
\end{table}

\paragraph{Preprocessing}
Preprocessing steps were performed using statistical parametric mapping (SPM$8$; \url{www.fil.ion.ucl.ac.uk/spm}). 
Scans were reconstructed, slice-time corrected, realigned to the first scan in the experiment for correction of head motion, and co-registered with the high-resolution T$1$-weighted image. 
Normalization was performed using the voxel-based morphometry (VBM) toolbox implemented in SPM$8$. 
The high-resolution T$1$-weighted image was segmented into tissue types, bias-corrected, registered to MNI space, and then normalized using Diffeomorphic Anatomical Registration Through Exponentiated Lie Algebra (DARTEL) \citep{Ashburner:2007}. 
The resulting deformation fields were then applied to the functional images. 
Smoothing of functional data was performed with an $8$ mm$^3$ kernel.

\paragraph{Connectome generation}
Functional connectomes were generated by placing $7.5$ mm radius nodes representing ROIs encompassing $33$ $3\times 3\times 3$ mm voxels in a regular grid spaced at $18\times 18 \times 18$~mm intervals throughout the brain. 
Spatially averaged time series were extracted from each of the ROIs. 
Next, linear detrending was performed, followed by nuisance regression. 
Regressors included six motion regressors generated from the realignment step, as well as their first derivatives. 
White matter and cerebrospinal fluid masks were generated from the VBM-based tissue segmentation step noted above, and eroded using the \textsf{fslmaths} program from FSL to eliminate border regions of potentially ambiguous tissue type. 
The top five principal components of the BOLD time series were extracted from each of the masks and included as regressors in the model -- a method that has been demonstrated to effectively remove signals arising from the cardiac and respiratory cycle \citep{Behzadi:2007}. 
The time-series for each ROI was then band-passed filtered in the $0.01$ -- $0.10$ Hz range. 
Individual frames with excessive head motion were then censored from the time series. 
Subjects with more than $50\%$ of their frames removed by scrubbing were excluded from further analysis, a threshold justified by simulations conducted by other groups \citep{Fair:2013}, as well as by our group. 
Pearson product-moment correlation coefficients were then calculated pairwise between time courses for each of the $347$ ROIs.
Standard steps in functional connectivity analysis (removing motion artifacts and nuisance covariates and calculating Pearson's product moment correlations between pairs of nodes) was performed with \texttt{ConnTool}, a functional connectivity analysis package developed by Robert C. Welsh, University of Michigan.

	\section{Results}
	\label{results}
\subsection{Results on synthetic functional connectome data}
\label{subsec:result,synth}
In order to evaluate the validity of our proposed method, we compared the performance of four linear classifiers trained on the synthetic functional connectome data described in Section~\ref{subsec:synthetic,4d,conn}, where the training set consists of $100$ samples with $50$ patients and $50$ controls.
Specifically, we solved the regularized ERM problem~\eqref{eqn:reg,erm} using the hinge-loss and the following four regularizers: Lasso, Elastic-net, GraphNet, and fused Lasso.
Lasso and Elastic-net were also solved using ADMM, although the variable splitting scenario and the optimization steps are different from Algorithm~\ref{alg:admm}.
The ADMM algorithm for Elastic-net is provided in \ref{appendix,admm,enet}, and the algorithm for Lasso follows directly from Elastic-net by setting $\gamma=0$.
The ADMM algorithm was terminated when the tolerance level~\eqref{eqn:admm,termin} fell below $\varepsilon=4\times 10^{-3}$ or the algorithm reached $400$ iterations.
Note that in our experiment, we let $y=+1$ indicate the ``patient class'' and $y=-1$ indicate the ``control class.''

\newcommand{\FOOTMESSAGE}{The grid search region for $\gamma$ is different for fused Lasso since we observed a clear drop-off in classification performance for any values of $\gamma$ higher than the range presented.  
We found this to be true for the real data experiment in Sec.~\ref{subsec:result,real,data} as well; see Fig.~\ref{fig:sim,grid,ntr} and Fig.~\ref{fig:grid,search}.}

With the exception of Lasso, the regularizers we investigated involve two tuning parameters: $\lambda\geq~0$ and $\gamma\geq 0$.
We tuned these regularization parameters by conducting a \mbox{$5$-fold} cross-validation on the training set over a two-dimensional grid, and tuned Lasso over a one-dimensional grid.
More precisely, the $\ellone$ regularization parameter $\lambda\geq 0$ was tuned over the range \sloppy{${\lambda\in\{2^{-11},2^{-10.75},\dots,2^{-3.5}\}}$} for all four regularizers.
The second regularization parameter $\gamma\geq 0$ was tuned over the range $\gamma\in\{2^{-16},2^{-15.5},\dots,2^{+2}\}$ for Elastic-net and GraphNet and $\gamma\in\{2^{-16},2^{-15.5},\dots,2^{-5}\}$ for fused Lasso\footnote{\FOOTMESSAGE}.
The final weight vector estimates are obtained by re-training the classifiers on the entire training set using the regularization parameter values $\{\lambda,\gamma\}$ that yielded the highest $5$-fold cross-validation classification accuracy.
For visualization, the estimated weight vectors are reshaped into $66\times 66$ symmetric matrices with zeroes on the diagonal (although these are matrices, we will refer to them as ``weight vectors'' as well), and the classification accuracies are evaluated on a testing set consisting of $500$ samples with $250$ patients and $250$ controls.

\begin{figure}[ptbh]
	\centering
	\renewcommand{\imwidth} {0.23\linewidth}
	\setlength{\tabcolsep}{4.25pt} 
	\begin{tabular}{cccc}
		\includegraphics[width=\imwidth]{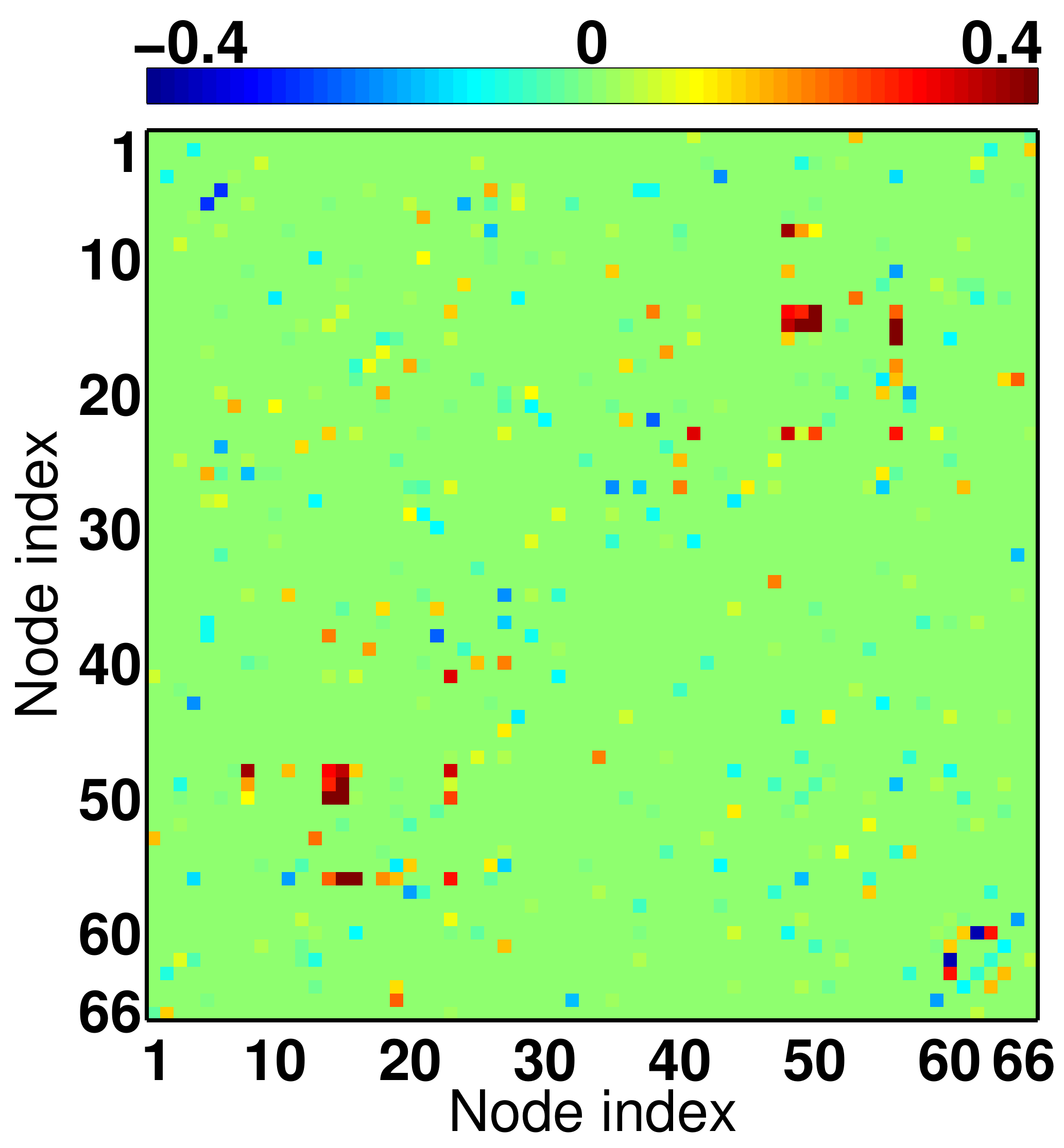} &
		\includegraphics[width=\imwidth]{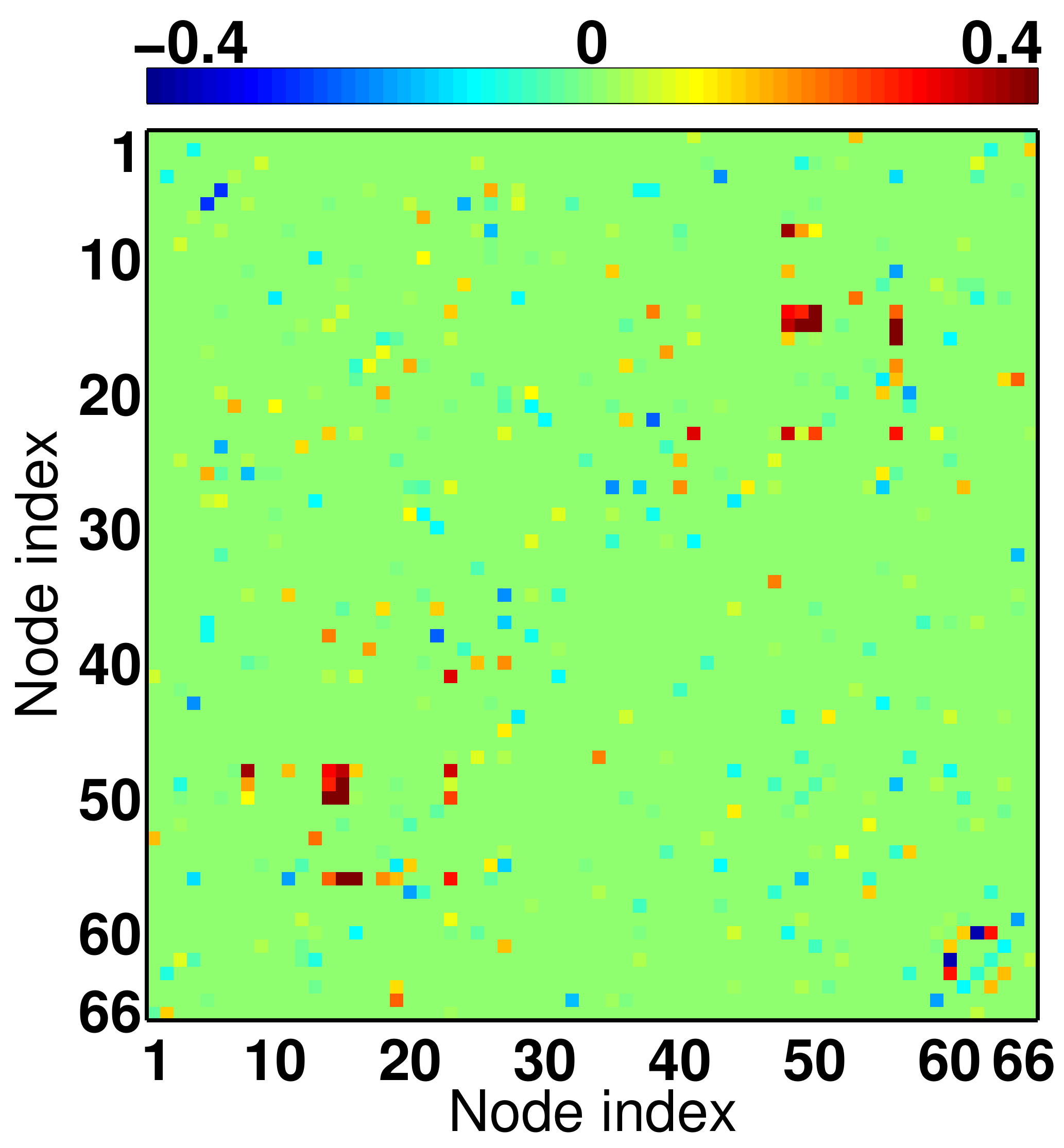} &
		\includegraphics[width=\imwidth]{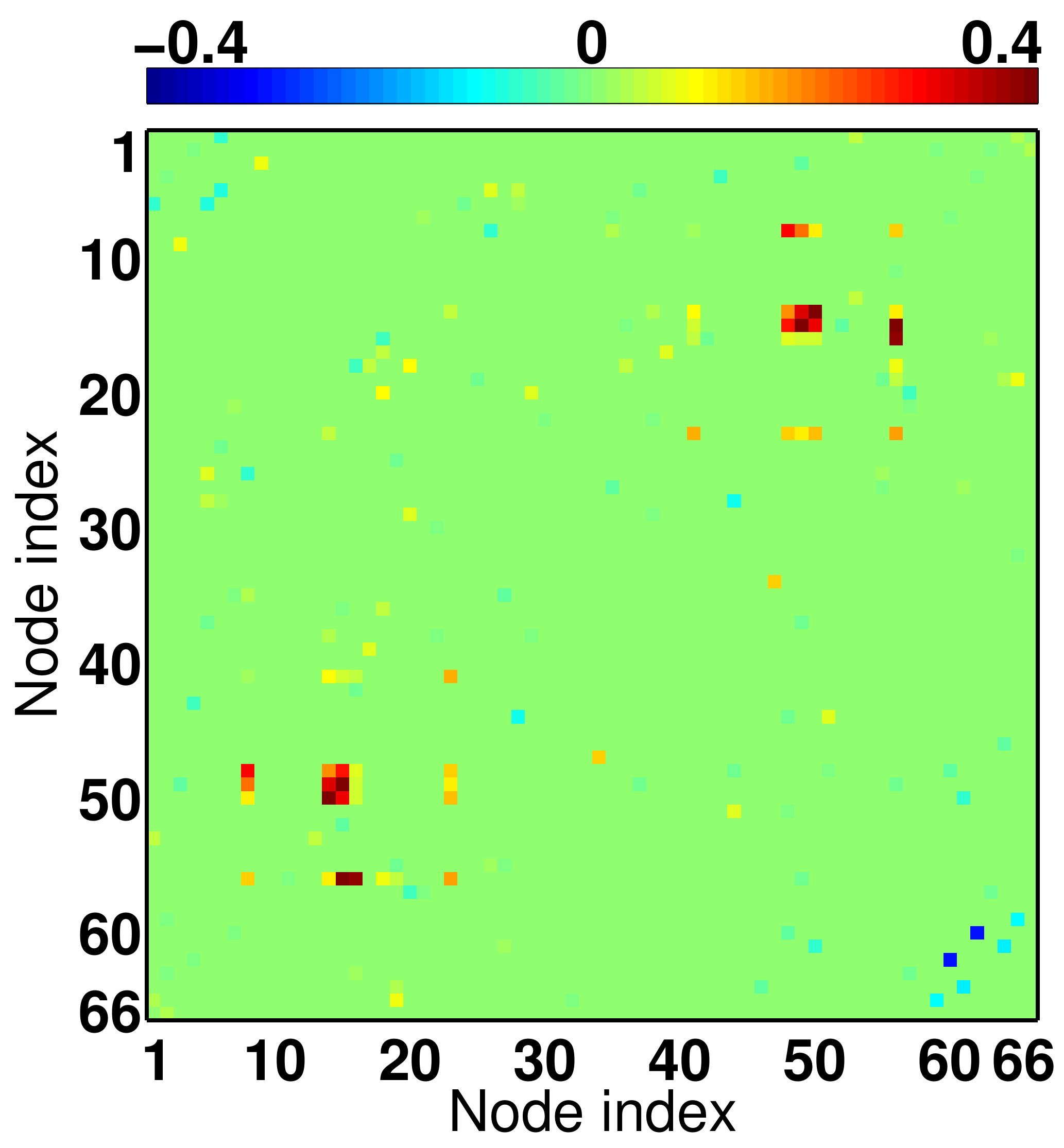} &
		\includegraphics[width=\imwidth]{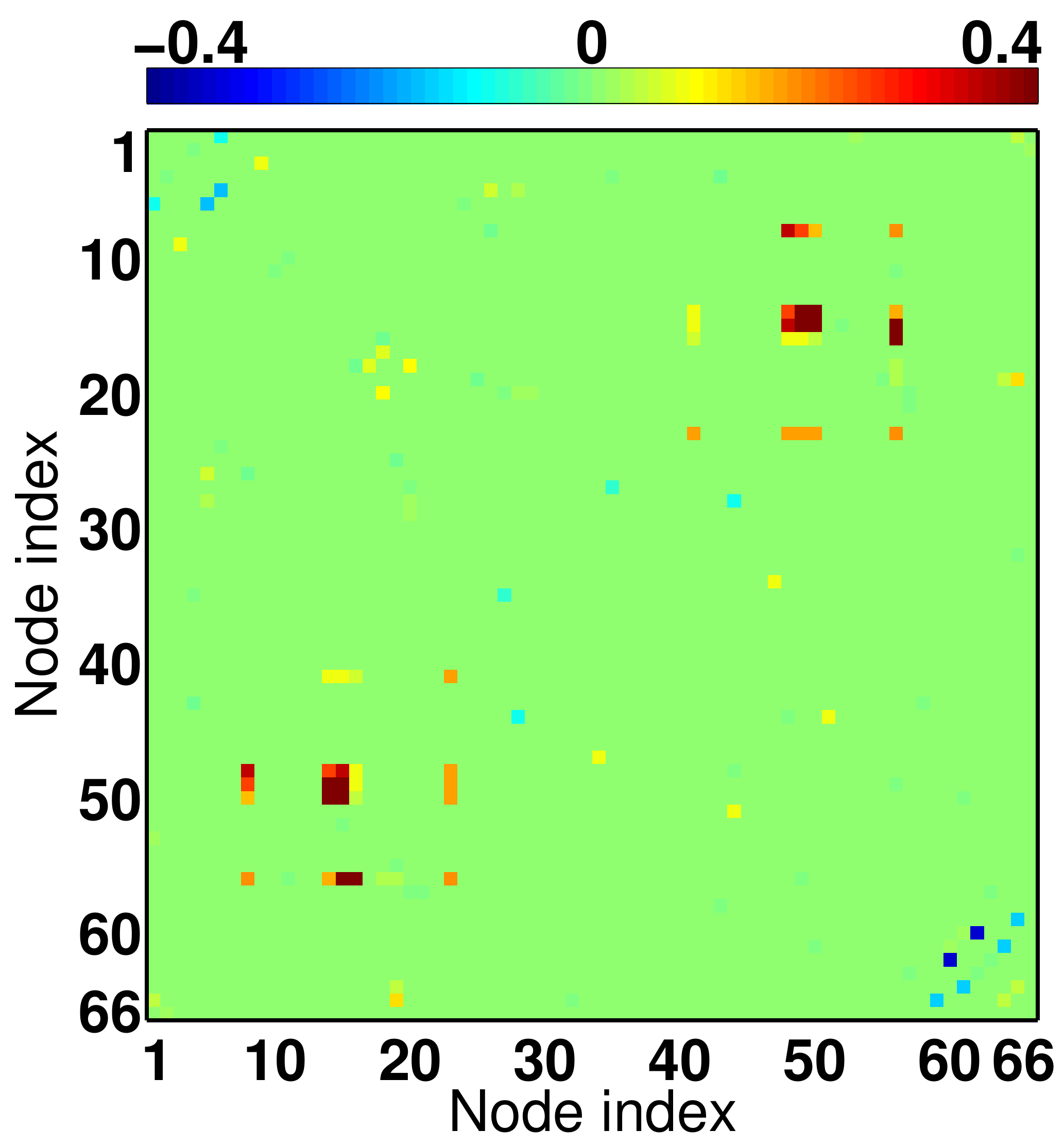} \vspace{-3pt}\\
		\small{(a) Lasso} & \small{(b) Elastic-net} & \small{(c) GraphNet} & \small{(d) Fused Lasso} \\
		\footnotesize{(classif. accuracy = $77.0\%$)} &
		\footnotesize{(classif. accuracy = $77.0\%$)} &
		\footnotesize{(classif. accuracy = $85.6\%$)} &
		\footnotesize{(classif. accuracy = $88.2\%$)} \\
	\end{tabular} \vspace{0pt}\\
	\renewcommand{\imwidth} {0.4\linewidth}	
	\begin{subfigure}[b]{\imwidth}\setcounter{subfigure}{4}
		\centering
		\includegraphics[height=95pt]{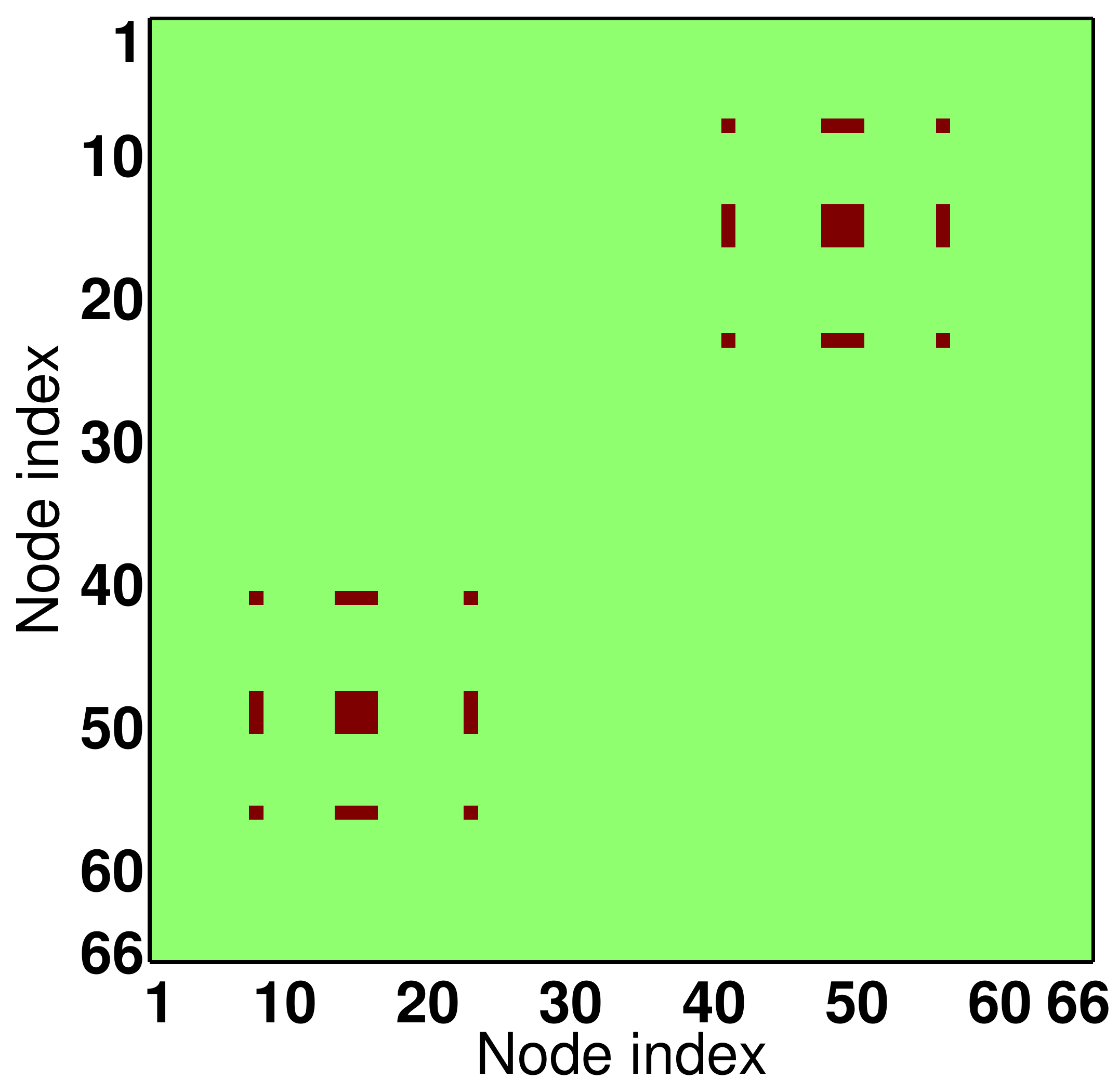}
		\vspace{-4pt}\\
		\caption{Support of the \emph{anomalous edges}}
		\label{subfig:sim,edge,truth}
	\end{subfigure}
	\hspace{10pt}
	\begin{subfigure}[b]{\imwidth}
		\centering
		\includegraphics[height=95pt]{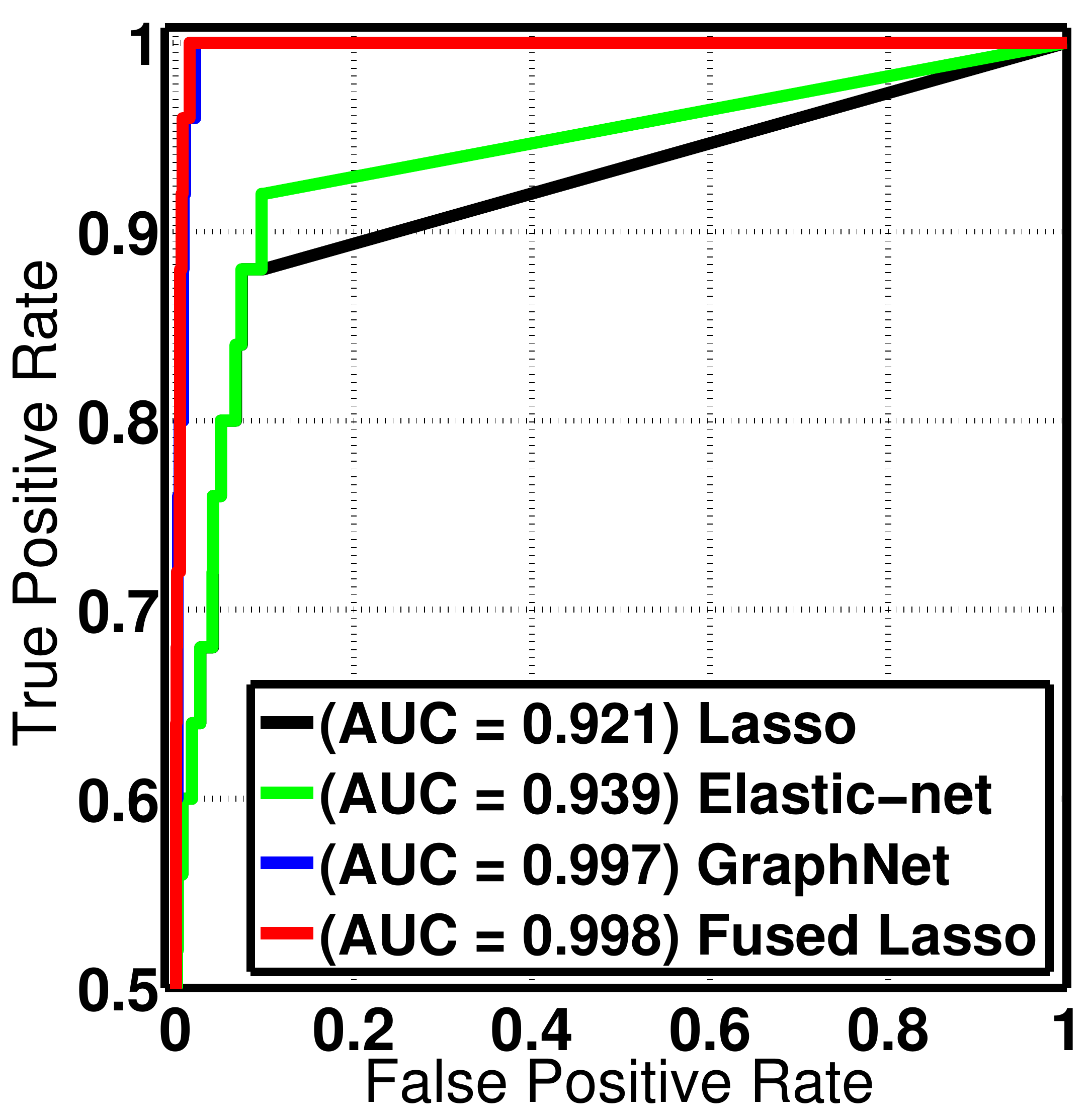}
		\vspace{-4pt}\\
		\caption{ROC (edge identification accuracy)}
		\label{subfig:sim,roc}
	\end{subfigure}
	\vspace{-5pt}\\
	\caption{
		Simulation experiment result: training set consists of $n=100$ samples with $50$ patients and $50$ controls (best viewed in color).  
		\mbox{(a)-(d) Weight} vectors (reshaped into symmetric matrices) estimated from solving the regularized ERM problem~\eqref{eqn:reg,erm} using the hinge-loss and four different regularizers.
		Regularization parameters were tuned via $5$-fold cross-validation on the training set, and classification accuracies were evaluated on a testing set consisting of $500$ samples with $250$ patients and $250$ controls.
		(e) Support matrix indicating the locations of the anomalous edges.
		(f) ROC curve representing the anomalous edge identification accuracy (not classification accuracy) of the four regularizers.
	}
	\label{fig:sim,weight,result}
	\vspace{10pt}
	\renewcommand{\imwidth}  {0.2449\linewidth}
	\renewcommand{\imheight}  {0.2449\linewidth}
	\setlength{\tabcolsep}{1pt} 
	\begin{tabular}{ccccc}
	\multicolumn{4}{c}{{\textbf{\normalsize{Classification accuracy}}}} \vspace{0pt} \\
	\includegraphics[height=\imheight,width=\imwidth]{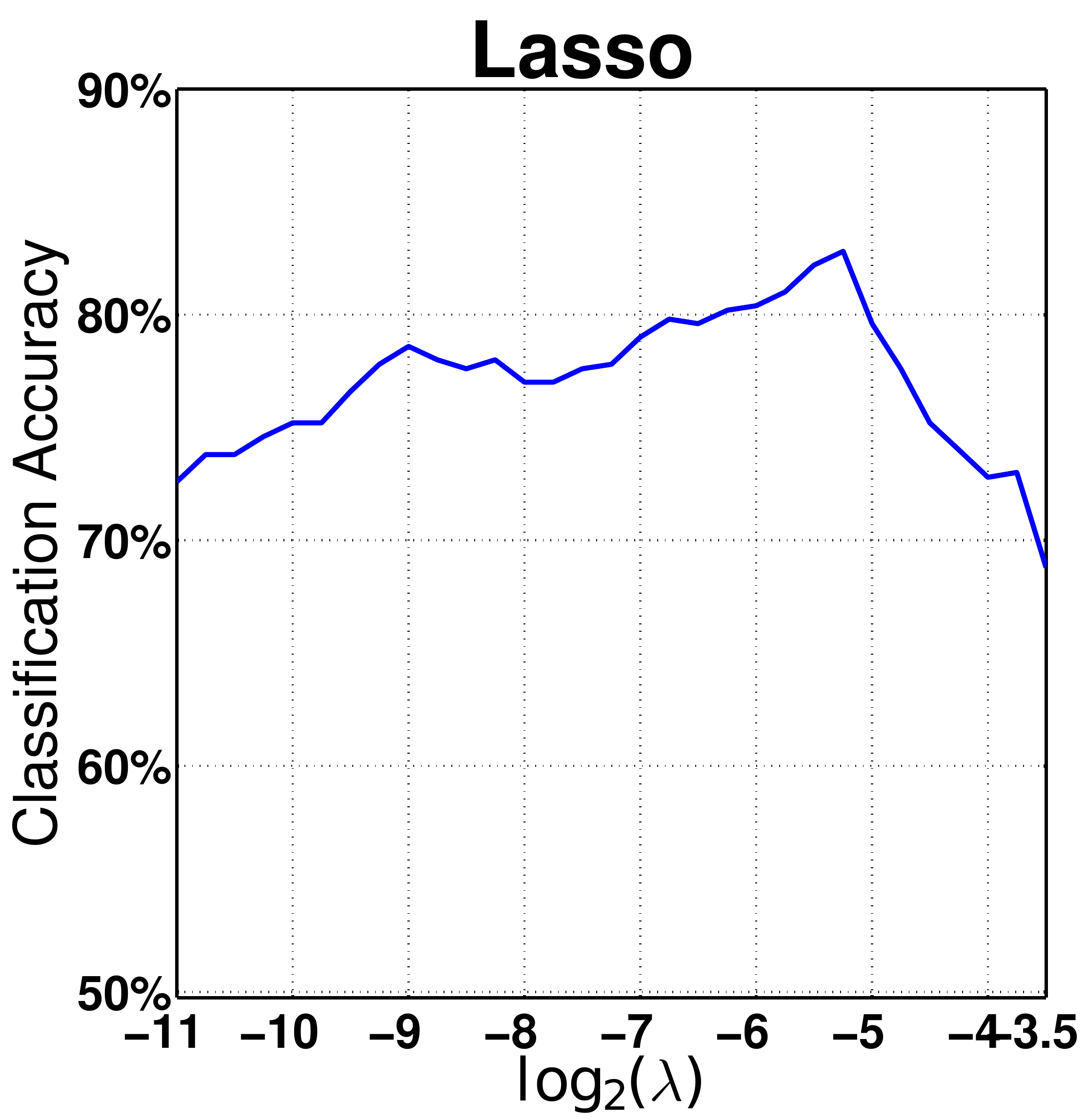} &
	\includegraphics[height=\imheight,width=\imwidth]{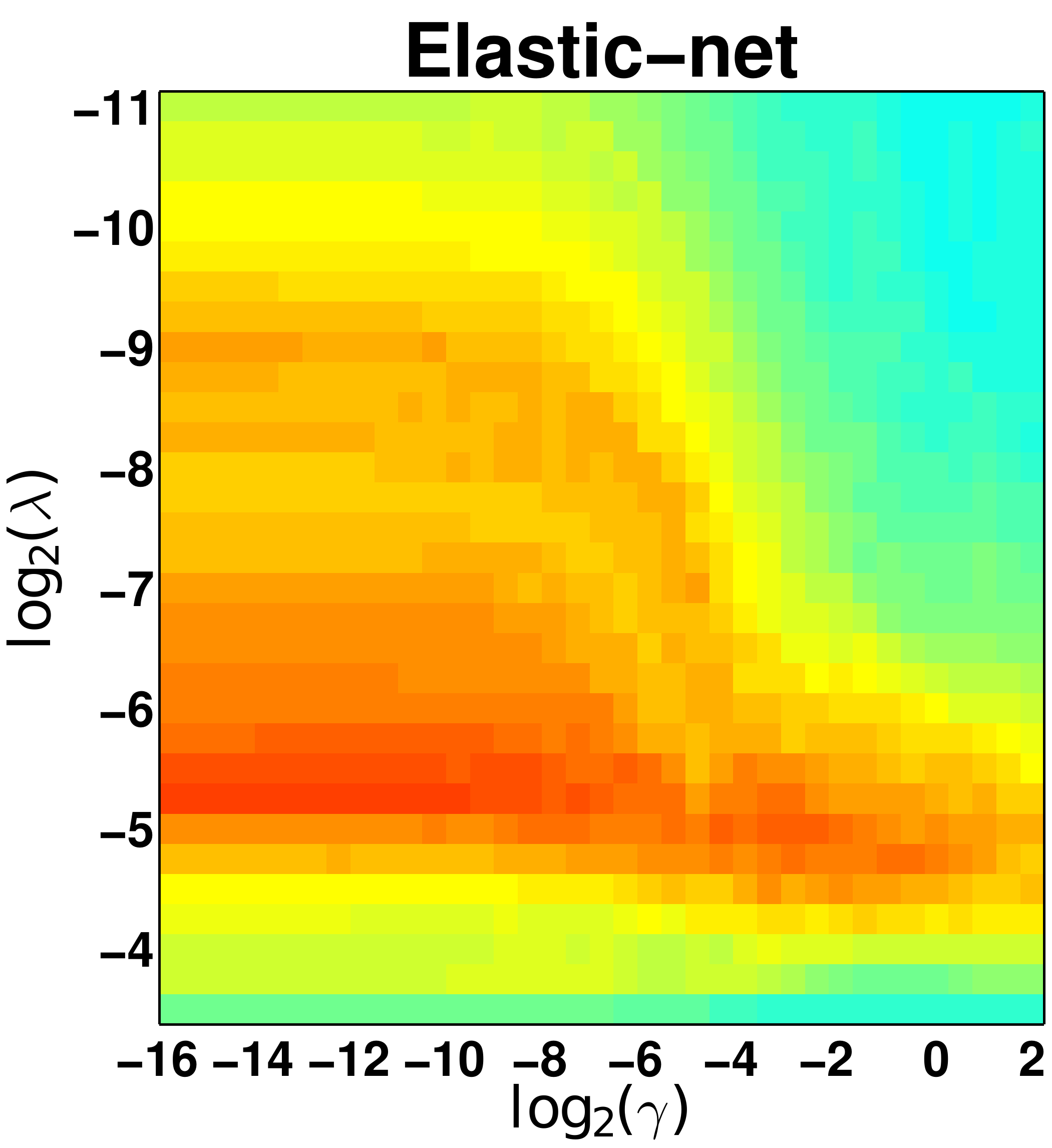} &
	\includegraphics[height=\imheight,width=\imwidth]{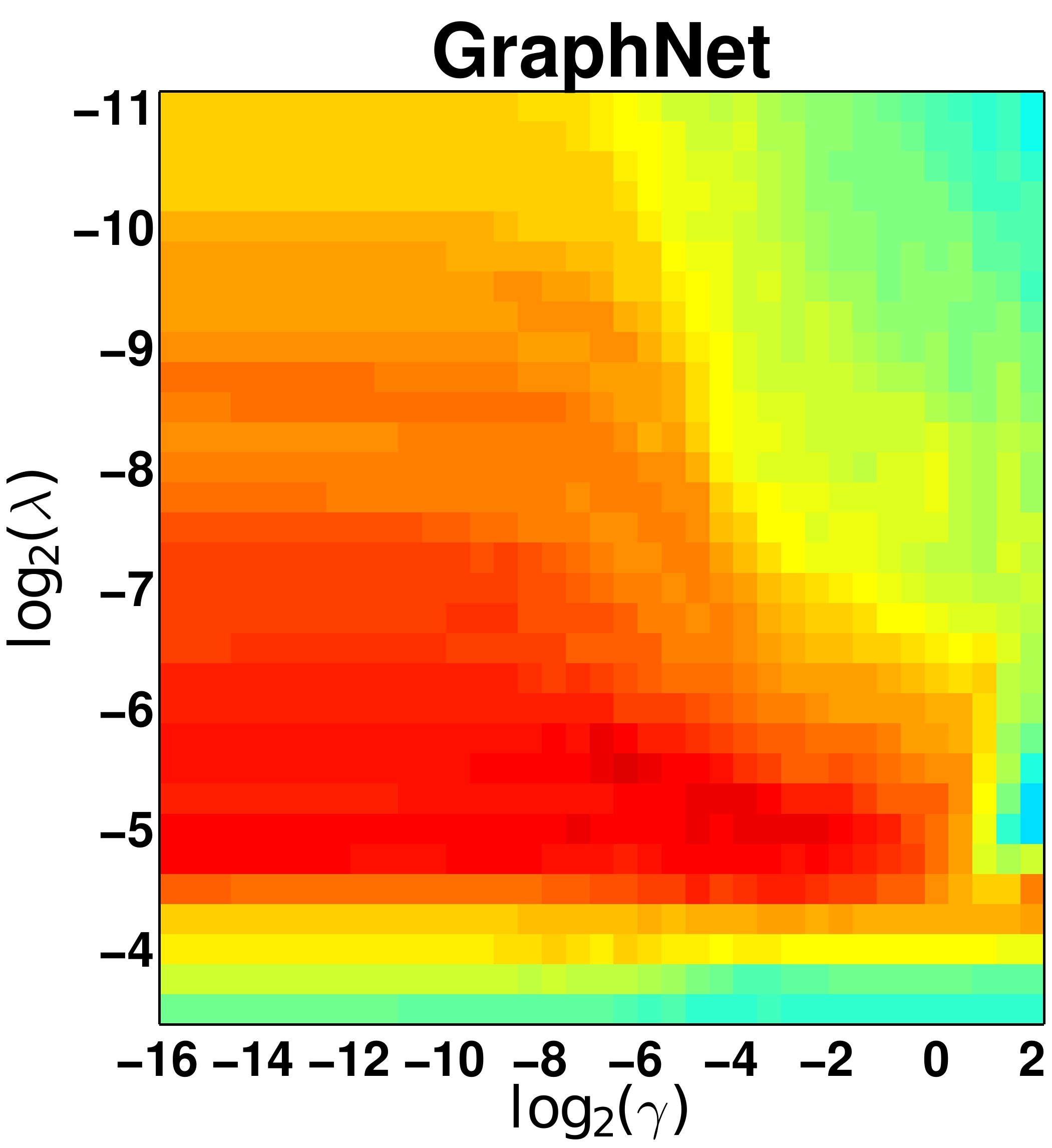} &
	\includegraphics[height=\imheight,width=\imwidth]{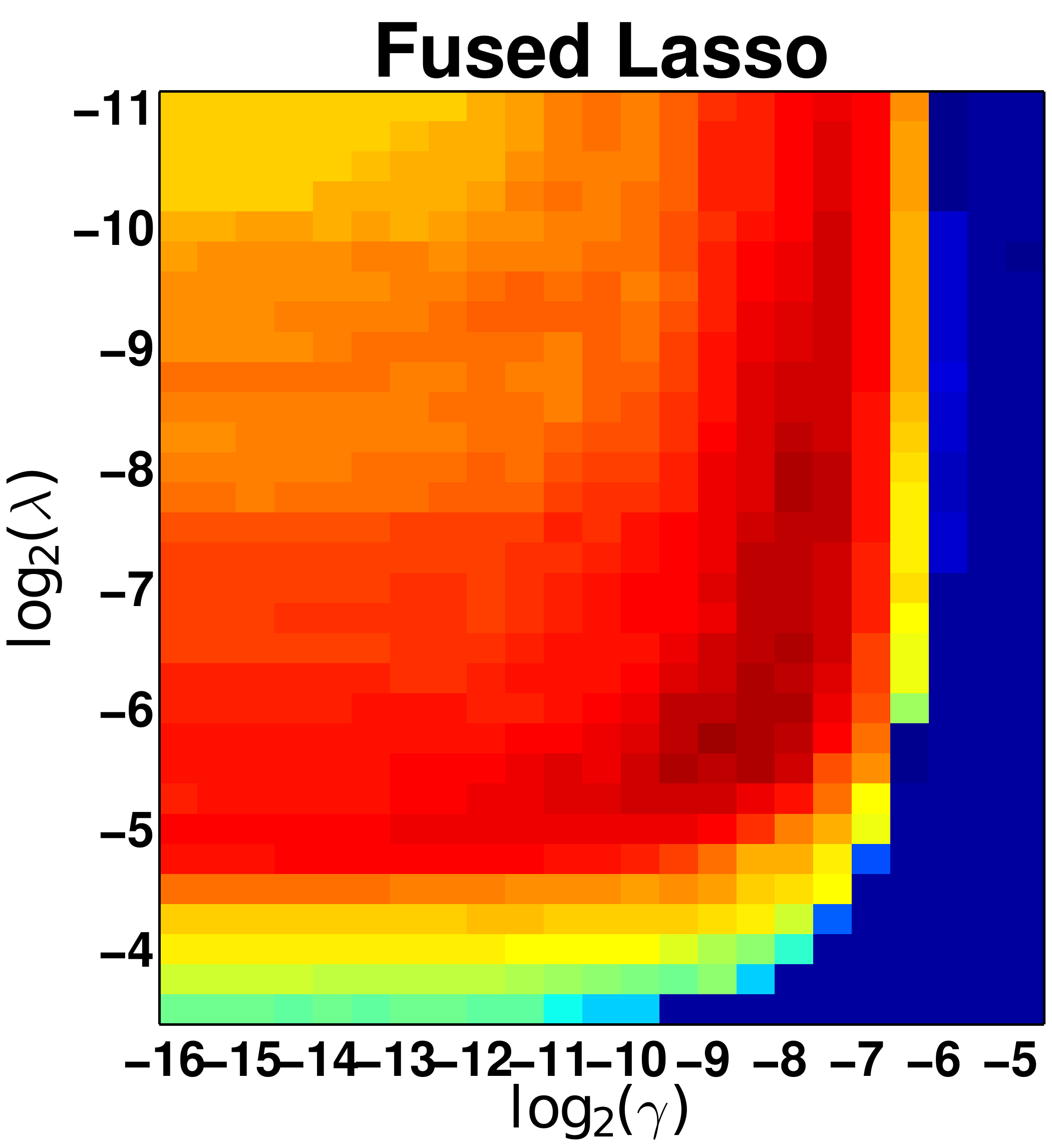} &
	\raisebox{0.02585\linewidth}{\includegraphics[height=0.2035\linewidth]{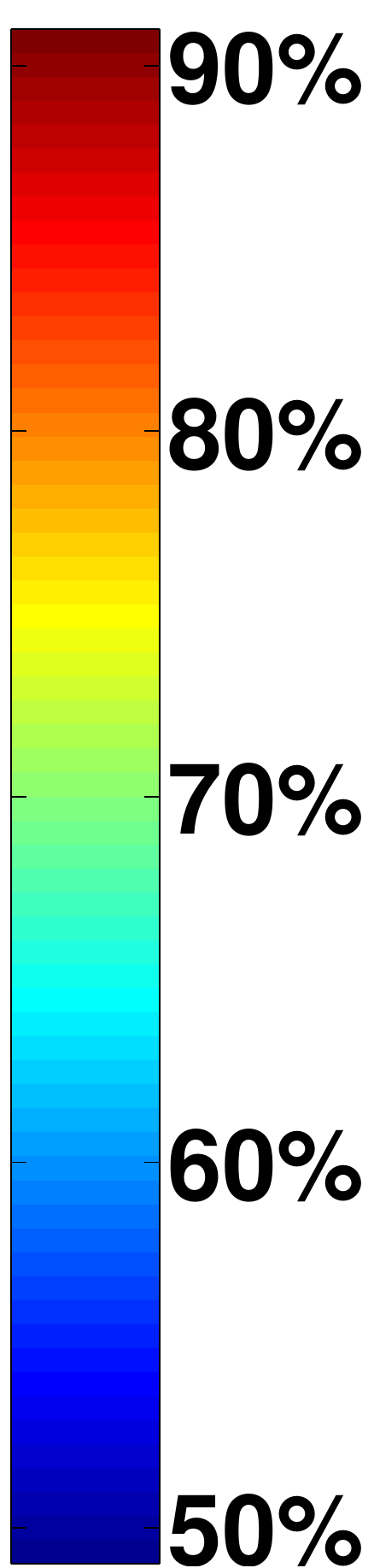}} 
	\vspace{1pt}\\
	\multicolumn{4}{c}{{\textbf{\normalsize{Sparsity level (number of features)}}}} \vspace{0pt}\\
	\includegraphics[height=\imheight,width=\imwidth]{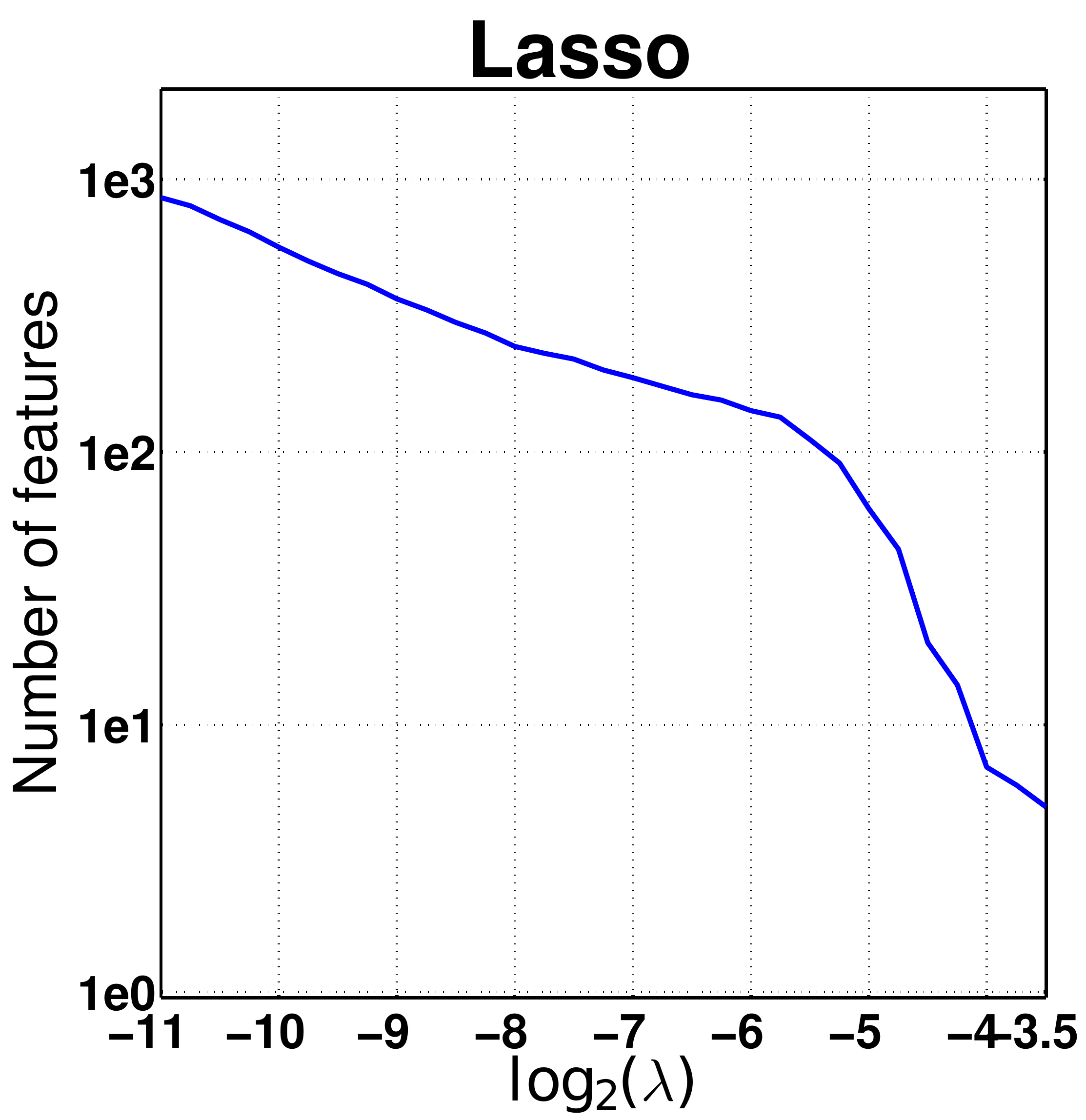} &
	\includegraphics[height=\imheight,width=\imwidth]{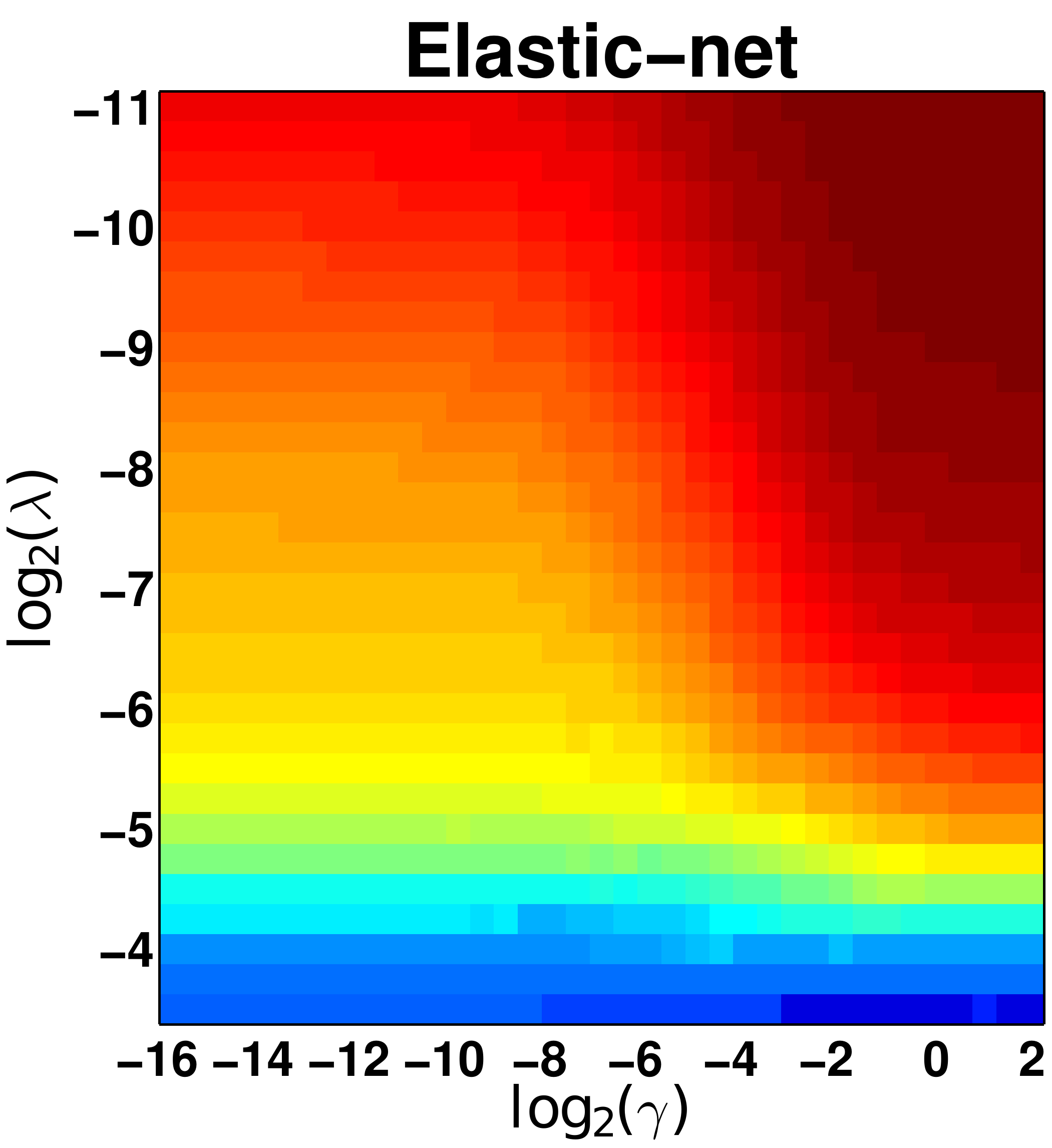} &
	\includegraphics[height=\imheight,width=\imwidth]{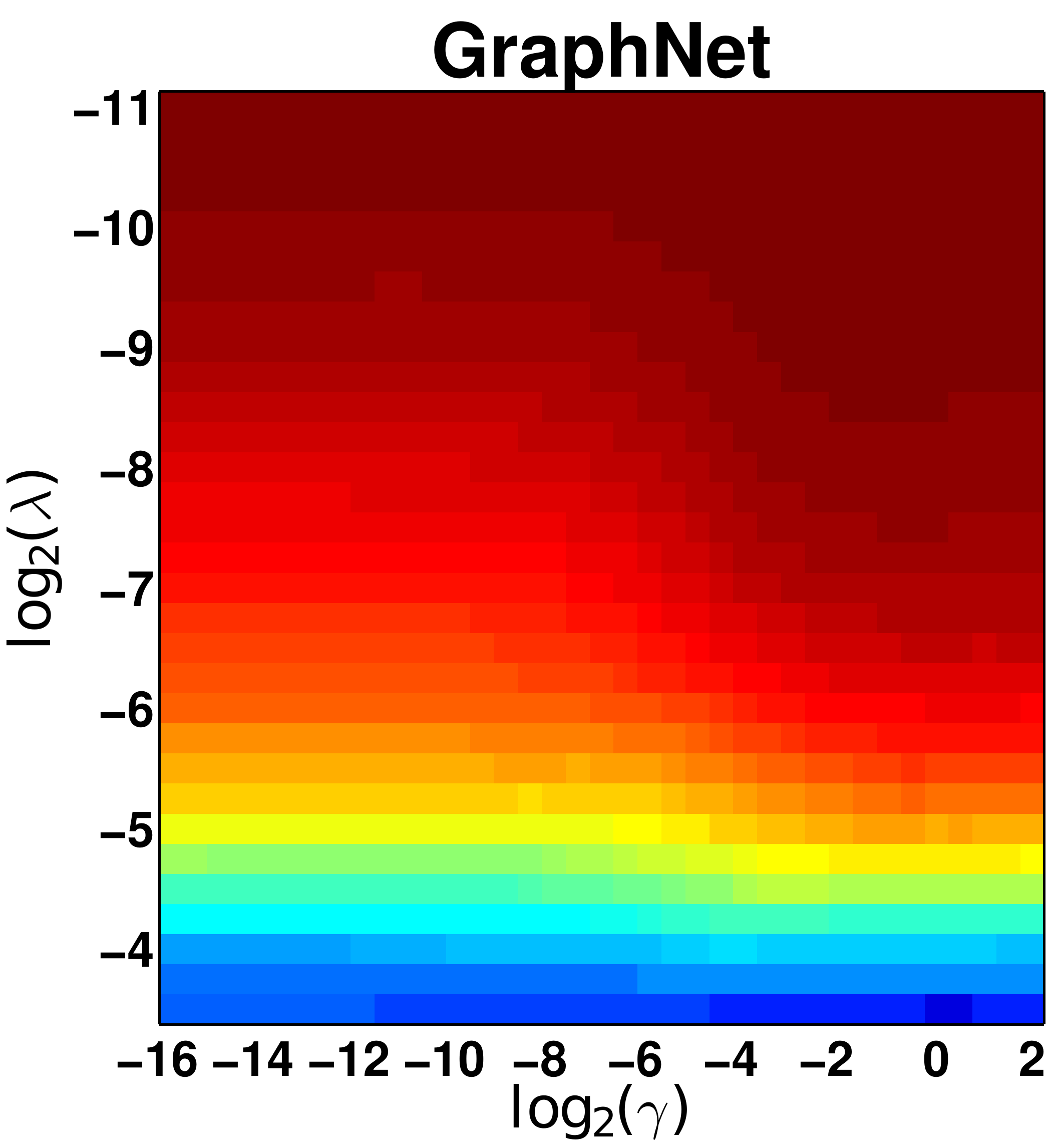} &
	\includegraphics[height=\imheight,width=\imwidth]{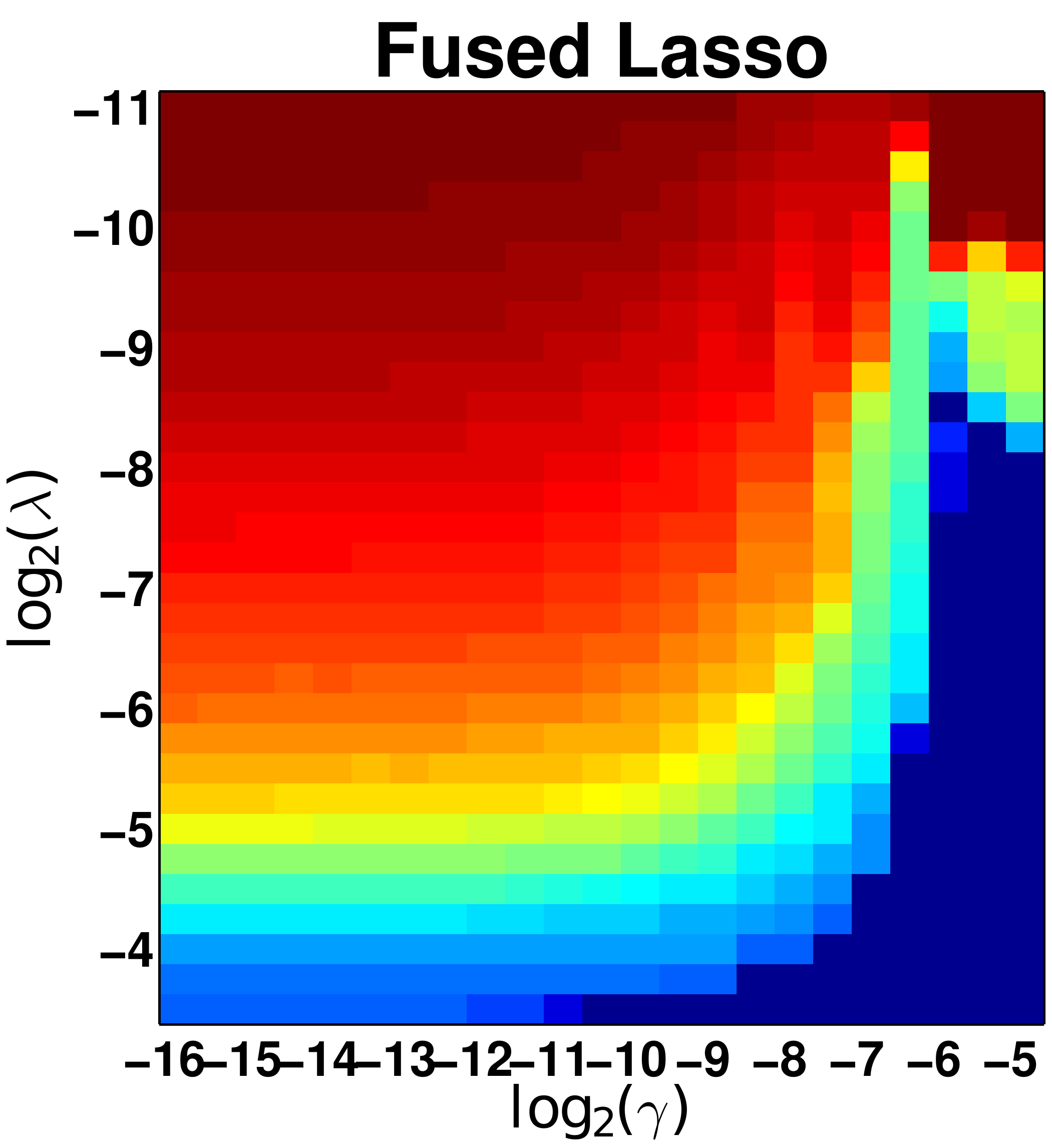} &
	\hspace{0pt}\raisebox{0.021423\linewidth}{\includegraphics[height=0.213\linewidth]{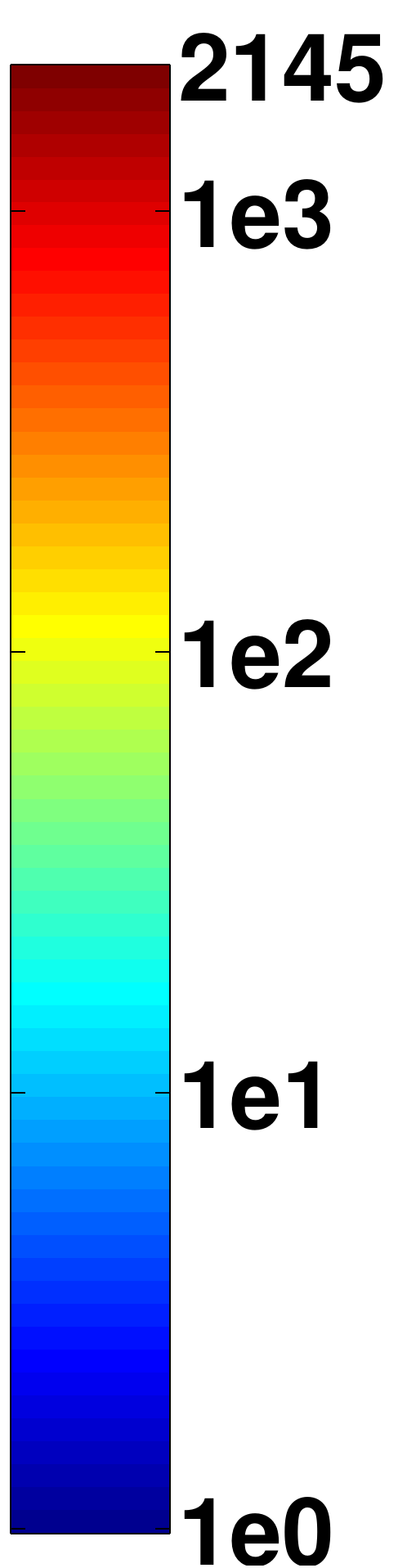}} \\
	\end{tabular}
	\vspace{-10pt}\\
	\caption{Grid search result for the simulation experiment (best viewed in color).
	 	All classifiers were learned using $100$ training samples consisting of $50$ patients and $50$ controls.	 	
		\textbf{Top row}: classification accuracy as a function of the regularization parameters $\{\lambda,\gamma\}$ (evaluated from $500$ testing samples consisting of $250$ patients and $250$ controls).
		\textbf{Bottom row}: the number of features selected as a function of the regularization parameters $\{\lambda,\gamma\}$.
	}
	\label{fig:sim,grid,ntr}
\end{figure}
The top row of Fig.~\ref{fig:sim,weight,result} displays the estimated weight vectors, and the corresponding testing classification accuracies are reported under the subcaptions. 
Here, the fused Lasso regularized SVM yielded the best classification accuracy at $88.2\%$ using $92$ features, followed by $85.6\%$ from GraphNet which used $104$ features; Lasso and Elastic-net both achieved $77.0\%$ classification accuracy using $230$ and $232$ features respectively.
However, a perhaps more interesting observation is that fused Lasso and GraphNet were able to recover the structure of the \emph{anomalous edges} much more clearly than Lasso and Elastic-net; this can be seen by comparing the weight vectors estimated by the four regularizers with the support of the anomalous edges displayed in Fig.~\ref{subfig:sim,edge,truth}.
While Lasso and Elastic-net yielded weight vector estimates with salt-and-pepper patterns that are difficult to interpret, the weight vector estimates for fused Lasso and GraphNet closely resembles the  structure of the \emph{anomalous edges}.
To quantify the regularizers' ability to identify the discriminative edges, we generated a receiver operating characteristic (ROC) curve by thresholding the absolute value of the elements of the estimated weight vector.
The resulting ROC curve for the four regularizers are plotted in Fig.~\ref{subfig:sim,roc}; we emphasize that this ROC curve summarizes the regularizers' ability to identify the informative edges, and does not represent classification accuracy.
From this ROC curve, we see that fused Lasso and GraphNet attain the best performances, achieving a nearly perfect \emph{area under the curve} (AUC) value of $0.998$ and $0.997$ respectively, whereas the AUC value for Lasso and Elastic-net were $0.921$ and $0.939$ respectively.
In short, Fig.~\ref{fig:sim,weight,result}a-f demonstrate that fused Lasso and GraphNet not only improved classification accuracy, but also exhibited superior performance in recovering the discriminatory edges with respect to their non-spatially informed counterparts, Lasso and Elastic-net.

In our next analysis, we studied how classification accuracy and sparsity (\ie, number of features selected) behave as a function of the regularization parameters $\{\lambda,\gamma\}$.
For this, we conducted a grid search over the same range of $\lambda$ and $\gamma$ values presented above, but the classifiers were trained over the entire training set.
Classification accuracy was evaluated on the same testing set as the above experiment.
The result of the grid search is presented in Fig.~\ref{fig:sim,grid,ntr}, where the top row plots the testing classification accuracy and the bottom row plots the number of features selected, both as a function of the regularization parameters $\{\lambda,\gamma\}$.

\begin{figure}[!t]
\noindent
\begin{minipage}{\textwidth}
	\centering
	\renewcommand{\imwidth}  {0.35\linewidth}
	\includegraphics[width=\imwidth]{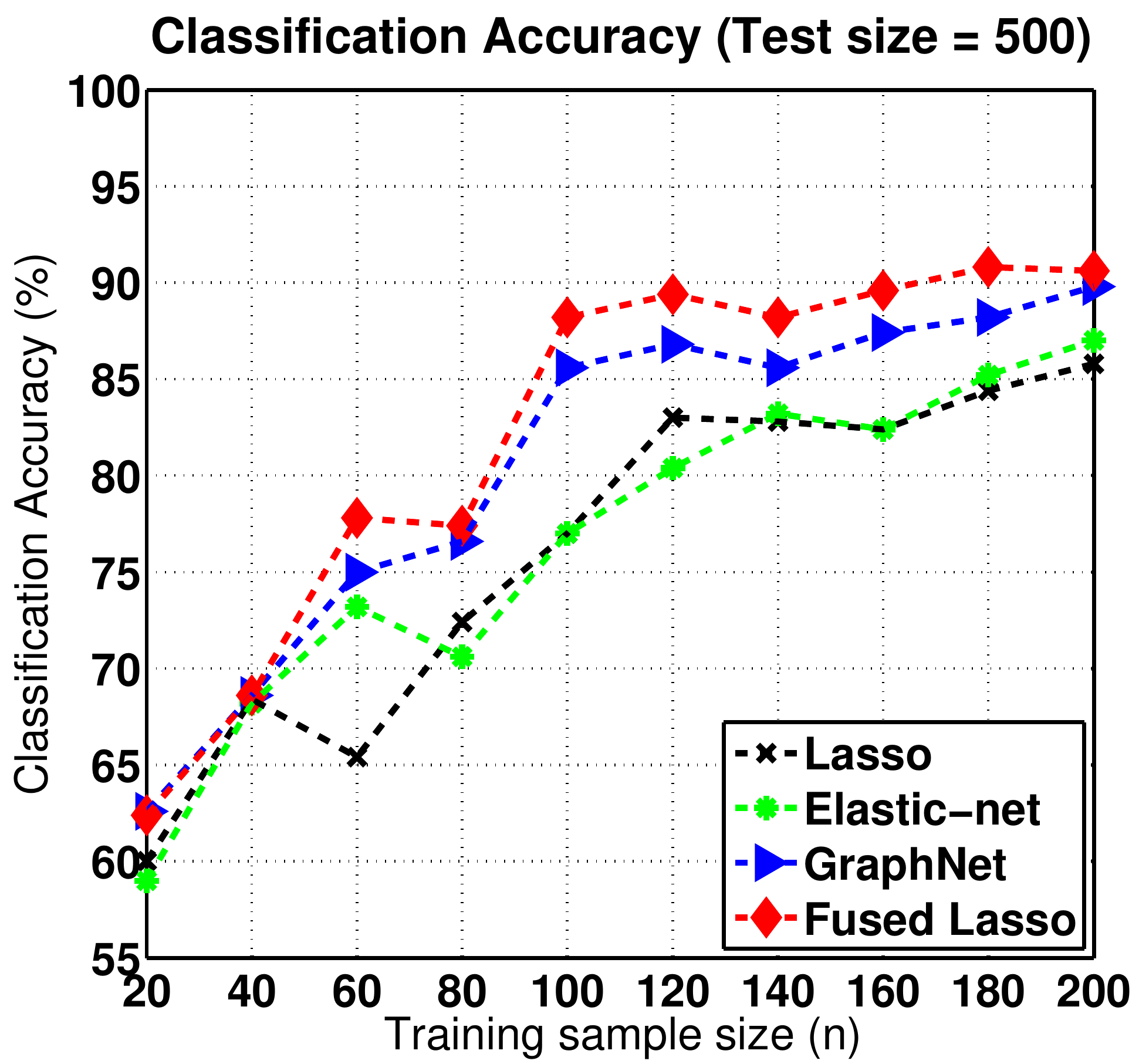} 
	\captionof{figure}{
		The testing classification accuracy of the different regularizers as a function as a number of training samples~$n$ in the simulation experiment.  
		Regularization parameters were tuned via $5$-fold cross-validation on the training set.  
		The testing set consists of $500$ samples with $250$ patients and $250$ controls.
		Table~\ref{table:sim,acc} reports the actual numbers.
	}
	 \label{fig:sim,acc,plot}
\vspace{10pt}
	\setlength{\tabcolsep}{4.5pt}  
	\begin{tabular}{c||c|c|c|c|c|c|c|c|c|c}
		\multicolumn{1}{l}{} &	\multicolumn{10}{c}{Testing Classification accuracy ($n$ = training sample size, $500$ = test size)}\\
		\hline
		\textbf{\small{Regularizer}} &
			\small{$n$=20} & \small{$n$=40} &\small{$n$=60}& \small{$n$=80} & \small{$n$=100} & \small{$n$=120} & \small{$n$=140} & \small{$n$=160} &\small{$n$=180} & \small{$n$=200} \\
		\hline\hline
		\textbf{\small{Lasso}} &
			\small{60.0\%}& \small{68.4\%}& \small{65.4\%}& \small{72.4\%}& \small{77.0\%}& \small{83.0\%}& \small{82.8\%}& \small{82.4\%}& \small{84.4\%}& \small{85.8\%}\\
		\textbf{\small{Elastic-net}} & 
			\small{59.7\%}& \small{68.2\%}& \small{73.2\%}& \small{70.6\%}& \small{77.0\%}& \small{80.4\%}& \small{83.2\%}& \small{82.4\%}& \small{85.2\%}& \small{87.0\%}\\
		\textbf{\small{GraphNet}} &
			\textbf{\small{62.6\%}}& \textbf{\small{68.6\%}}& \small{75.0\%}& \small{76.6\%}& \small{85.6\%}& \small{86.8\%}& \small{85.6\%}& \small{87.4\%}& \small{88.2\%}& \small{89.8\%}\\
		\textbf{\small{Fused Lasso}} &
			\small{62.4\%}& \textbf{\small{68.6\%}}& \textbf{\small{77.8\%}}& \textbf{\small{77.4\%}}& \textbf{\small{88.2\%}}& \textbf{\small{89.4\%}}& \textbf{\small{88.2\%}}& \textbf{\small{89.6\%}}& \textbf{\small{90.8\%}}& \textbf{\small{90.6\%}}\\
		\hline
	\end{tabular}
	\captionof{table}{
		The testing classification accuracy of the different regularizers as a function as a number of training samples~$n$ in the simulation experiment (the best classification accuracy for each $n$ is denoted in bold font).
		See Fig.~\ref{fig:sim,acc,plot} for a plot of this result.
	}
	\label{table:sim,acc}
\end{minipage}
\end{figure}

To further study the performance of our method, we next conducted a \emph{sample complexity analysis} \citep{Gramfort:2011}, where we studied how the classification accuracy of the four regularizers behaved as a function of the training sample size $n$.
This was done by repeating our earlier experiment of tuning the regularization parameters via $5$-fold cross-validation on the training set, but here we varied the training sample size over the range $n\in\{20,40,60,\dots,200\}$; the same testing set of size $500$ was used throughout for evaluating the classification accuracy.
Note the labels are balanced for all datasets, \ie, the training set consists of $n/2$ patients and $n/2$ controls, and similarly the testing set consists of $250$ patients and $250$ controls.
The result of this experiment is reported in Fig.~\ref{fig:sim,acc,plot} and Table~\ref{table:sim,acc}.
A key observation from this analysis is that the classification accuracy for GraphNet and fused Lasso consistently outperformed Lasso and Elastic-net, which can be attributed to the spatial information injected by these spatially-informed regularizers.
Overall, fused Lasso yielded the best classification accuracy.

It is important to note that the inclusion of the anomalous node clusters in the data generating process certainly favors fused Lasso and GraphNet.
However, we remind the readers that these anomalous node clusters are not some arbitrary structures we introduced to favor the spatially-informed regularizers, but are motivated from the ``patchiness assumption'' of brain disorders, a neuroscientific viewpoint which we discuss in detail in Sec.~\ref{subsec:why,flasso}.
The results from the simulation experiments confirm the intuition that if the ``patchiness assumption'' of brain disorders holds true, spatially-informed classifiers can be a powerful tool for recovering relevant biosignatures.

\subsection{Results on resting state fMRI data from a schizophrenia dataset}
\label{subsec:result,real,data}
In this experiment, we examined the performance of linear classifiers trained using regularized ERM \eqref{eqn:reg,erm} with the hinge-loss, and three regularizers were subject to comparison: Elastic-net, GraphNet, and fused Lasso.  
The study involved $121$ participants, consisting of $54$ schizophrenic subjects (SZ) and $67$ healthy controls (HC).
We adopt the convention of letting $y=+1$ indicate SZ and $y=-1$ indicate HC subjects.
The ADMM algorithm was terminated when the tolerance level~\eqref{eqn:admm,termin} fell below $\varepsilon=4\times 10^{-3}$ or the algorithm reached $400$ iterations.
Empirically, we found the algorithm to converge at around $180\mytilde 300$ iterations.
For the two regularization parameters, we conducted a two-dimensional grid search: the \ellone regularization parameter $\lambda\geq 0$ was searched over the range $\lambda\in\{2^{-20},2^{-19},\cdots,2^{-3}\}$ for all three regularizers, and the second regularization parameter $\gamma\geq 0$ was searched over $\gamma\in\{2^{-20},2^{-19},\cdots,2^{3}\}$ for Elastic-net and GraphNet and $\gamma\in\{2^{-20},2^{-19},\cdots,2^{-3}\}$ for fused Lasso.
Ten-fold cross-validation to evaluate the generalizability of the classifiers.
Furthermore, we analyzed the sparsity level achieved during the grid search by computing the average number of features selected across the cross-validation folds.

The resulting testing classification accuracy and sparsity level for different combinations of $\{\lambda,\gamma\}$ are rendered as heatmaps in Fig.~\ref{fig:grid,search}.
The general trend observed from the grid search is that for all three regularization methods, the classification accuracy improved as more features entered the model.
We observed the same trend when using other loss functions as well, specifically the truncated-least squares loss and the huberized-hinge loss (using $\delta=0.5$) function.
Although this behavior may be somewhat surprising, it has been reported that in the $p\gg n$ setting, the unregularized SVM often performs just as well as the best regularized case, and accuracy can degrade when feature pruning takes place (see Ch.18 in \cite{Hastie:2009:book}).

\newcommand{\addhspace}{\hspace{0.04\linewidth}}
\begin{figure}[t]
	\setlength{\tabcolsep}{1pt} 
	\renewcommand{\imwidth}  {0.3\linewidth}
	\renewcommand{\imheight}  {0.3105\linewidth}	
	\begin{tabular}{cccc}	
	\multicolumn{4}{c}{{\textbf{\large{Classification accuracy}}}} \vspace{0pt} \\
	\includegraphics[width=\imwidth,height=\imheight]{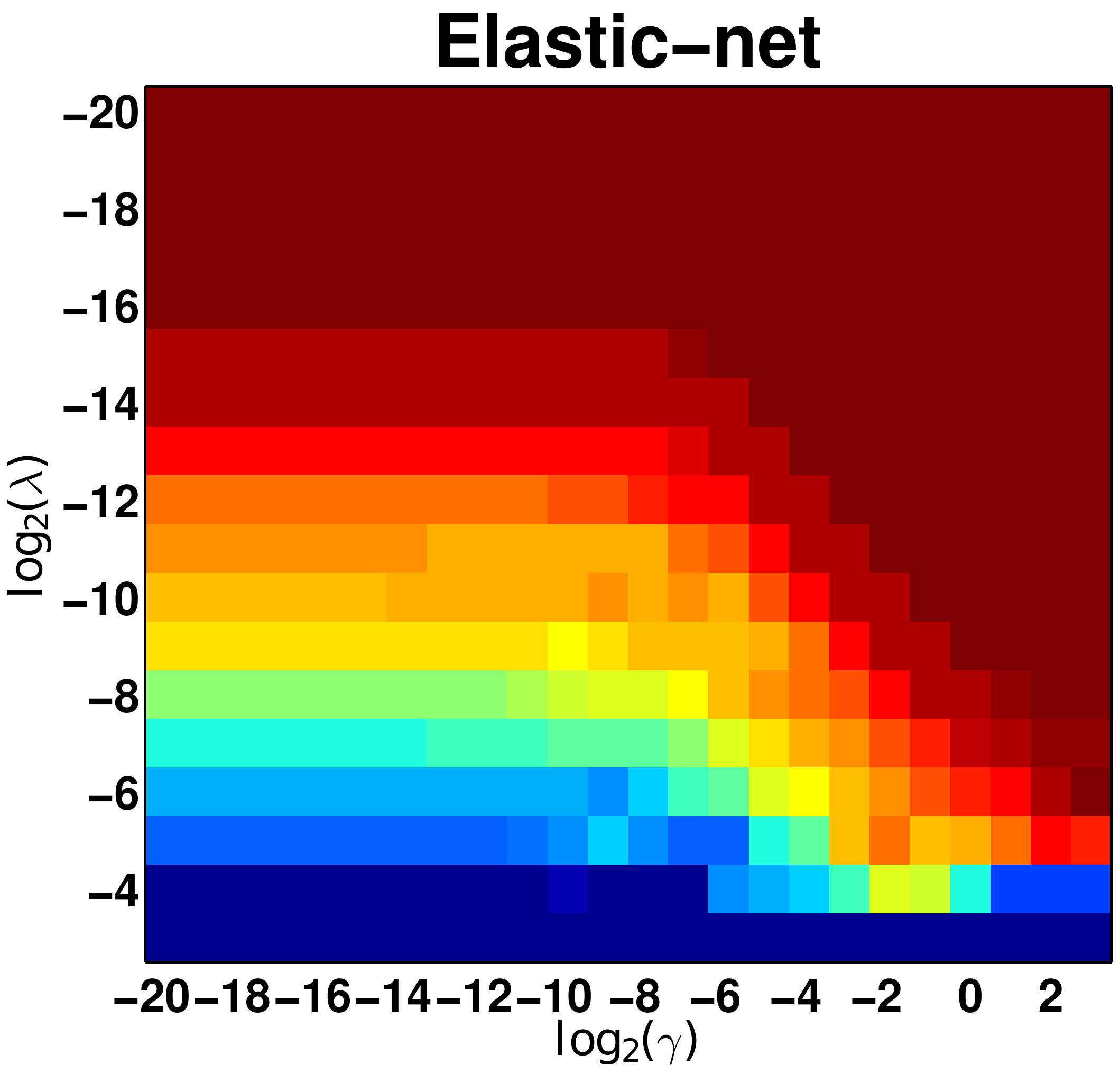} &
	\includegraphics[width=\imwidth,height=\imheight]{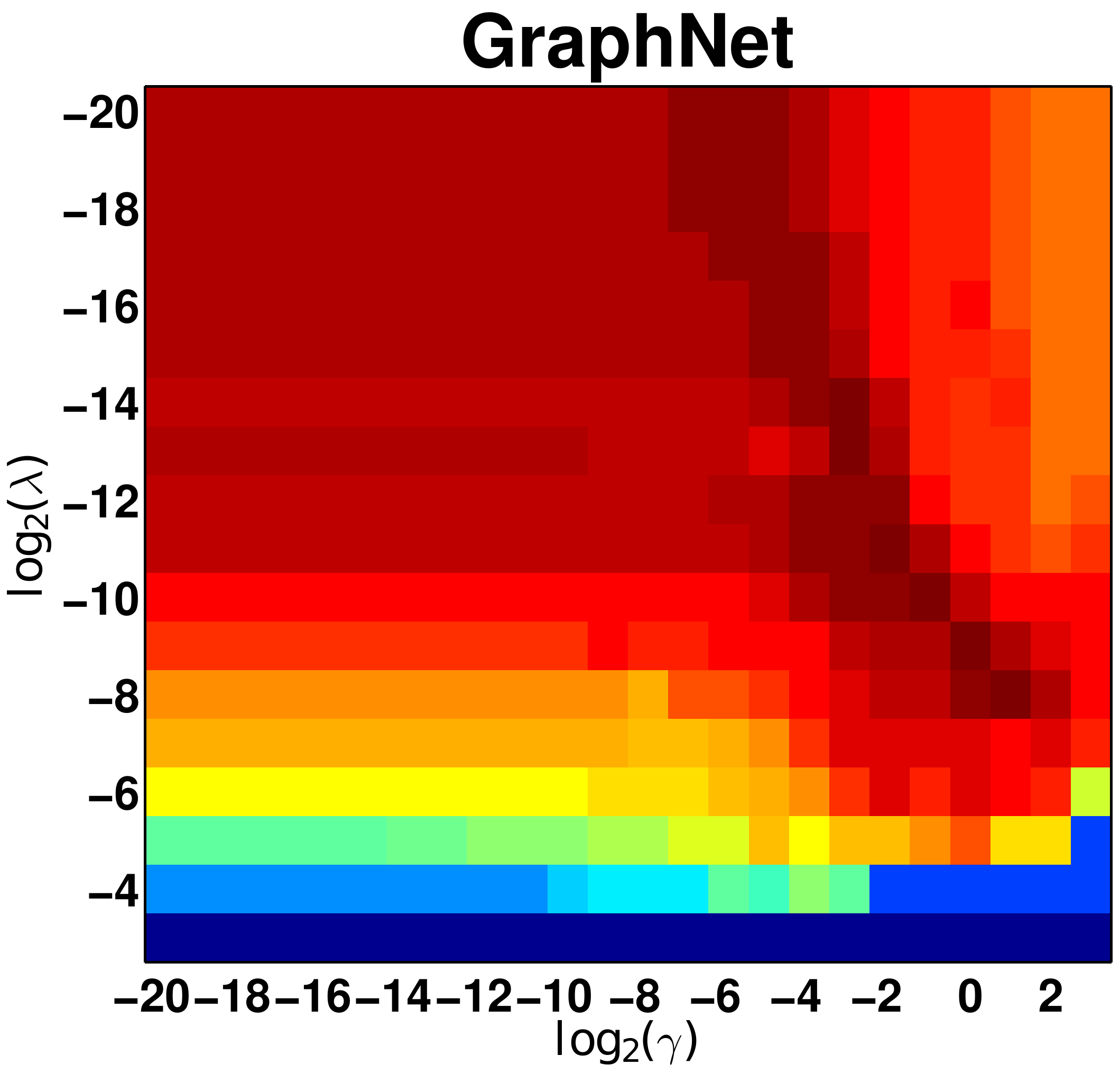} &
	\includegraphics[width=\imwidth,height=\imheight]{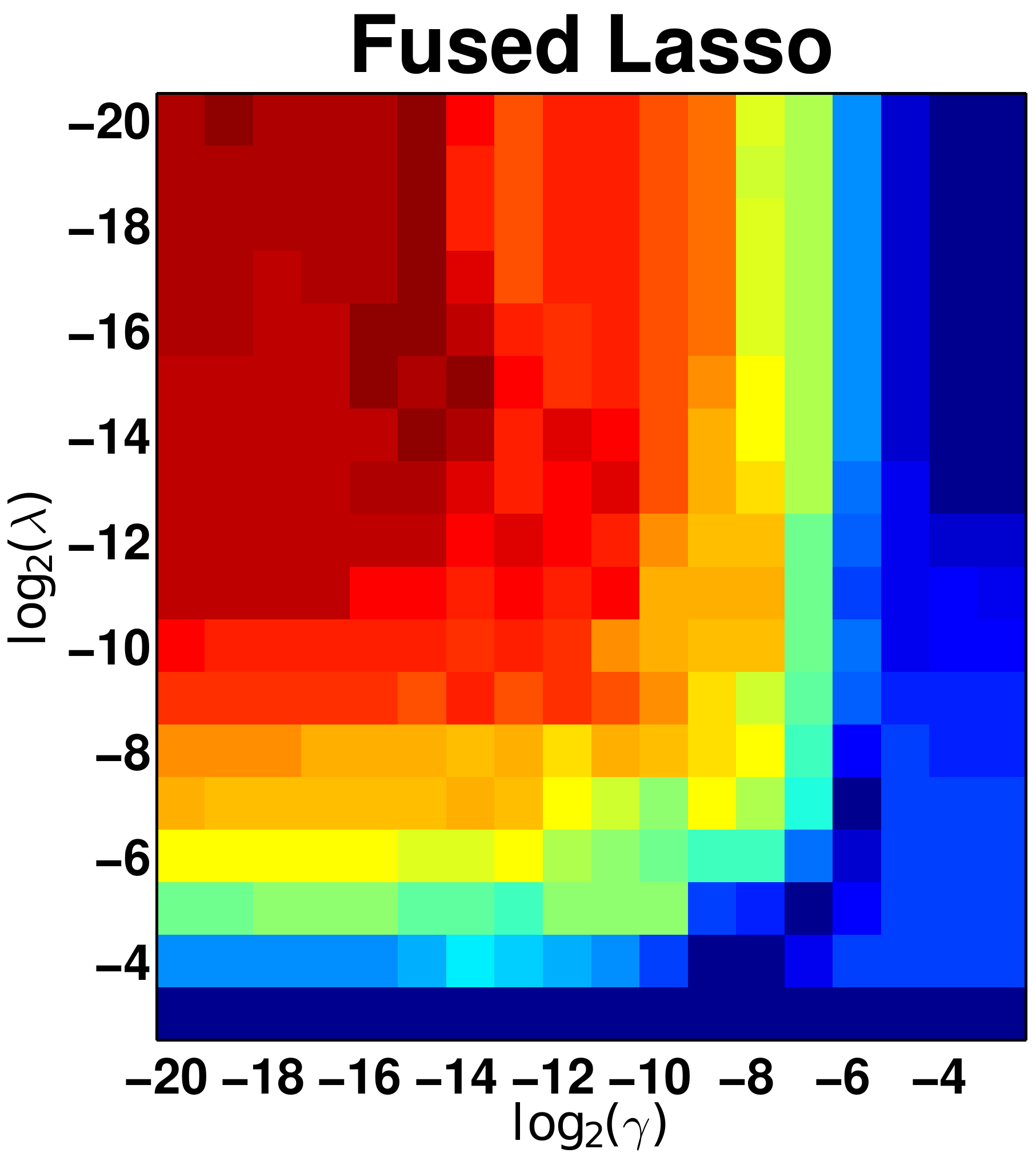} &
	\raisebox{0.02725\linewidth}{\includegraphics[height=0.268\linewidth]{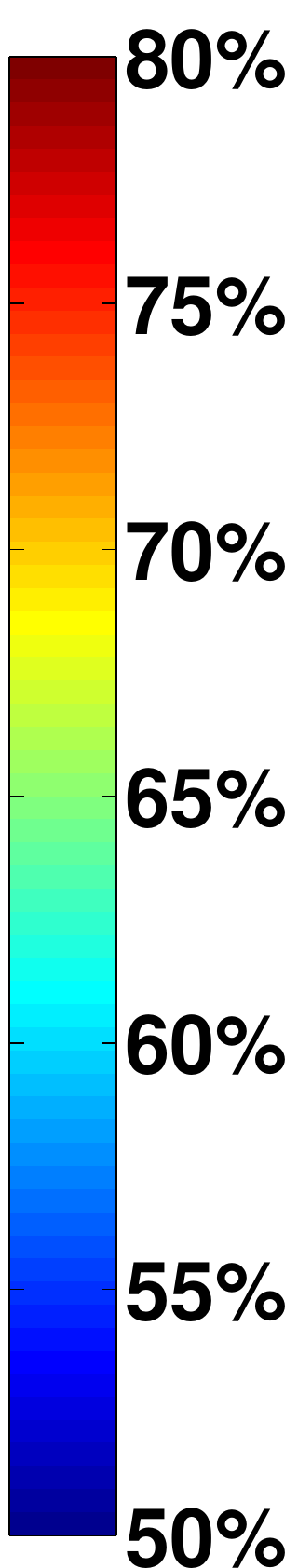}} \vspace{9pt}\\
	\multicolumn{4}{c}{{\textbf{\large{Mean sparsity level (number of features)}}}} \vspace{0pt}\\
	\includegraphics[width=\imwidth,height=\imheight]{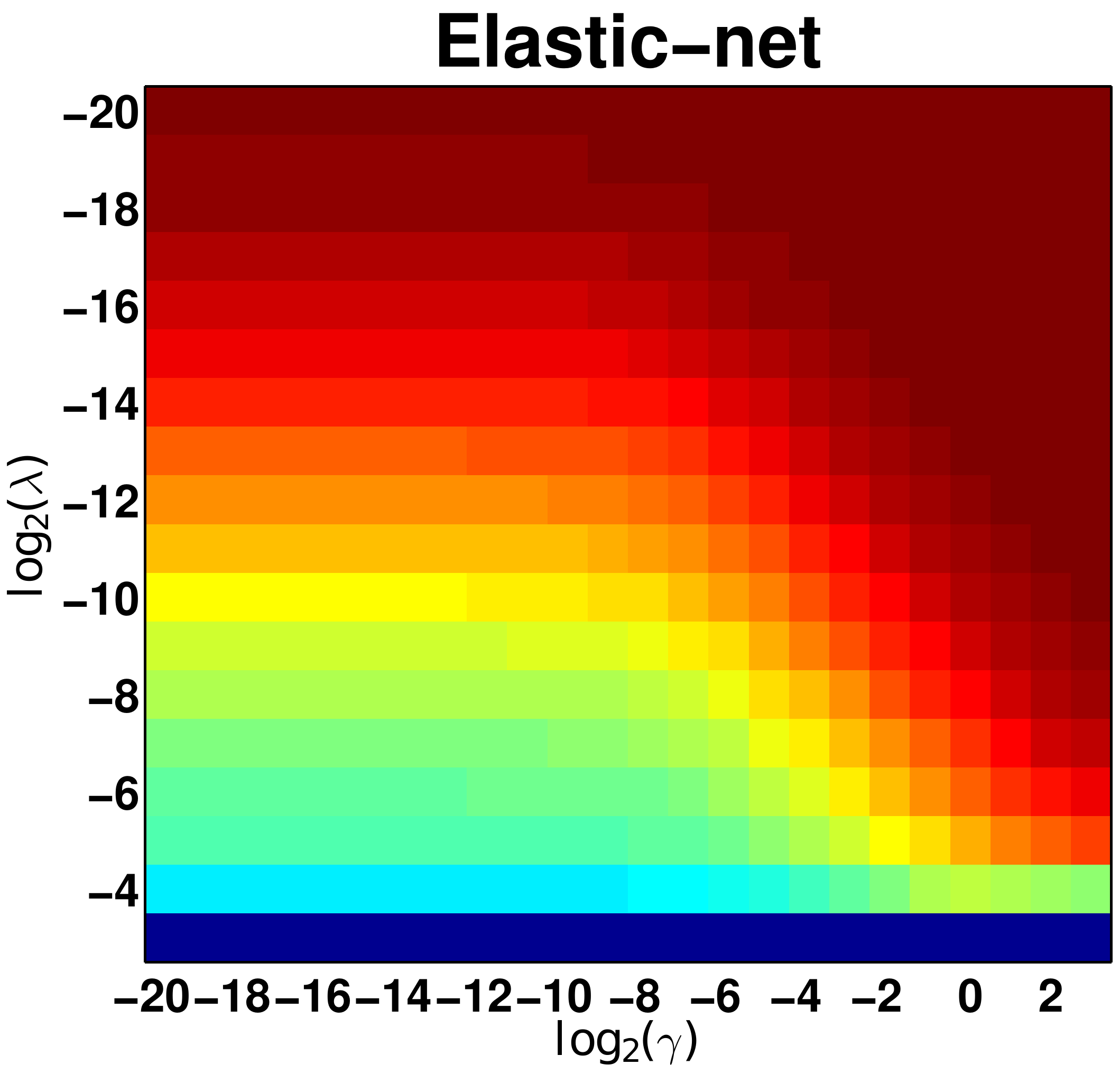} &
	\includegraphics[width=\imwidth,height=\imheight]{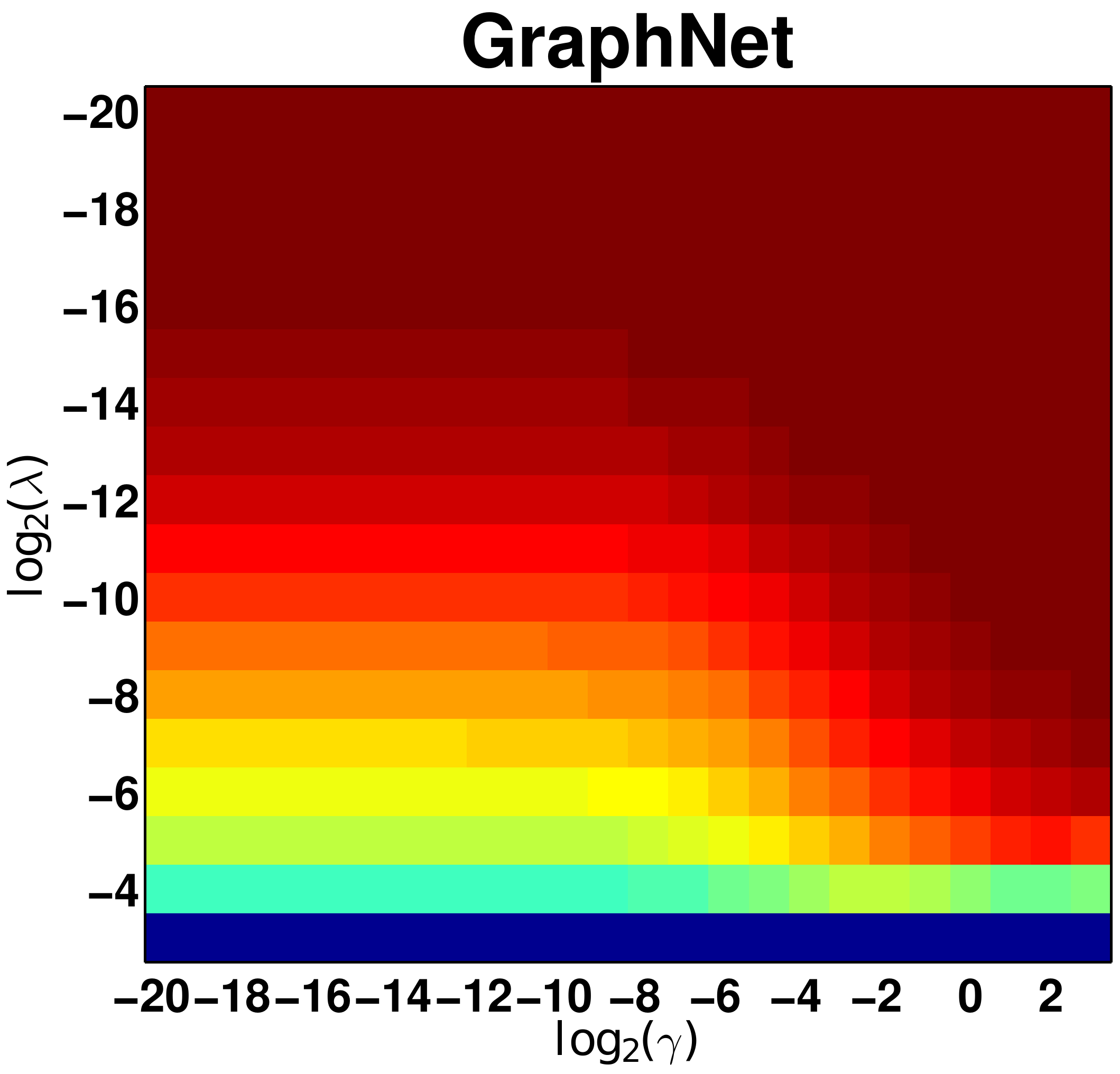} &
	\includegraphics[width=\imwidth,height=\imheight]{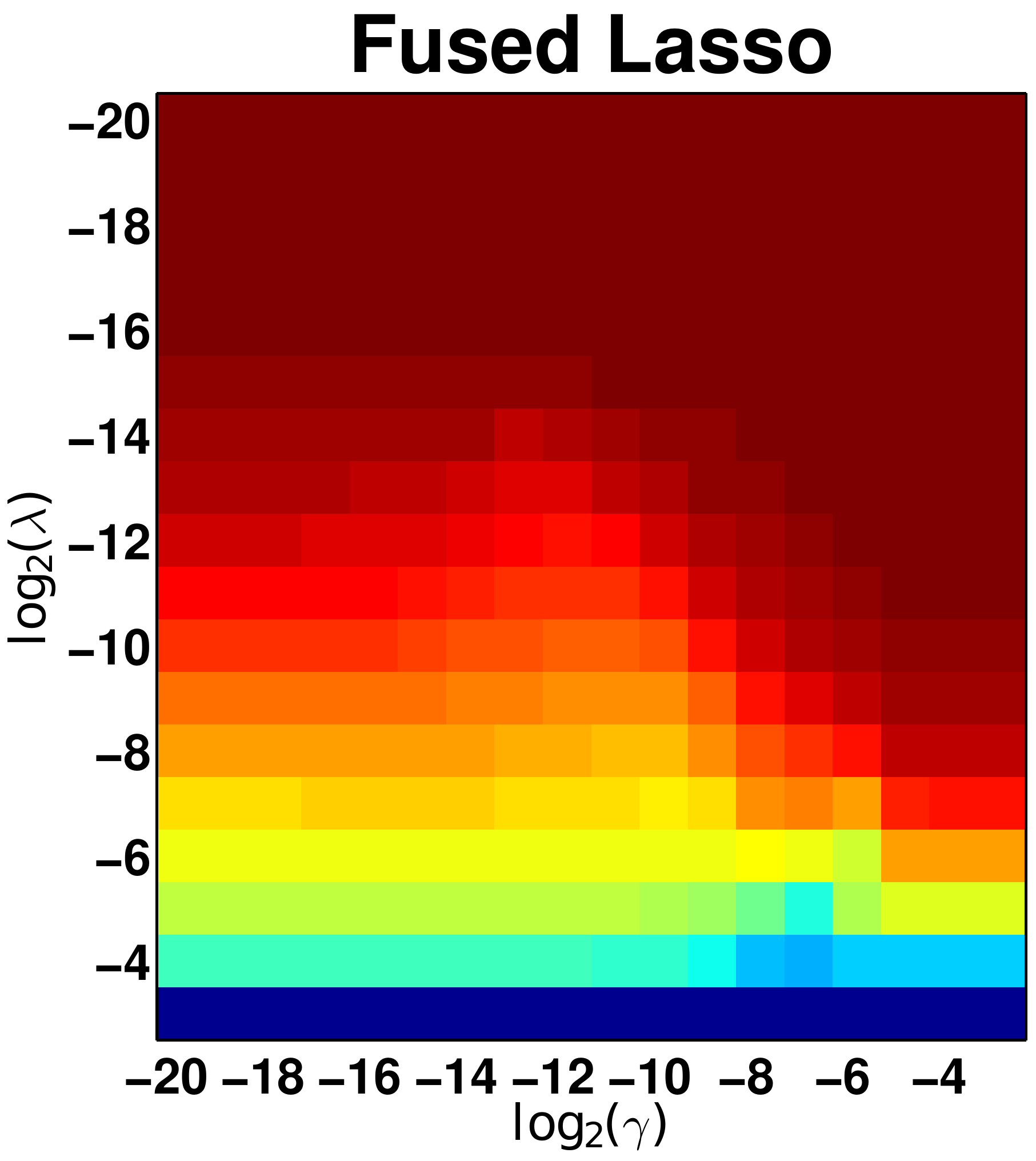} &
	\hspace{-4pt}\raisebox{0.03225\linewidth}{\includegraphics[height=0.26\linewidth]{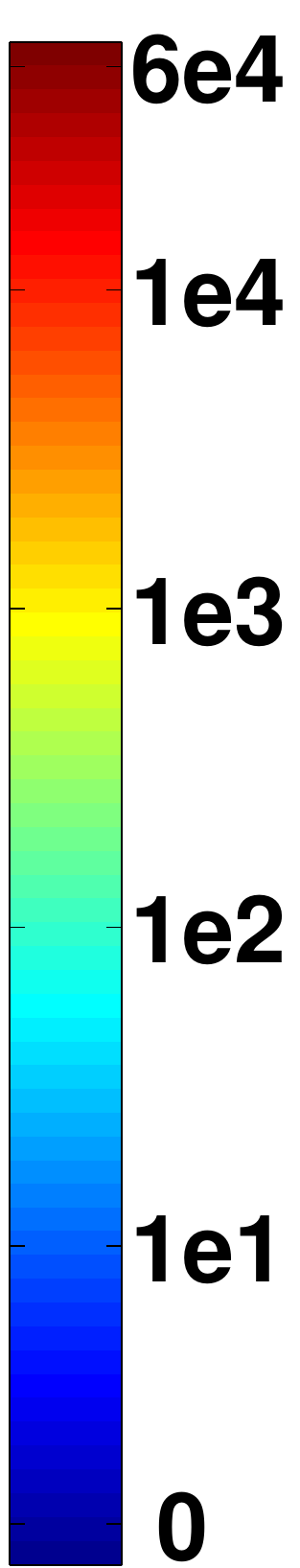}}\\
	\addhspace{{(a) Elastic-net}} & \addhspace{{(b) GraphNet}} & \addhspace{{(c) Fused Lasso}} & \vspace{-4pt}	\\
	\end{tabular} 
	\caption{
		Grid search result for the real resting state data (best viewed in color).
		\textbf{Top row}: the classification accuracy evaluated from 10-fold cross-validation.
		\textbf{Bottom row}: the average number of features selected across the cross-validation folds.
		The $(x,y)$-axis corresponds to the two regularization parameters $\lambda$ and $\gamma$.
	}
	\label{fig:grid,search}
\end{figure}

A common practice for choosing the final set of regularization parameters is to select the choice that gives the highest prediction accuracy.
Based on the grid search result reported in Fig.~\ref{fig:grid,search}, one may be tempted to conclude that the prediction models from GraphNet and fused Lasso are not any better than Elastic-net.
However, the ultimate goal in our application is the discovery and validation of connectivity-based biomarkers, thus classification accuracy by itself is not sufficient.
It is equally important for the prediction model to be interpretable  (\eg, sparse) and inform us about the predictive regions residing in the high dimensional connectome space.
From the grid search, we found that for all three regularization methods, the classifiers achieved a good balance between accuracy and sparsity when approximately $3,000$ features ($\approx 5\%$) were selected out of $p=60,031$.
More specifically, Elastic-net, GraphNet, and fused Lasso achieved classification accuracies of $73.5\%$, $70.3\%$, and $71.9\%$, using an average of $3076$, $3403$, and $3140$ features across the cross-validation folds.
Corresponding regularization parameter values $\{\lambda,\gamma\}$ were: $\{2^{-6},2^{-1}\}$, $\{2^{-5},2^{-2}\}$, and $\{2^{-9},2^{-10}\}$.
Therefore, we further analyzed the classifiers obtained from these regularization parameter values. 

\begin{figure}[t!]\begin{minipage}{\textwidth}
	\renewcommand{\imwidth}  {0.3\linewidth}
	\setlength{\tabcolsep}{1pt} 
	\begin{tabular}{cccc}
	\multicolumn{4}{c}{{\textbf{\large{Median Weight Vector}}}} \vspace{0pt} \\
	\includegraphics[width=\imwidth]{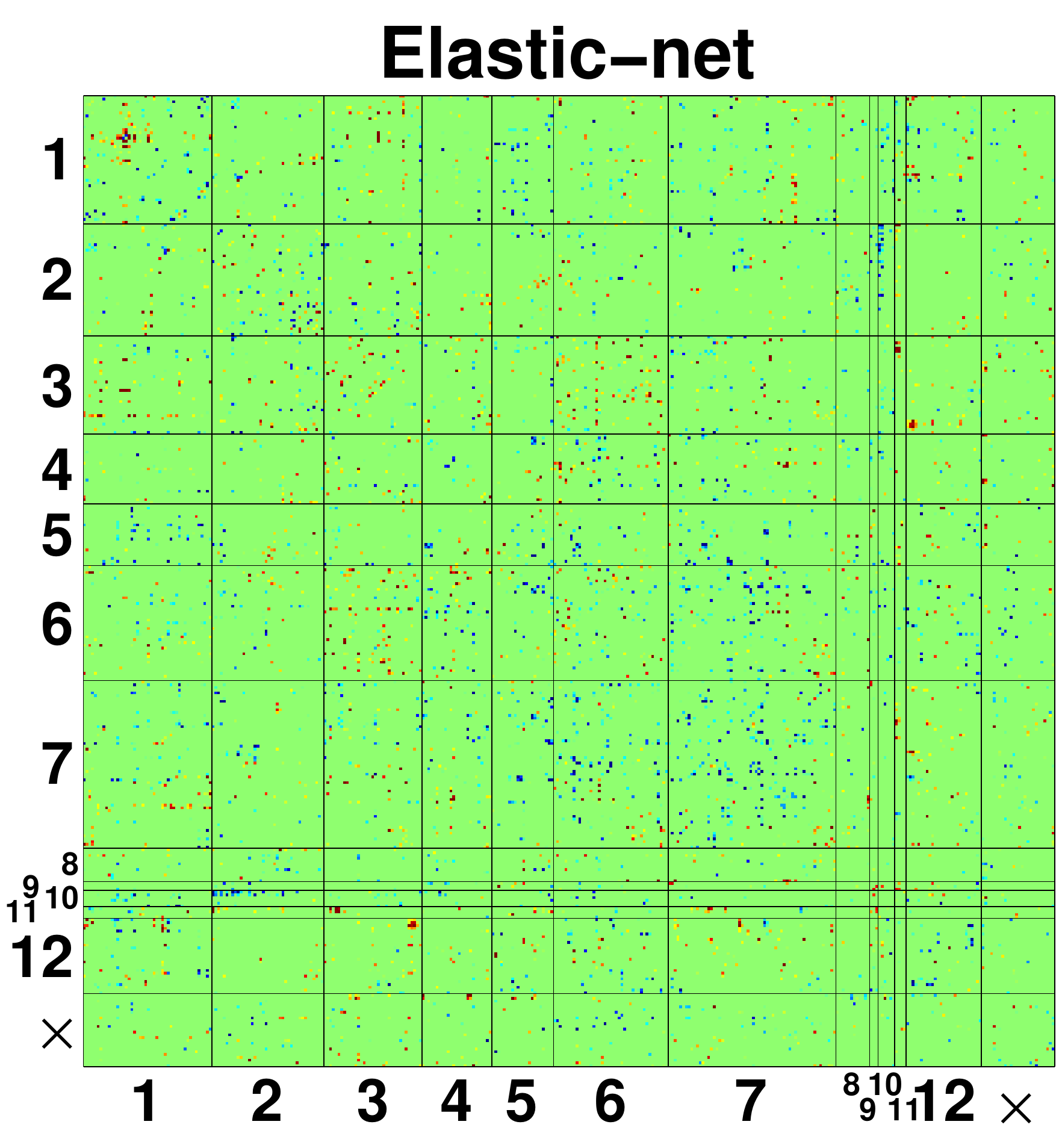} &
	\includegraphics[width=\imwidth]{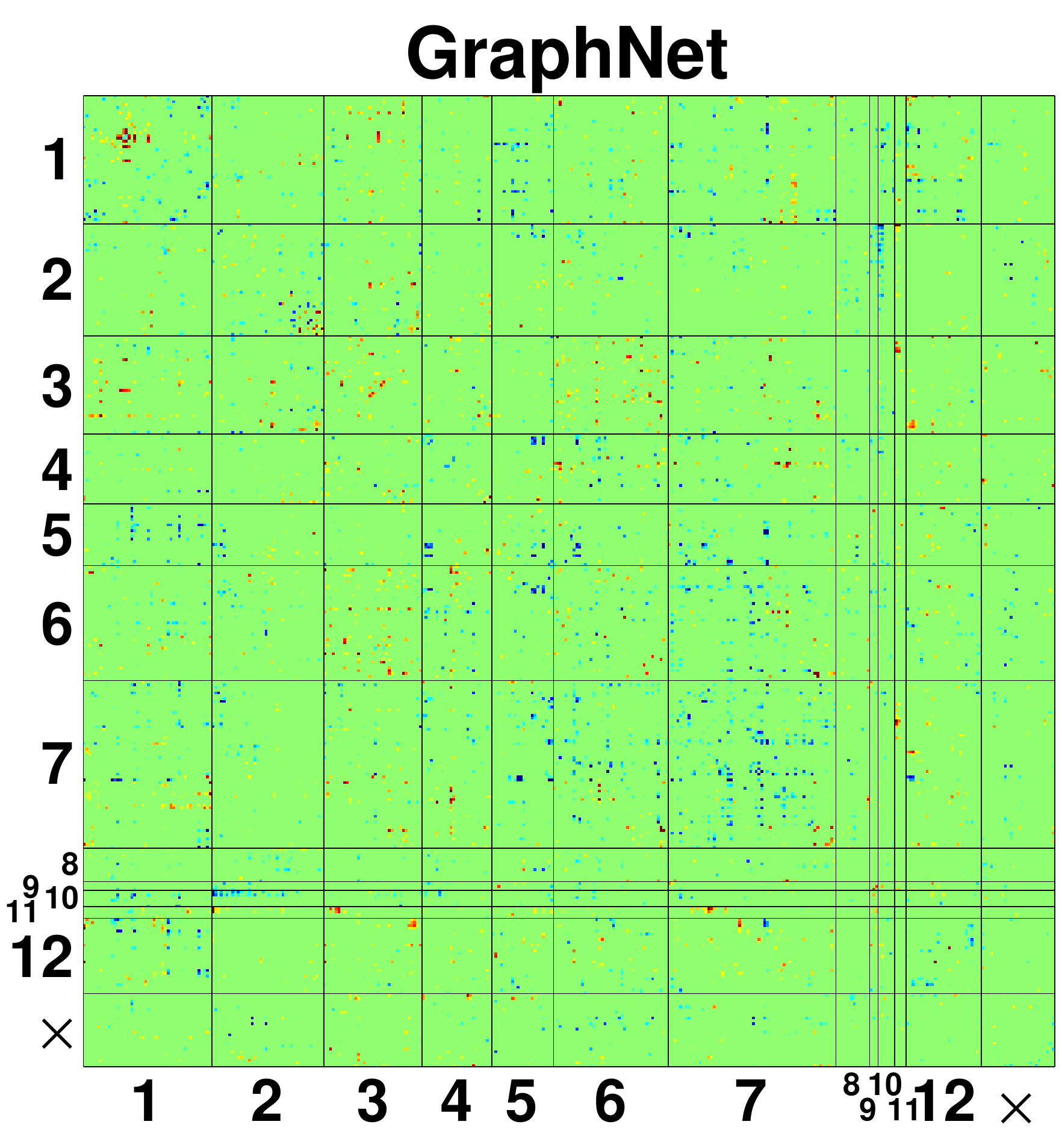} &
	\includegraphics[width=\imwidth]{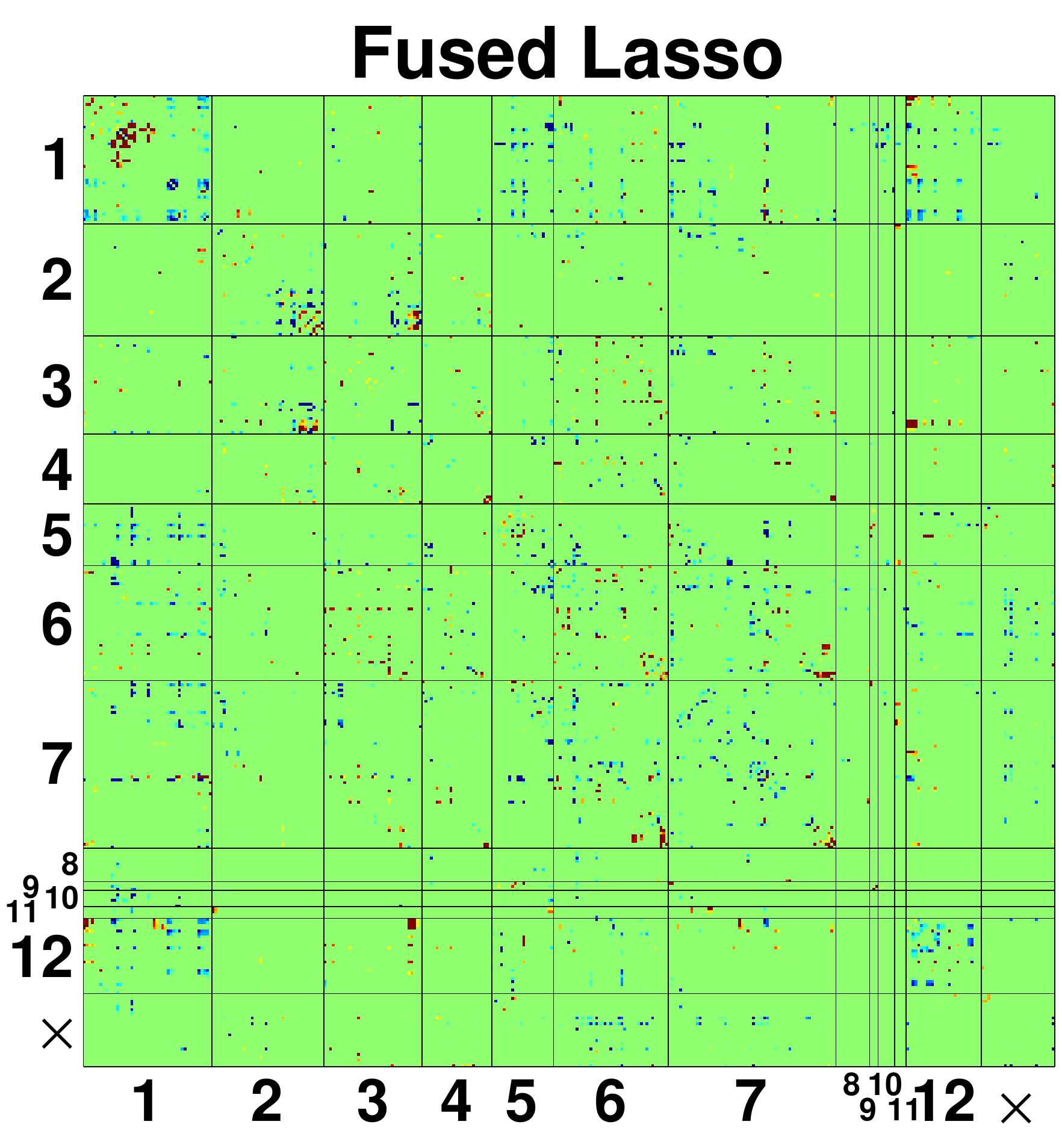} &
	\raisebox{0.02225\linewidth}{\includegraphics[height=0.27675\linewidth]{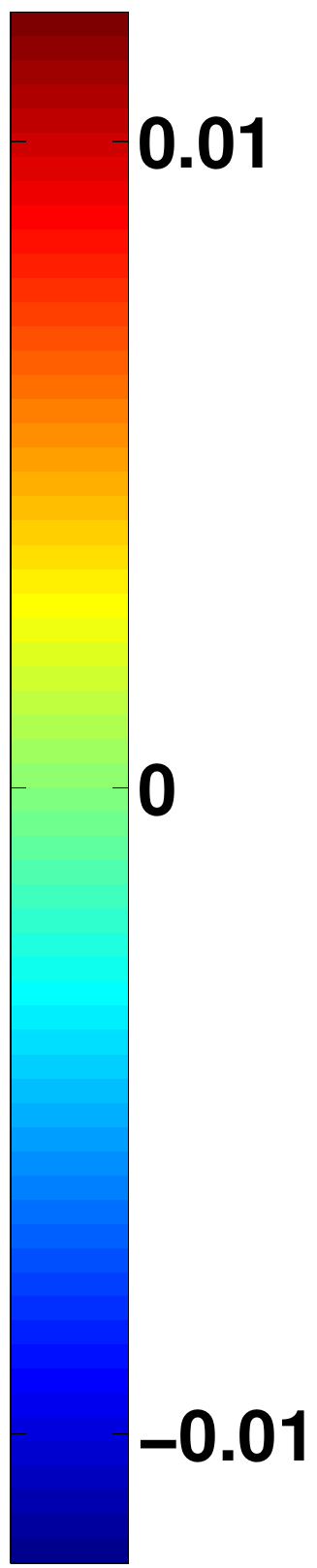}} \vspace{4pt}\\
	\multicolumn{4}{c}{{\textbf{\large{Median Weight Vector Support}}}} \vspace{0pt} \\
	\includegraphics[width=\imwidth]{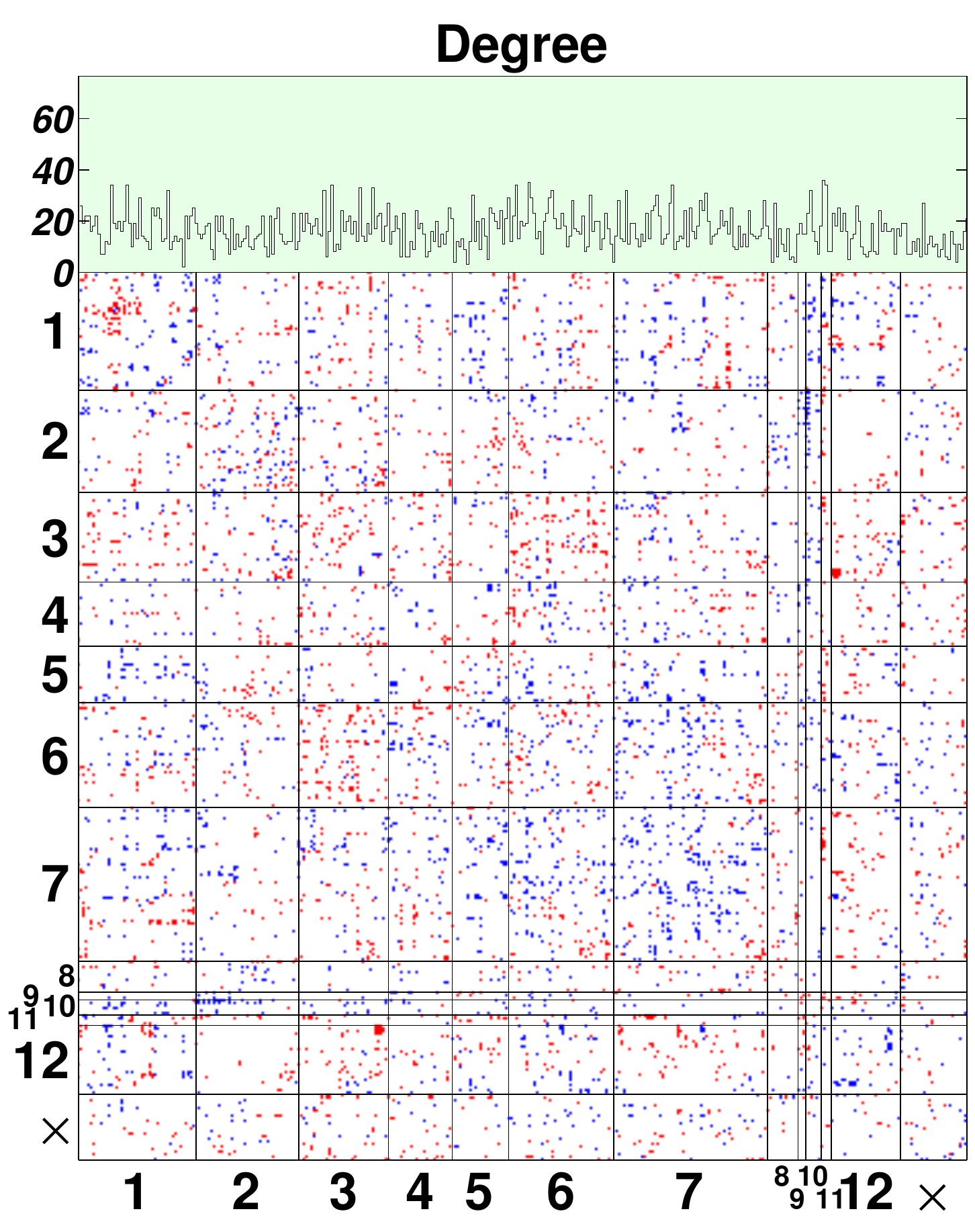} &
	\includegraphics[width=\imwidth]{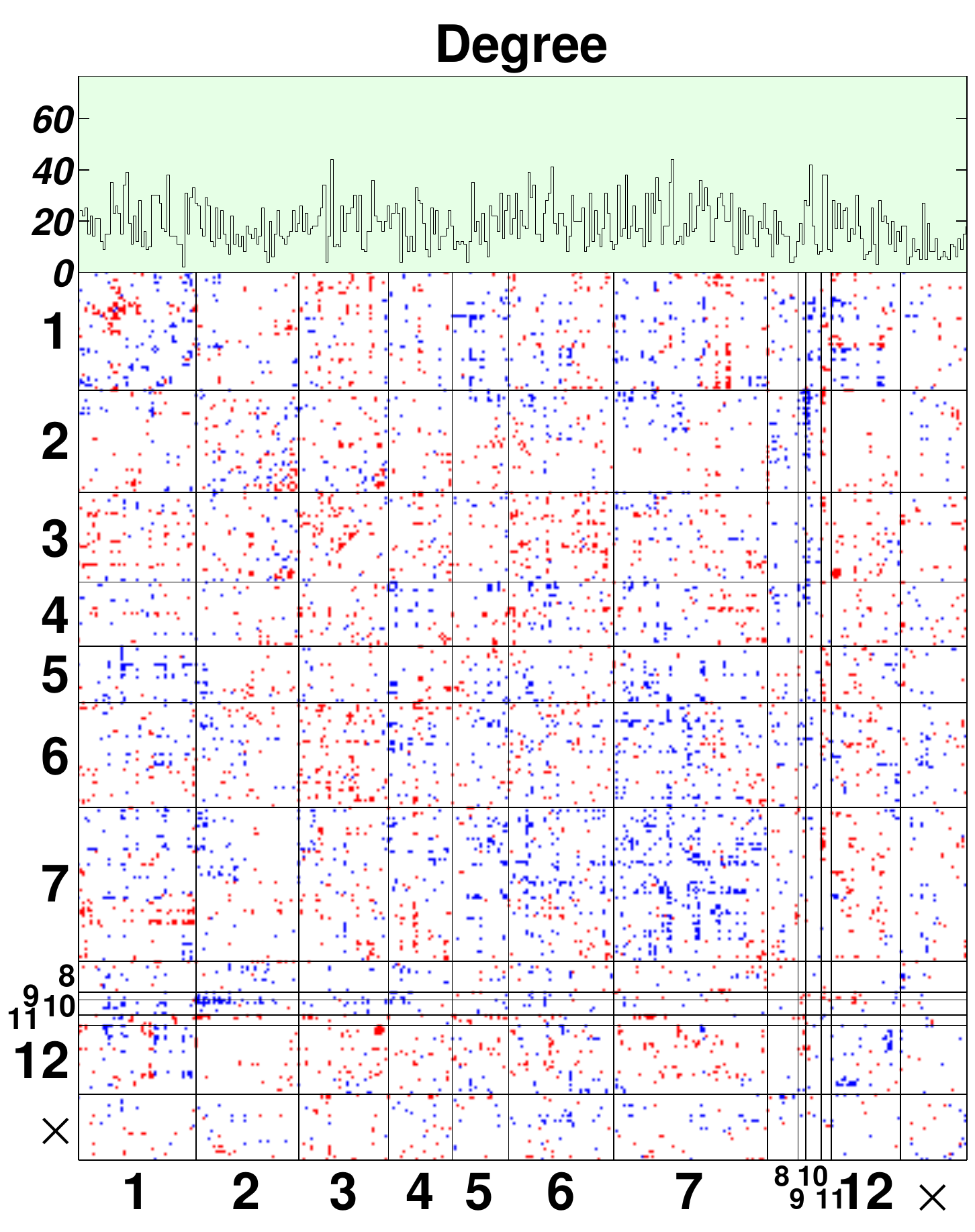} &
	\includegraphics[width=\imwidth]{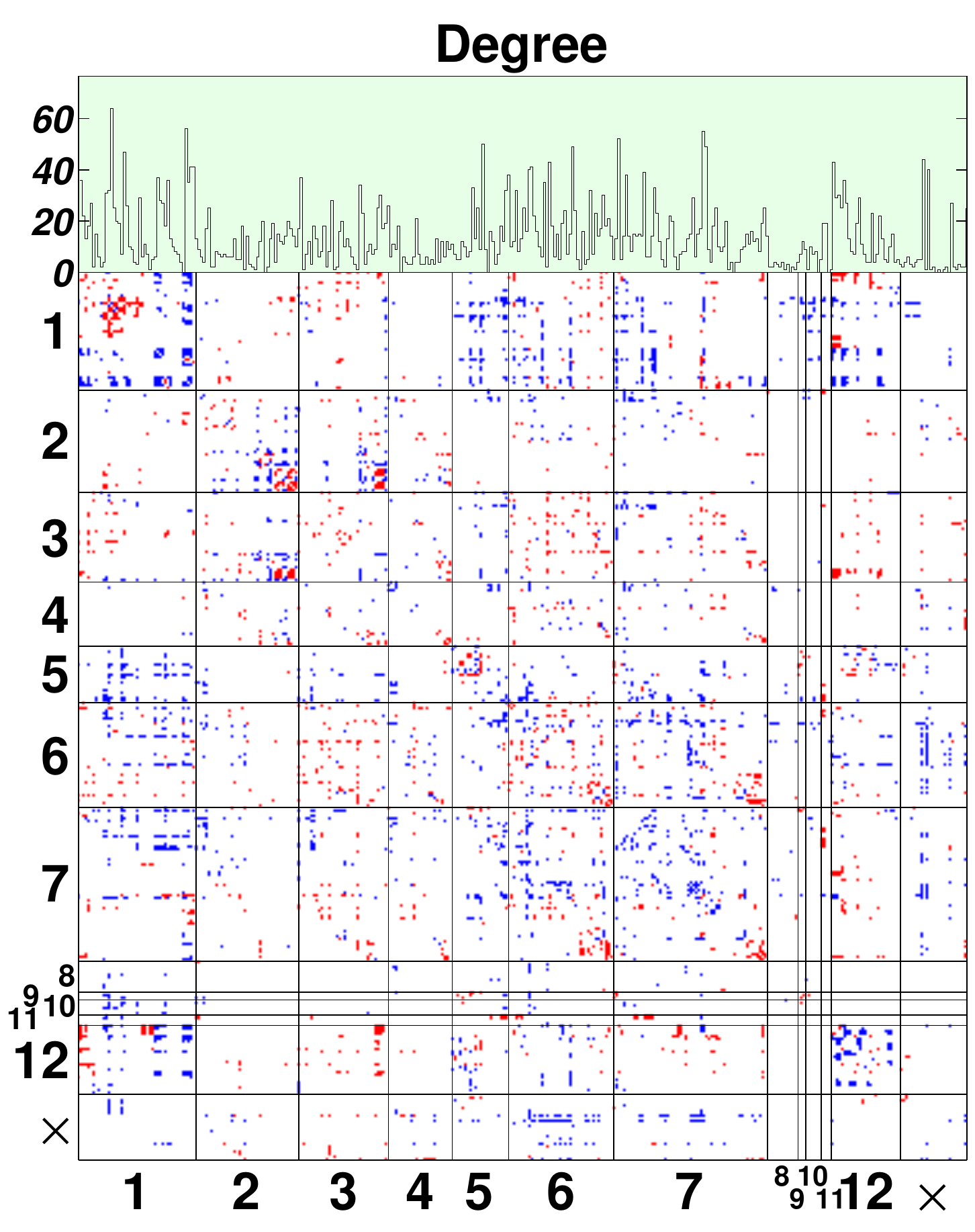} &
	\hspace{-2pt}\raisebox{0.0215\linewidth}{\includegraphics[height=0.2755\linewidth]{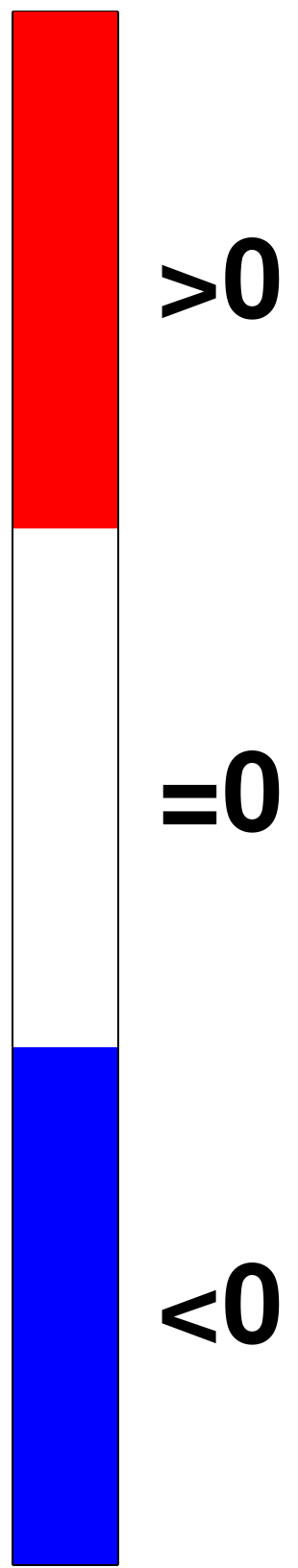}} \vspace{-2pt} \\
\addhspace{{(a) Elastic-net}} & \addhspace{{(b) GraphNet}} & \addhspace{{(c) Fused Lasso}} & \vspace{-5pt}	\\
	\end{tabular} 
	\captionof{figure}{
	Weight vectors (reshaped into symmetric matrices) generated by computing the elementwise median of the estimated weight vectors across the cross-validation folds (best viewed in color).  
	The rows and columns of these matrices are grouped according to the network parcellation scheme proposed by \cite{Yeo:2011}, which is reported in Table~\ref{table:network}.
	The top row displays the heatmap of the estimated weight vectors, whereas the bottom row displays their support structures, with red, blue, and white indicating positive, negative, and zero entries respectively.
	In order to highlight the structure of the estimated weight vectors, the bottom row further plots the degree of the nodes, \ie, the number of connections a node makes with the rest of the network.
	}
	\label{fig:exp,median}
	\vspace{12pt}
	\centering
	\setlength{\tabcolsep}{8pt} 
	\begin{tabular}{llll}
		\hline
		\multicolumn{4}{c}{Network Membership Table ($\times$ is ``unlabeled'')}  \\ 
		\hline\hline
		 1. Visual		& 2. Somatomotor 	 & 3. Dorsal Attention 	& 4. Ventral Attention \\
		 5. Limbic 		& 6. Frontoparietal  & 7. Default	 		& 8. Striatum 		\\
		 9. Amygdala 		& 10. Hippocampus	 & 11. Thalamus		& 12. Cerebellum \\
		\hline
	\end{tabular}\vspace{-4pt}
	\captionof{table}{Network parcellation of the brain proposed by \cite{Yeo:2011}.  In our real resting state fMRI study, the indices of the estimated weight vectors are grouped according to this parcellation scheme; see Fig.~\ref{fig:exp,median}.}
	\label{table:network}
	\vspace{-8pt}
\end{minipage}\end{figure}

During cross-validation, we learned a different weight vector for each partitioning of the dataset.
In order to obtain a single representative weight vector, we took the approach of \cite{Grosenick:2013}, computing the elementwise median of the weight vectors across the cross-validation folds.
Note that this approach possesses attractive theoretical properties; see \cite{Grosenick:2013} and \cite{Minsker:2013} for a detailed discussion.
For visualization and interpretation, we grouped the indices of these weight vectors according to the network parcellation scheme proposed by \cite{Yeo:2011}, and augmented this parcellation with subcortical regions and cerebellum derived from the parcellation of \cite{AAL:2002} (see Table~\ref{table:network}); these weight vectors are then reshaped them into $347\times 347$ symmetric matrices with zeroes on the diagonal.
Furthermore, we generated trinary representations of these matrices in order to highlight their support structures, where red, blue, and white denotes positive, negative, and zero entries respectively.
The resulting matrices are displayed in Fig.~\ref{fig:exp,median}.

From these figures, one can observe that Elastic-net yields solutions that are scattered throughout the connectome space, which can be problematic for interpretation.
In contrast, the weight vector returned from GraphNet has a much smoother structure, demonstrating the impact of the smooth spatial penalty; this is arguably a far more sensible structure from a biological standpoint.
Finally, the weight vector from fused Lasso reveals systematic sparsity patterns with multiple contiguous clusters present, indicating that the predictive regions are compactly localized in the connectome space (\eg, see the rich connectivity patterns present in the intra-visual and intra-cerebellum network).
It is noteworthy the fused Lasso not only appears to identify more densely packed patches of abnormalities in certain regions, it also generates large areas of relative sparsity (\eg, see somatomotor network interconnections with other networks, and the nodes that fall outside the augmented Yeo parcellation scheme, which are labeled~``$\times$''). 
These areas are more sparse in the fused Lasso map, and this appears to be consistent with existing knowledge of connectivity alterations in schizophrenia (see Sec.~\ref{subsec:disc,real,data} of the Discussion). 
In addition, the weight vector estimate from fused Lasso appears to implicate certain nodes more often in connectivity alterations. 
In order to emphasize this point, the bottom row in Fig.~\ref{fig:exp,median} also plots the degree of the nodes, \ie, the number of connections a node makes with the rest of the nodes (this is another example of ``spatial contiguity'' in the $6$-D connectome space).

Finally, in order to convey the regional distribution of the edges recovered by fused Lasso, we rendered implicated edges on canonical $3$-D brains (Fig.~\ref{fig:bnv,median}; these figures were generated with the BrainNet Viewer, \url{http://www.nitrc.org/projects/bnv/}).
We focus on the three sets of network-to-network connections, intra-frontoparietal, frontoparietal-default, and intra-cerebellum, as these three networks have particularly extensive evidence of their involvement in schizophrenia (see Discussion in Sec.~\ref{sec:discussion}).
It is noteworthy that lateral prefrontal cortex, an important region in frontoparietal network, is well represented in the fused Lasso map. 
Edges involving this region represent $39.3\%$ of the intra-frontoparietal connections and $43.6\%$ of the frontoparietal-default network connections. 
This finding is consistent with previous studies of schizophrenia that emphasize the importance of this region (see Discussion in Sec.~\ref{sec:discussion}). 

\renewcommand{\imheight}  {0.3\textwidth}
\renewcommand{\VSPACE}{\vspace{12pt}}
\renewcommand{\VSPACEE}{\vspace{0pt}}
\begin{figure}[ptbh]
	\centering	
	\setlength{\tabcolsep}{8pt}
	\myfbox{\begin{tabular}{cc}
		\vspace{-8pt}\\
		\includegraphics[width=0.225\textwidth]{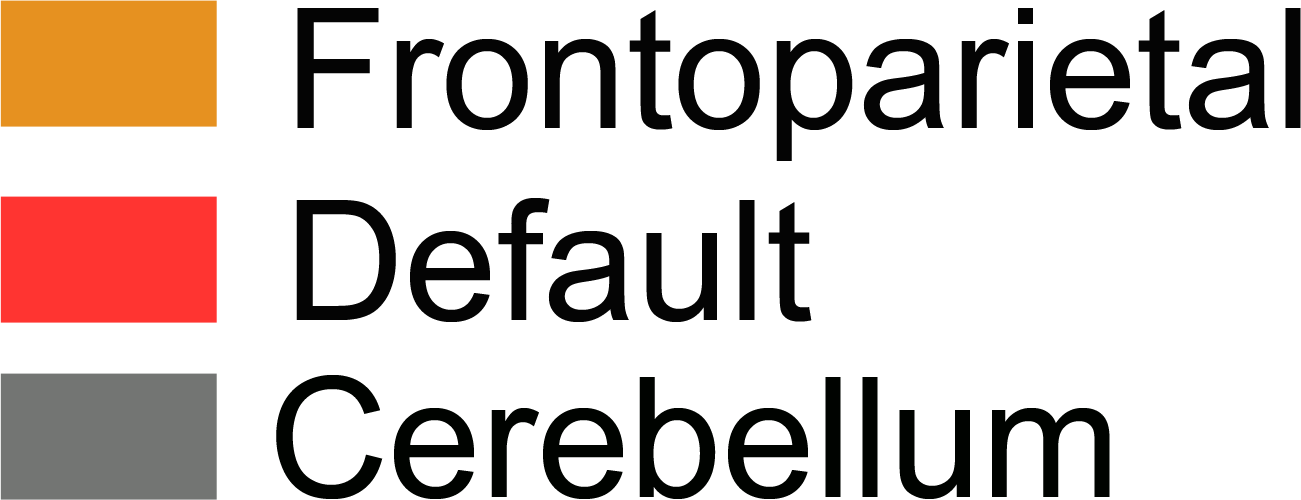} & \hspace{10pt}
		\raisebox{7pt}{\includegraphics[width=0.4\textwidth]{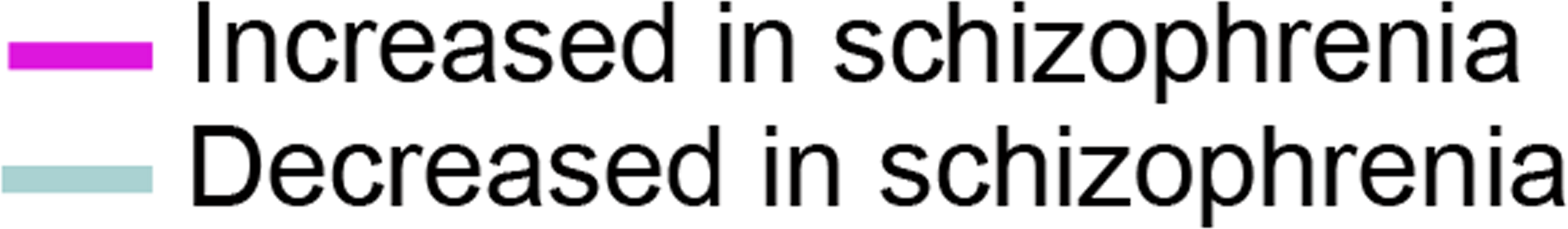}}  \vspace{0pt}\\
		\textbf{(node color)} & \textbf{(edge color)} \vspace{-2pt} \\ 
	\end{tabular}}\vspace{11pt} \\
	\begin{tabular}{ccc}
		\includegraphics[height=\imheight]{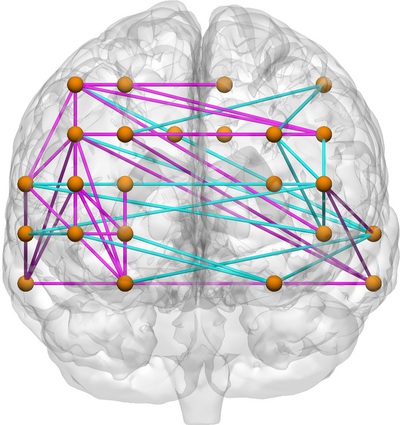}  &
		\includegraphics[height=\imheight]{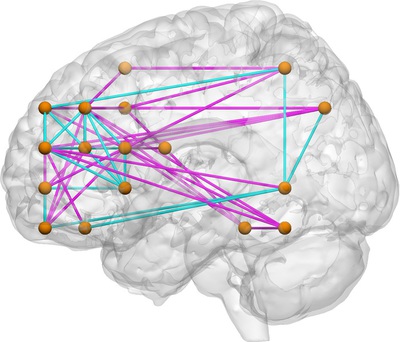}  &
		\includegraphics[height=\imheight]{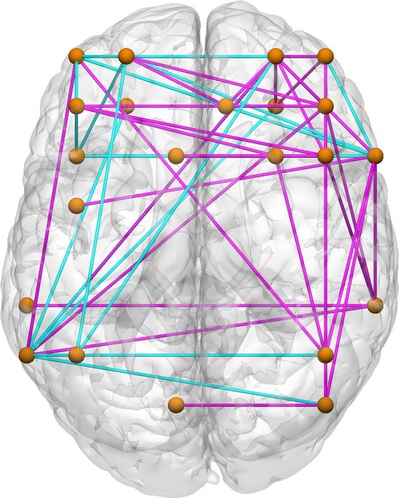} \VSPACEE\\
		\multicolumn{3}{c}{\textbf{\large{Intra-Frontoparietal (6-6)}}} \VSPACE \\
		\includegraphics[height=\imheight]{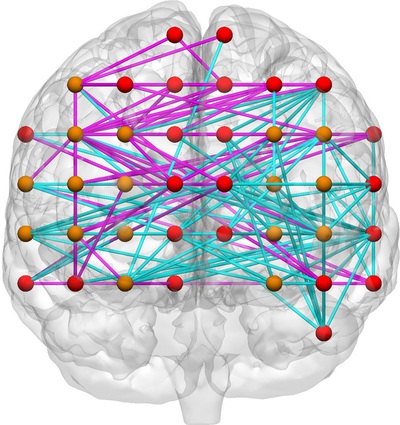}  &
		\includegraphics[height=\imheight]{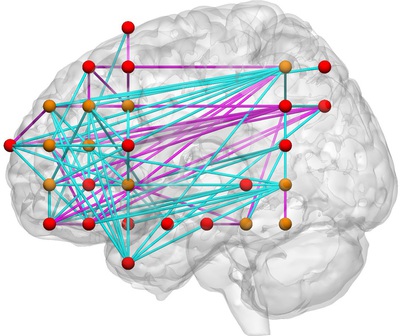}  &
		\includegraphics[height=\imheight]{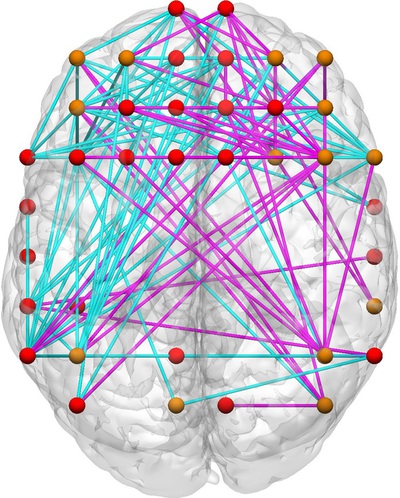} \VSPACEE\\
		\multicolumn{3}{c}{\textbf{\large{Frontoparietal-Default (6-7)}}} \VSPACE \\
		\includegraphics[height=\imheight]{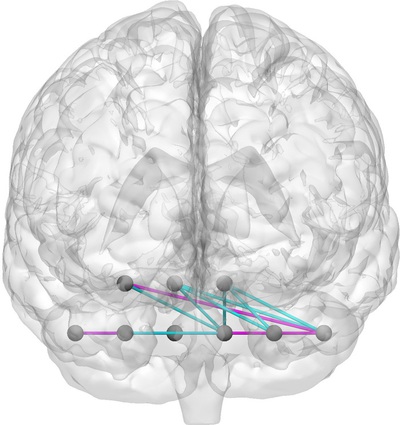}  &
		\includegraphics[height=\imheight]{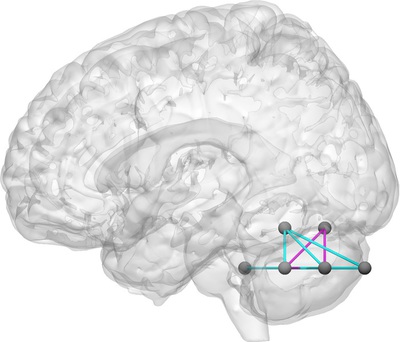}  &
		\includegraphics[height=\imheight]{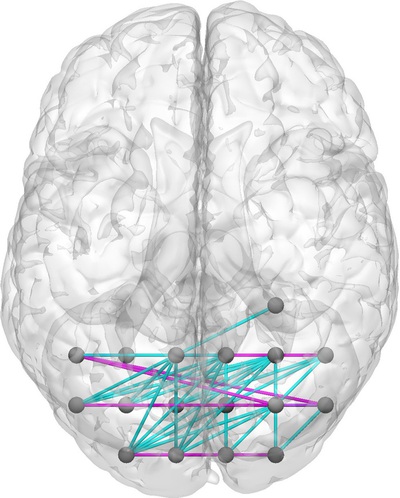} \VSPACEE\\
		\multicolumn{3}{c}{\textbf{\large{Intra-Cerebellum (12-12)}}}  \\
	\end{tabular}
	\caption{
	Nonzero edge values of the median weight vector generated from the fused Lasso regularized SVM.  
	For three sets of network-to-network connections, we rendered abnormal connections separately on anterior, sagittal, and axial views of a canonical brain. 
	Notice the prominent involvement of lateral prefrontal regions in connections within frontoparietal network and in connections between frontoparietal network and default network.
	}
	\label{fig:bnv,median}
\end{figure}

\subsection{Computational considerations}
It is important to note that the benefit of spatial regularization comes with higher computational expense.
To illustrate this point, we ran the ADMM algorithms for Elastic-net, GraphNet, and fused Lasso for $1000$ iterations on the full resting state dataset using regularization parameter values $\{\lambda,\gamma\}=\{2^{-15},2^{-15}\}$ and compared their computation times
(the algorithm for Elastic-net is reported in \ref{appendix,admm,enet}, whereas the algorithms for GraphNet and fused Lasso are reported in Algorithm~\ref{alg:admm}).
This timing experiment was implemented in MATLAB version $7.13.0$ on a desktop PC with Intel quad-core $3.40$ GHz CPU and $12$ GB RAM.
The total computation times for Elastic-net, GraphNet, and fused Lasso were $17.04$ seconds, $96.07$ seconds, and $112.45$ seconds respectively.
The increase in computation time for GraphNet and fused Lasso stems from the fact that unlike the \elltwo-penalty in Elastic-net, the spatial penalty $\norm{\C\w}_q^q,\; {q\in\{1,2\}}$ is not separable across the coordinates of \w.
To address this difficulty, the variable splitting strategy proposed for GraphNet and fused Lasso~\eqref{eqn:admm,splitting2} contains four constraint variables, which is two more than the splitting proposed for Elastic-net~\eqref{eqn:splitting,enet}; as a consequence, the ADMM algorithms for GraphNet and fused Lasso contain two additional subproblems.
Furthermore, the computational bottlenecks of the ADMM algorithms for GraphNet and fused Lasso are the $6$-D FFT and inverse-FFT operations~\eqref{eqn:v4,fft}, which are not conducted for the Elastic-net.
Therefore, if achieving high classification accuracy is the central goal, then Elastic-net would be the most sensible and practical choice, as it yields good classification accuracy and is by far the fastest among the three regularization methods we studied.

Finally, in order to assess the practical utility of our proposed algorithm with respect to existing methods, we conducted another timing experiment using the ADMM algorithm proposed by \cite{Gui-Bo-Ye:2011}, which also solves fused Lasso regularized SVM.
It is important to note that the variable splitting scheme they employ is different from the one we introduce, and consequently, their method requires the following matrix inversion problem to be solved for one of the ADMM updates:
\begin{equation}
\w\iterp \leftarrow 
	\left(
		\X^T\X + \C^T\C + \BI_p
	\right)\inv  
	\big(
		\X^T\Y^T [\va\iter-\ua\iter] +[\vb\iter-\ub\iter] + \C^T[\vc\iter-\uc\iter]
	\big).
	\nonumber
\end{equation}
As suggested in \cite{Gui-Bo-Ye:2011}, we applied the conjugate gradient algorithm to numerically solve this large scale matrix inversion problem\footnote{The conjugate gradient algorithm was ran until either the \elltwo-norm of the residual fell below $1\times 10^{-3}$ or the algorithm reached $60$ iterations.}.
Using the same experimental protocol as our first timing experiment, we ran Ye and Xie's algorithm for $1000$ iterations on the full resting state dataset, which resulted in a total computation time of $331.36$ seconds,
which is nearly three times longer than the algorithm we proposed. 
This illustrates the practical benefit of our proposed variable splitting and data augmentation scheme, which allows all the ADMM updates to be solved analytically.

	\section{Discussion}
	\label{sec:discussion}
Abundant neurophysiological evidence indicates that major psychiatric disorders are associated with distributed neural dysconnectivity \citep{Stephan:2006,Konrad:2010,Muller:2011}.
Thus, there is strong interest in using neuroimaging methods to establish connectivity-based biomarkers that accurately predict disorder status \citep{Kloppel:2012,Sundermann:2013,Cohen:2011}.
Multivariate methods that use whole-brain functional connectomes are particularly promising since they comprehensively look at the network structure of the entire brain \citep{Castellanos:2013,Fornito:2012}, but the massive size of connectomes requires some form of dimensionality reduction. 

In this work, we developed and deployed a multivariate approach based on the SVM~\citep{Cortes:1995} and regularization methods that leverage the 6-D spatial structure of the functional connectome, namely the fused Lasso \citep{Tibshirani:2005} and the GraphNet regularizer \citep{Grosenick:2013}.
In addition, we introduced a novel and scalable algorithm based on the classical alternating direction method \citep{Boyd:2011,Gabay:1976,Glowinski:1975} for solving the nonsmooth, large-scale optimization problem that results from the structured sparse SVM.
Note that most existing multivariate methods in the literature rely on some form of \apriori feature selection or feature extraction (\eg, principal component analysis, locally linear embedding) before invoking some \mbox{``off the shelf''} classifier (\eg, nearest-neighbor, SVM, linear discriminant analysis) \citep{Castellanos:2013}.
In contrast, our feature selection method is not only spatially informed, but is also \emph{embedded} \citep{Guyon:2003}, meaning that feature selection is conducted together with model fitting.
This type of joint feature selection and classification has been rarely applied in the disease prediction framework with functional connectomes.

We used a grid-based parcellation scheme for producing whole-brain resting state functional connectomes (see Section~\ref{subsec:FC,def}), and this has two advantages.
First, it endows a natural ordering and a notion of nearest neighbors among the coordinates of functional connectomes, which is important when defining the neighborhood set for fused Lasso and GraphNet (one may consider predefining an arbitrary graph structured neighborhood set, but we prefer an approach that enforces little \apriori assumption on the structure of the predictive regions).
Second, the finite differencing matrix corresponding to this (augmented) functional connectome has a special structure that allows efficient FFT-based matrix inversion to be applied (this structure is absent when a functional or an anatomical based parcellation scheme is adopted).  When this property is used in tandem with variable splitting, the inner subproblems associated with the proposed ADMM algorithm admit closed form solutions that can be carried out efficiently and non-iteratively.

Using a simulation method and a large real-world schizophrenia dataset, we demonstrate that the proposed spatially-informed regularization methods can achieve accurate disease prediction with superior interpretability of discriminative features.
To the best of our knowledge, this is the first application of structured sparse methods in the context of disease prediction using functional connectomes.

\subsection{Rationale behind spatial regularization}
\label{subsec:why,flasso}
The rationale for using the fused Lasso and GraphNet regularizer can be better appreciated by considering the ``patchiness assumption'' -- the view that major psychiatric diseases manifest in the brain by impacting moderately-sized (\eg, $1,000$ mm$^3$ to $30,000$ mm$^3$) spatially contiguous neural regions. 
This assumption has been repeatedly born out across different imaging modalities.
In structural studies and task-based activation studies, theorists have consistently identified mid-sized blobs in maps of differences between patients and controls \citep{Dickstein:2006, Glahn:2005, Wright:2000}.
In studies of functional connectivity, the patchiness assumption has found clear support. 
The vast majority of previous connectivity studies are seed-based; they create maps of connectivity with a single or a handful of discrete seeds, and compare these maps between patients and controls. 
These studies nearly always report connectivity between patients and controls is altered at one or more discrete medium-sized blobs, similar to structural studies and activation-based studies \citep{Heuvelemail:2010, Konrad:2010,Etkin:2007}.

In addition to actual findings from previous connectivity studies, the patchiness assumption is justified by careful examination of the hypotheses proposed by theorists. 
It is exceedingly common for theorists to state their hypotheses in terms of altered connectivity between two discrete regions or discrete sets of regions. 
For example, based on hypofrontality models of auditory hallucinations in schizophrenia, Lawrie and colleagues \mbox{\citep{Lawrie:2002}} predicted that individuals with schizophrenia would exhibit decreased connectivity between dorsal lateral prefrontal cortex (DLPFC; Brodman's areas 9 and 10), involved in top-down control, and superior temporal gyrus (STG), which is involved in auditory processing. 
Both DLPFC and STG are large structures, and they encompass roughly a dozen nodes each in our grid-based parcellation. 
If Lawrie and colleagues' conjecture is correct, then we should observe alterations in connectivity between the large set of connections that link the nodes that fall within the respective brain structures. 
Moreover, Lawrie and colleagues' hypothesis implies that the predicted changes will be relatively discrete and localized to connections linking these two regions. 
For example, the finding of salt and pepper changes throughout the connectome would of course not support their conjecture. 
Moreover, their hypothesis predicts that even regions that are relatively close to dorsal lateral prefrontal cortex, for example precentral gyrus, involved in motor processing, do not change their connectivity with STG -- the connectivity changes they predict are relatively localized and discrete.

In addition to hypotheses about region-to-region abnormalities, the patchiness assumption is also evident in recent network models of mental disorders. 
In recent years, theorists have recognized that the human brain is organized into large-scale networks that operate as cohesive functional units \mbox{\citep{Bressler:2010,Laird:2011,Yeo:2011}}. 
Each individual network is composed of a set of discrete regions, and each region itself encompasses multiple nodes given a standard, suitably dense parcellation scheme (such as our grid-based scheme).
Concurrent with the rise of this network understanding of neural organization, theorists have proposed models in which psychiatric disorders are seen to involve perturbations in the interrelationships between individual pairs of network, where the remainder of the network interrelationships remain essentially unaffected \mbox{\citep{Menon:2011,Lynall:2010,Tu:2013}}. 
If these network models of disease are correct, then using functional connectivity methods, we should discover that in a psychiatric disease that is proposed to affect the interrelationship between network A and network B, the set of regions that make up network A change their relationship with the set of regions in network B. 
The regions that abut the regions in networks A and B are, by hypothesis, not proposed to alter their connectivity. 
In connectomic space, this pattern would be represented as patchy changes in the sets of connections linking the blobs of contiguous nodes that represent networks A and B, with the remainder of the connectome remaining largely unaffected. 

In sum, actual results from structural, task-based, and connectivity studies suggest the patchiness assumption is reasonable, while close examination of the form of the hypotheses routinely made by psychiatric researchers suggests the assumption underlies theorists' conjectures about disease processes. 
If these claims are correct, then this provides a powerful rationale for both the fused Lasso and GraphNet penalty. 
Fused Lasso penalizes abrupt discontinuities, favoring the detection of piecewise constant patches in noisy contexts. 
Similarly, GraphNet also promotes spatial contiguity, but encourages the clusters to appear in smoother form.
Given that there is a solid basis for expecting that the disease discriminative patterns in functional connectomes will consist of spatially contiguous patches, rather than consisting of salt-and-pepper patterns randomly dispersed throughout the brain, then fused Lasso and GraphNet are well very positioned to uncover these patchy discriminative signatures. 
In addition, the spatial coherence promoted by these spatially-informed regularizers helps decrease model complexity and facilitates interpretation.

\subsection{Simulation study and interpretability of results}
The analytic intuitions discussed above were confirmed in our simulation study.
Here, we imposed ``patchiness'' in the ground truth by introducing clusters of \emph{anomalous nodes} in the synthetic functional connectomes that represent the patient group (see Section~\ref{subsec:synthetic,4d,conn}).
For comparison, we learned SVM classifiers from the training data using the hinge-loss and one of the following regularizers: Lasso, Elastic-net, GraphNet, and fused Lasso.
Our results indicate that fused Lasso and GraphNet not only improved classification accuracy, but also exhibited superior performance in recovering the discriminatory edges with respect to their non-spatially informed counterparts, Lasso and Elastic-net.

\subsection{Application: classifying healthy controls vs. schizophrenic subjects}
\label{subsec:disc,real,data}
Our results indicate that at similar sparsity level, the classification accuracy with Elastic-net, GraphNet, and fused Lasso are comparable.
However, studying the structure of the learned weight vectors reveals the key advantage of GraphNet and fused Lasso: they facilitate interpretation by promoting sparsity patterns that are spatially contiguous in the connectome space.
Fused Lasso recovers highly systematic sparsity patterns with multiple spatially contiguous clusters, including nodes with diffuse connectivity profiles, which is one manifestation of the ``patchiness assumption'' discussed earlier.
On the other hand, the smooth sparsity structure that GraphNet recovers is biologically more sensible than the salt-and-pepper like structure yielded by the Elastic-net.
These decreases in model complexity come without sacrificing prediction accuracy, which fits well with the principle of \emph{Occam's razor} -- given multiple equally predictive models, the simplest choice should be selected.

Finally, additional evidence that fused Lasso recovered more interpretable discriminative features for the schizophrenia dataset comes from comparing visualizations of the respective weight vectors from the three regularizers (see Fig.~\ref{fig:exp,median}). 
The map of the fused Lasso support shows more prominent and clearly localized alterations in connectivity involving frontoparietal network, default network, and cerebellum, among other regions. These networks also exhibited increased node degree, indicating diffuse connectivity alterations with other networks. 
Interestingly, these networks are among the most commonly implicated in schizophrenia.
Frontoparietal network, which has multiple important hubs in prefrontal cortex, is involved in executive processing and cognitive control \citep{Cole:2013}, and has been shown to exhibit abnormal activation (see \cite{Minzenberg:2009} for a quantitative meta-analysis) and connectivity (\cite{Repovs:2011}; \cite{Tu:2013}; see \cite{Fornito:2012} for a review) in schizophrenia. 
Fused Lasso also recovered altered connectivity between frontoparietal network and default mode network, an important brain network involved in autobiographical memory and internally generated mental simulations \citep{Buckner:2008, Raichle:2001}. 
The weight vectors shown in Fig.~\ref{fig:exp,median} and the $3$-D brains shown in Fig.~\ref{fig:bnv,median} evidence a substantial number of aberrant connections between frontoparietal network and default network, with a predominance of reduced connectivity in schizophrenia. 
Frontoparietal network and default network become more interconnected throughout childhood and adolescence \citep{Fair:2007,Anderson:2011}, which might reflect development of top-down cognitive control by frontoparietal regions over default network. 
Reduced connectivity between these two networks is among the most commonly observed findings in connectivity research in schizophrenia \citep{Jafri:2008, Repovs:2011, Woodward:2011, Zhou:2007a, Zhou:2007b}, and has been proposed to reflect disruptions and/or delays in normal trajectories of maturation \citep{Repovs:2011}.
It is also noteworthy that a sizable portion of the aberrant connection within frontoparietal cortex and between frontoparietal network and default network involved dorsal lateral prefrontal cortex (see results in Sec.~\ref{subsec:result,real,data}). 
This region is perhaps the most frequently described as being abnormal in schizophrenia \citep{Bunney:2000, Callicott:2000, Zhou:2007a}. 
A third network highlighted by fused Lasso is cerebellum, which is featured in the influential `cognitive dysmetria' hypothesis of schizophrenia \citep{Andreasen:1998}. 
Abnormalities in cerebellum have been found in post-mortem \citep{Weinberger:1980}, structural \citep{Wassink:1999}, and functional connectivity studies \citep{Mamah:2013}. 

Fused Lasso also tended to generate more sparsity in regions of the connectome that are not associated with schizophrenia pathology. 
For example, connectivity abnormalities in somatomotor network, and in particular its interconnections with attention network and frontoparietal network, have as far as we know not been described in previous schizophrenia connectivity studies. 
The same is true of the nodes that fell outside the Yeo parcellation augmented with subcortical regions and cerebellum. 
These too have not been associated with schizophrenia pathology and tended to be sparser with fused Lasso. 
Overall, fused Lasso appeared to identify regions known from prior research to be involved in schizophrenia and appeared to generate more sparsity outside of these regions, providing some corroboration for the interpretability of fused Lasso findings.

\subsection{Future Directions}
While the spatially-informed disease prediction framework we introduced is capable of yielding predictive and highly interpretable results, there are several open questions that remain for future investigation.
For example, with little modification, the variable splitting and the data augmentation procedure we introduced should be applicable to the isotropic TV penalty, which also promotes spatial contiguity \citep{Wang:2008tv}.
This is important because on one hand, fused Lasso lacks the rotational invariance property of the isotropic TV penalty, whereas on the other hand, isotropic TV penalty is known to introduce artifacts at corner structured regions~\citep{Birkholz:2011,Grasmair:2010}.
Therefore, fused Lasso and isotropic TV penalty can both potentially be problematic for connectomic investigations, and a thorough comparison between these two penalties with our functional connectome data would be an important direction for future investigation.
In addition, there are multiple works that have introduced a framework for achieving structured sparsity by coupling the isotropic TV penalty with the differentiable logistic loss function \citep{Gramfort:2013, Baldassarre:2012,Michel:2011}.
Although our method has the advantage that it can handle non-differentiable loss functions and hence the SVM, the algorithm employed in the above works enjoy a faster rate of convergence than the ADMM algorithm we employ \citep{He:2012,Beck:2009}.
Investigating ways to accelerate our proposed ADMM algorithm will be important for future work  \citep{Deng:2012, Goldstein:2012}.

There are several other interesting extensions that remain for future research as well.
First, functional and anatomical parcellations (which lack a grid structure and hence the BCCB structure) are often used in connectomic investigations.  
Future work should extend our methodology so the ADMM subproblems can be solved efficiently in analytic form even when a irregularly structured parcellation scheme is used (although the ADMM algorithm proposed by \cite{Gui-Bo-Ye:2011} is applicable in this setup, their approach requires an iterative update to be used to numerically solve one of the ADMM subproblems).
Furthermore, with the emergence of various data sharing projects in the neuroimaging community such as Autism Brain Imaging Data Exchange (ABIDE) \citep{Martino:2013}, ADHD-$200$ \citep{ADHD200}, $1000$ Functional Connectomes Project, and the International Neuroimaging Data-sharing Initiative (INDI) \citep{Mennes:2012}, there is a need for a principled framework to handle the heterogeneity introduced by aggregating the data from multiple imaging centers.
Toward this end, we are seeking ways to combine the currently presented spatial regularization scheme and multi-task learning \citep{Caruana:1997}, where the tasks correspond to the imaging centers from which the resting state scans originate.
One particular approach we have in mind for this is to replace the \ellone-regularizer in the objective function~\eqref{eqn:costfx} with the $\ellone/\elltwo$ mixed-norm regularizer \citep{Lounici:2009, Gramfort:2012}, which encourages the weight vectors across the different tasks to share similar sparsity patterns (a structure often referred to as block-sparsity).
Our proposed ADMM algorithm can easily be modified to handle this change, as this simply amounts to replacing the scalar soft-threshold operator for the \vb update~\eqref{eqn:v2,update2} with the vector soft-threshold operator (see \cite{Gramfort:2012}).
Finally, a more sophisticated approach for parameter tuning is needed, ideally a model selection strategy that provides statistical guarantees \citep{Cawley:2010}. 
Resampling-based approaches  \citep{Bach:2008c,Varoquaux:2012} such as stability selection \citep{Meinshausen:2010} may be considered, albeit these methods can be computationally demanding in high dimension.

	\section{Conclusions}
	\label{conclusions}
In this work, we introduced a regularized ERM framework that explicitly accounts for the $6$-D spatial structure in the connectome via the fused Lasso and the GraphNet regularizer. 
We demonstrate that our method recovers sparse and highly interpretable patterns across the connectome while maintaining predictive power, and thus could generate new insights into how psychiatric disorders impact brain networks.

\subsection*{Acknowledgments}
T. Watanabe and C. Scott's research was supported by NIH grant P01CA087634 and by NSF Grant CCF 1217880.
C. Sripada's research was supported by NIH grant {K23-AA-020297}, Center for Computational Medicine Pilot Grant, and the John Templeton Foundation.
The authors would like to thank A. Hero and J. Fessler, University of Michigan, for the valuable discussions and their insightful feedbacks.
The authors would also like to thank Robert C. Welsh, University of Michigan, for providing us with \texttt{ConnTool}, a functional connectivity analysis package.

\appendix 
\setcounter{figure}{0}

\section{Details on the data augmentation scheme}
\label{appendix:data,aug}

As discussed in Sec.~\ref{subsec:var,split}, the augmentation matrix $\A\in\reals^{\ptil\times p}$ aims to rectify the irregularities in the Laplacian matrix $\C^T\C$.
To gain a better understanding about \A, it is best to think of it as a concatenation of two matrices, ${\A=\A_2\A_1}$.
We refer to $\A_1\in\reals^{\pstar\times p}$ and $\A_2\in\reals^{\ptil\times \pstar}$ as the \emph{first level} and the \emph{second level} augmentation matrix respectively.

\begin{figure}[t!]
	\centering
	\renewcommand{\imwidth}  {0.425\linewidth}
	\begin{subfigure}[t]{\imwidth}
		\centering
		\Large
		$\underbrace{
			\myfbox{\includegraphics[width=0.85\linewidth]{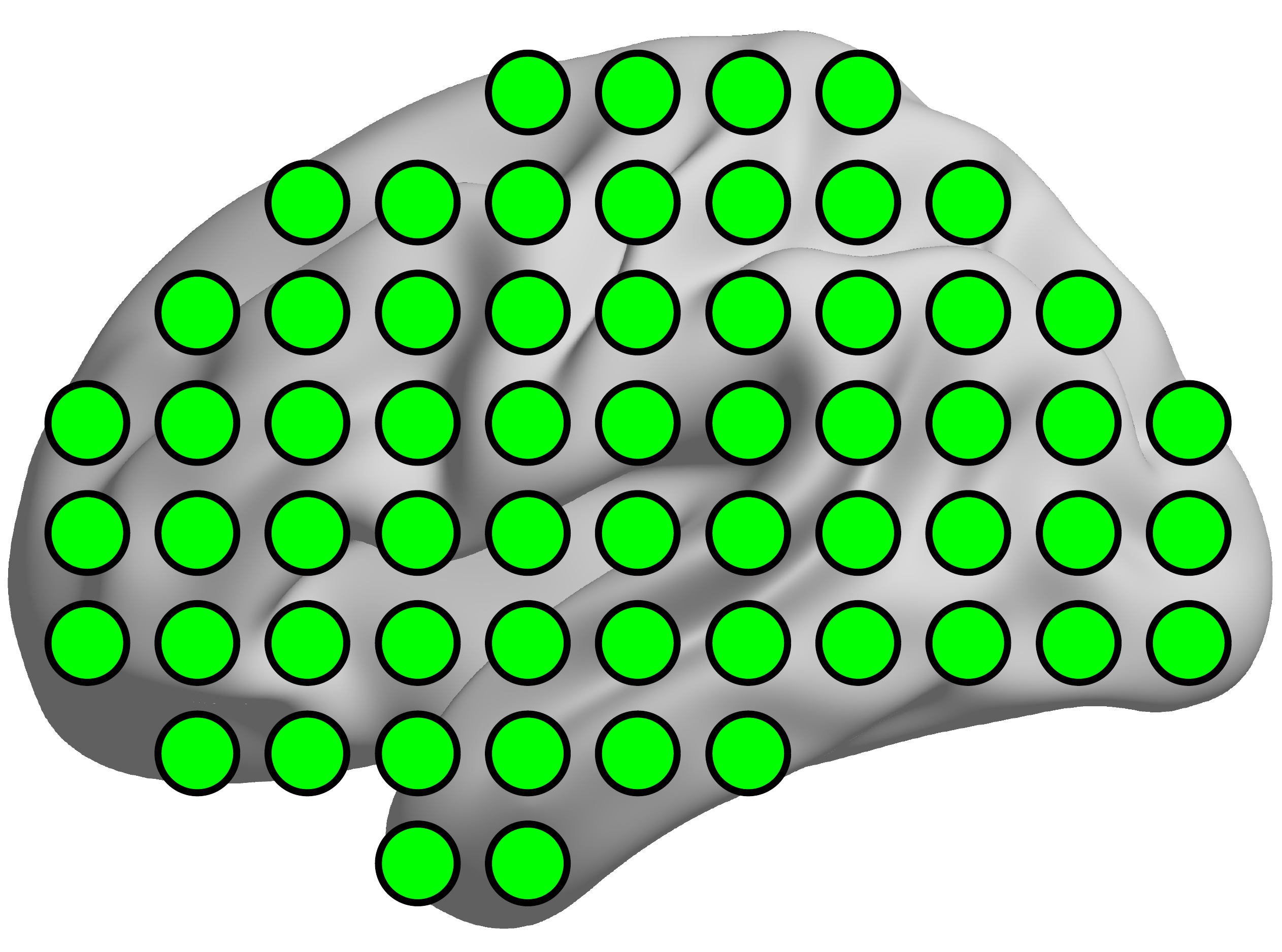}}
			}_\x
		$
		\caption{Original Functional connectome}
	\end{subfigure}
	\hspace{0.05\linewidth}
	\begin{subfigure}[t]{\imwidth}
		\centering
		{\Large
		$\underbrace{
			\myfbox{\includegraphics[width=0.85\linewidth]{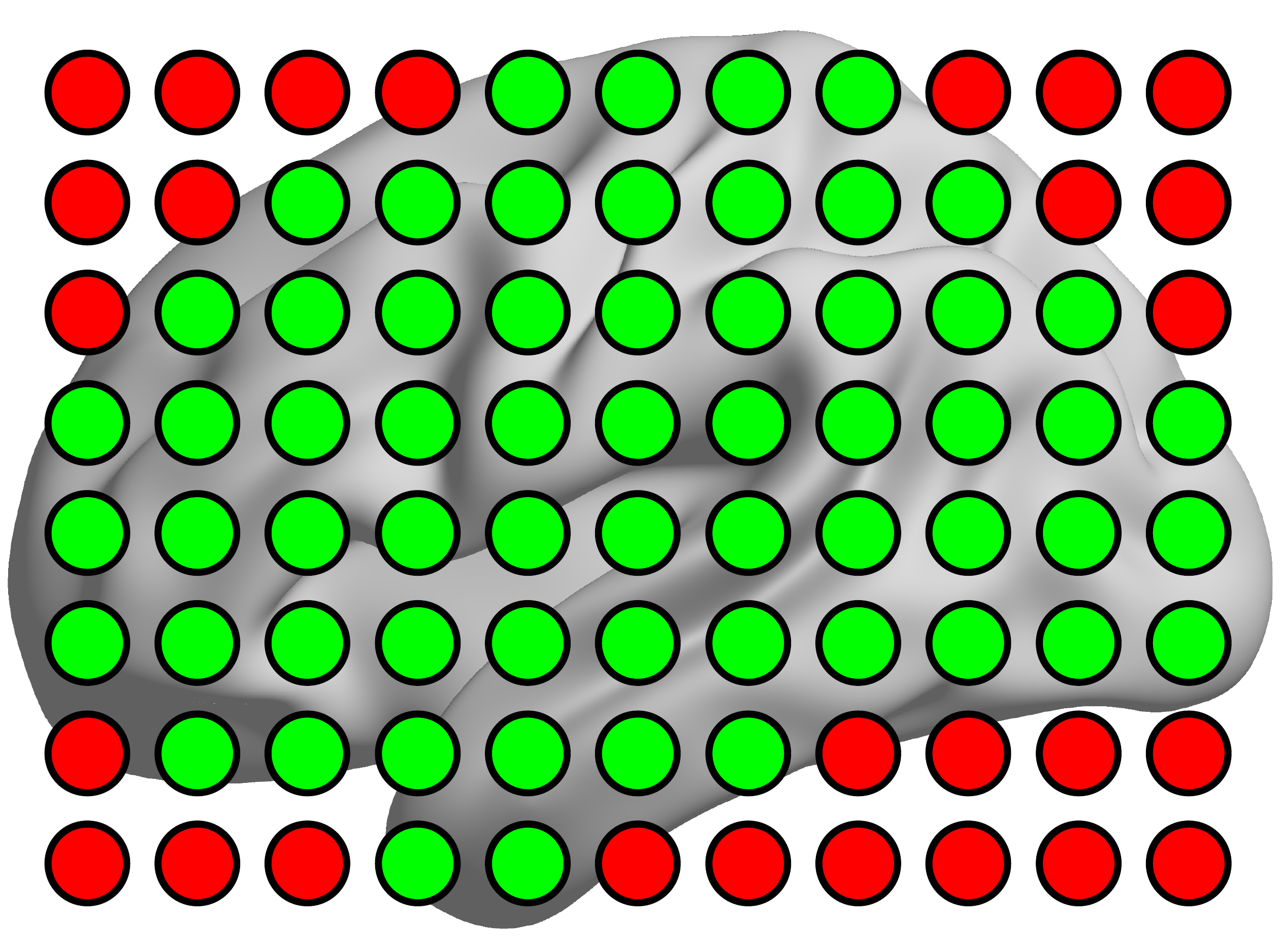}}
			}_{\xstar=\A_1\x}
		$}
		\caption{Intermediate augmented connectome}
	\end{subfigure}
	\caption{
		The effect of the first level augmentation matrix $\A_1$.
		\textbf{Left:} the original functional connectome \x only contains edges between the nodes placed on the support of the brain (represented by the green nodes).
		\textbf{Right:}
		$\A_1$ pads extra zero entries on \x to create the intermediate augmented connectome \xstar.
		Here, \xstar can be treated as if the nodes were placed throughout the entire rectangular FOV (the red bubbles represent nodes that are outside the brain support), as its entries contain all possible edges between the green and red nodes; the edges that connect with the red nodes all have zero values. 
	}\label{fig:augmat1}\vspace{0.025\linewidth}
	\renewcommand{\VSPACE}{\vspace{-0.135in}}
	\begin{subfigure}[b]{0.39\linewidth}
		\centering
		\small
		\(
\xstar = \ba{c} \xstar(\r_2,\r_1) \\ \xstar(\r_3,\r_1) \\ \vdots \\ \xstar(\r_d,\r_1) \\ \hline \VSPACE \\
\xstar(\r_3,\r_2) \\ \xstar(\r_4,\r_2) \\ \vdots \\ \xstar(\r_d,\r_2) \\ \hline
\vdots \\ \hline \VSPACE \\  \xstar(\r_d,\r_{d-1})  \ea,
		\)
		\vspace{8pt}
		\caption{Intermediate augmented connectome}
	\end{subfigure}
	\begin{subfigure}[b]{0.6\linewidth}
		\small\centering
		\(
\xtil=\A_2\xstar
=
\ba{c} \dored{\xtil(\r_1,\r_1)} \\ \xtil(\r_2,\r_1) \\ \xtil(\r_3,\r_1) \\ \vdots \\ \xtil(\r_{d},\r_1) \\ \hline \VSPACE \\  
\dored{\xtil(\r_1,\r_2)} \\ \dored{\xtil(\r_2,\r_2)} \\ \xtil(\r_3,\r_2) \\ \vdots \\ \xtil(\r_{d},\r_2) \\ \hline \vdots \\ \dored{\xtil(\r_{d},\r_{{d}})}  \ea
=
\ba{c} \dored{0} \\ \xstar(\r_2,\r_1) \\ \xstar(\r_3,\r_1) \\ \vdots \\ \xstar(\r_{d},\r_1) \\ \hline \VSPACE \\ \dored{0} \\ \dored{0} \\ \xstar(\r_3,\r_2) \\ \vdots \\ \xstar(\r_{d},\r_2) \\ \hline \vdots \\ \dored{0}  \ea
		\)
		\caption{Augmented functional connectome}
	\end{subfigure}
	\caption{
		The effect of the second level augmentation matrix $\A_2$.  
		The entries of \xstar represent edges localized by $6$-D coordinate points $\left\{(\r_j,\r_k) \; |\;  j>k\right\}$, where $\r_j=(x_j,y_j,z_j)$ and $\r_k=(x_k,y_k,z_k)$ are the $3$-D locations of the node pairs defining the edges.
		$\A_2$ fixes the asymmetry in the coordinates of \xstar by padding zero entries to accommodate for the $6$-D coordinate points $\left\{(\r_j,\r_k) \; |\;  j\leq k\right\}$; these are the diagonal and the upper-triangular entries in the cross-correlation matrix that were disposed for redundancy.
	}
	\label{fig:augmat2}
\end{figure}

\paragraph{Role of $\A_1$}
The first source of irregularities is that the nodes defining the functional connectome $\x\in\reals^p$ are placed only on the brain, not the entire rectangular FOV.
As a consequence, \x only contains edges among the nodes placed on the support of the brain (represented by the green nodes in Fig.~\ref{fig:augmat1}).
To fix these irregularities, $\A_1$ pads extra zero entries on \x to create an \emph{intermediate} augmented connectome $\xstar=\A_1\x$, where $\xstar\in\reals^\pstar$.
Here, \xstar can be treated as if the nodes were placed throughout the entire rectangular FOV; the red nodes in Fig.~\ref{fig:augmat1} represent a set of \emph{ghost nodes} that were not originally present. 
The coordinates of \xstar contain all possible edges between the \emph{ghost nodes} and the original set of nodes, where the edges connected with the \emph{ghost nodes} have zero values.

\paragraph{Role of $\A_2$}
The second source of irregularities is that \x (and \xstar) lack a complete \mbox{$6$-D} representation since it only contains the lower-triangular part of the cross-correlation matrix. 
Consequently, the coordinates of \xstar lack symmetry, as their entries only contain edges for the following set of $6$-D coordinate points: $\left\{(\r_j,\r_k) \; |\;  j>k\right\}$, where $\r_j=(x_j,y_j,z_j)$ and $\r_k=(x_k,y_k,z_k)$ are the $3$-D locations of the node-pairs defining the edges.
Matrix $\A_2$ fixes this asymmetry by padding zero entries to fill in for the $6$-D coordinate points $\left\{(\r_j,\r_k) \; |\;  j\leq k\right\}$, which correspond to the diagonal and the upper-triangular entries in the cross-correlation matrix that were disposed due to redundancy (see Fig.~\ref{fig:augmat2}).
Applying $\A_2$ on $\xstar=\A_1\x$ provides the desired augmented functional connectome $\xtil=\A_2\xstar=\A\x$, and similarly the augmented weight vector $\wtil=\A\w$.
Here, \xtil and \wtil contain the full set of $6$-D coordinate points $\left\{(\r_j,\r_k) \; |\;  j,k \in [d]\right\}$, where $d$ is the total number of nodes on the rectangular FOV including the \emph{ghost nodes} (\ie, both the green and the red nodes in Fig.~\ref{fig:augmat1}).
Note that dimension $\ptil$ of the augmented functional connectome is $\ptil=d^2$, and the total number of adjacent coordinates $\tilde{e}$ in this augmented $6$-D connectome space is $\tilde{e}=6\ptil$.

\section{ADMM updates for Elastic-net}
\label{appendix,admm,enet}
The unconstrained formulation of the Elastic-net regularized ERM problem reads
\begin{equation}
	\argmin{\w\in\reals^p} \frac{1}{n}\Loss(\YXw) + \lambda\norm{\w}_1 + \frac{\gamma}{2}\norm{\w}^2_2 \;,
	\nonumber
\end{equation}
which can be converted into the following equivalent constrained formulation:
\begin{equation}
	\begin{array}{c}
		\dstyle\minimize{\w,\va \vb}
			\frac{1}{n}\Loss(\va)+\lambda\norm{\vb}_1+ \frac{\gamma}{2}\norm{\w}^2_2 \;
			\vspace{0.01\linewidth}
		\text{ subject to }  \YXw =\va, \; \w=\vb \;.
	\end{array}
	\label{eqn:splitting,enet}
\end{equation}
With this variable splitting scheme, the correspondence with the ADMM formulation~\eqref{eqn:canonical,admm} is
\begin{equation}
	\begin{array}{c}
		\fbar(\xbar)=\frac{\gamma}{2}\norm{\w}^2_2 , \;\;\;
		\gbar(\ybar)=\dstyle\frac{1}{n}\Loss(\va) + \lambda\norm{\vb}_1  
		\vspace{0.02\linewidth} \\
		\Abar = 	\ba{cc} 	\Y\X	 \\ \I \ea 	, \quad
		\xbar = \w, \quad
		\Bbar = 	-\I		, \quad
		\ybar = \bmat \va \\ \vb \emat \;.
		\end{array}
	\nonumber
\end{equation}
and the ADMM updates for \xbar~\eqref{eqn:admm,xbar,update} and \ybar~\eqref{eqn:admm,ybar,update} decomposes into subproblems
\begin{alignat}{1} 
	\w\iterp 	\leftarrow \argmin{\w}&
		 \Bigg\{ \frac{\gamma}{2}\norm{\w}^2
		+ \normsq{\Y\X\w-\left(\va\iter-\ua\iter\right)} 
		+ \normsq{\w-\left(\vb\iter-\ub\iter\right)} \Bigg\} \nonumber \\
	\va\iterp \leftarrow \argmin{\va}
		& \left\{ \frac{1}{n}\Loss(\va) + \frac{\rho}{2}\normsq{\va-\left(\Y\X\w\iterp+\ua\iter\right)} \right\}
			\nonumber \\
	\vb\iterp \leftarrow \argmin{\vb} 
		& \left\{\lambda\norm{\vb}_1 + \frac{\rho}{2}\normsq{\vb-\left(\w\iterp+\ub\iter\right)} 
			\nonumber\right\} \;.
\end{alignat}
The update for \w is
\begin{align}
	\w\iterp \leftarrow \left(\rho\X^T\X + [\gamma+\rho]\BI_p\right)\inv  
		\Big(&\rho\X^T\Y^T [\va\iter-\ua\iter] 
		+\rho[\vb\iter-\ub\iter] \Big) \, \nonumber
\end{align} which can be solved efficiently via inversion Lemma \eqref{eqn:inv,lemma}.
The update for \va and \vb is identical to \eqref{eqn:v1,update1} and \eqref{eqn:v2,update1} described in Sec.~\ref{subsec:admm,steps}, which can be solved via coordinate-wise proximal operators \eqref{eqn:v1,update2} and \eqref{eqn:v2,update2}.
The dual variable update \eqref{eqn:admm,dual,update} is a trivial matrix-vector multiplication.

	\bibliographystyle{IEEEtran}	
	\bibliography{mybib}
\end{document}